%% file: iclr2027_main.tex
\documentclass{article} % For LaTeX2e
\usepackage{arxiv_style,times}

\input{math_commands.tex}

\usepackage{hyperref}
\usepackage{url}
\usepackage{subcaption}
\usepackage{graphicx}
\usepackage{amssymb}
\usepackage{booktabs}
\usepackage{tabularx}
\usepackage{pifont}
\usepackage[table]{xcolor}

\title{Examining Social Attribution in LLM Reasoning: A Theory-Guided Probing Methodology}

\author{%
\textbf{Zhaoxin Yu\textsuperscript{1,2}\quad
Qingchao Kong\textsuperscript{1}\quad
Dajun Zeng\textsuperscript{1,2}\quad
Wenji Mao\textsuperscript{1,2}\thanks{Corresponding author. Email: \{\href{mailto:yuzhaoxin2024@ia.ac.cn}{yuzhaoxin2024},\href{mailto:wenji.mao@ia.ac.cn}{wenji.mao}\}@ia.ac.cn}}\\[0.5em]
{\small
\textsuperscript{1}State Key Laboratory of Multimodal Artificial Intelligence Systems,\\
Institute of Automation, Chinese Academy of Sciences\\[0.25em]
\textsuperscript{2}School of Artificial Intelligence, University of Chinese Academy of Sciences\endgraf}}

\begin{document}

\maketitle

\begin{abstract}
Large language models (LLMs) are increasingly deployed in sociotechnical systems to facilitate social interactions, in which \textit{social attribution}, the reasoning process attributing external events to the causes and reasons of agents' social behaviors, plays a critical role.
Such reasoning processes involve the judgments of social cause, responsibility, and blame/credit to agents.
Although attributional models are well-studied in social psychology and social cognition, represented by Attribution Theory, it is still underexplored in the AI and computing community, and LLM social reasoning in particular.
This paper aims to advance the understanding of LLM in a fundamental social aspect by providing the first systematic exploration of LLM social attribution. 
Our work focuses on responsibility and blame attributions, examining current LLMs' performances in making responsibility and blame judgments, as well as their internal mechanisms underlying such judgments.
Specifically, guided by attribution theory, we first construct a social attribution benchmark consisting of a \textit{Vignette} subset based on classic scenarios from attribution theory research and a \textit{Reality} subset constructed based on real-world social narratives, yielding \textbf{7,639} responsibility/blame judgment questions.
On this basis, we evaluate \textbf{32} representative LLMs and \textbf{5} basic non-LLM baselines.
To further explore the internal mechanisms underlying the LLM judgment process, we develop a probing-based methodology to investigate the latent-space representations of \textbf{5} key attribution dimensions and the consistency of their influences on LLM judgments compared to those in human social attribution.
Our research findings reveal that current LLMs exhibit measurable but incomplete agreement with human responsibility and blame judgments, and meanwhile, this agreement is positively correlated with model size.
At a deep level, some attribution dimensions are systematically decodable from specific positions in LLM hidden states, and their influences on the final judgment are consistent with those indicated by human Attribution Theory.
The dataset and associated code are available at \url{https://github.com/Yuzhaoxin946/SAB-Bench}.
\end{abstract}

\input{old_intro}

\section{Related Work}

\noindent \textbf{LLMs for normative judgments.} Recent research systematically evaluates LLMs' attribution-related behaviors in social contexts \citep{xu2024llmsocial_survey_ipm}.
Benchmarks typically address two categories of judgment: (1) social-norm judgments, using everyday, context-dependent rules of thumb to evaluate whether models generate or select norm-consistent explanations and behavioral recommendations \citep{forbes2020socialnorm, jiang2021socialnorm, yuan2024socialnorm, agarwal2024ethical, jinlanguage, ji2025moralbench, chakraborty2025structuredmoral}; and (2) legal judgments, grounded in statutes, precedents, and legal reasoning, assessing performance and reliability in legal question answering, case analysis, and decision recommendation \citep{guha2023legalbench, fei2024lawbench, dai2025law}.
These studies primarily adopt a normative or prescriptive perspective.
We instead take a descriptive perspective, modeling systematic patterns in LLM responsibility and blame judgments and evaluating their alignment with human judgments, without assuming that human judgments conform to universal principles of rationality.
Our benchmark further annotates five theory-derived attribution dimensions to examine the intermediate variables underlying these judgments.

\noindent \textbf{LLM Probing for Mechanism Exploration.} Probing measures the decodability of target variables from LLM hidden states using lightweight predictors \citep{belinkov2019analysis, belinkov2022probing}.
Studies examine factual and relational knowledge \citep{meng2022locating, hernandezlinearity}, instruction following \citep{heollms}, and truth-related signals \citep{orgadllms}, sometimes combining representation analysis with activation interventions to assess their influence on outputs.
Social and cognitive studies investigate modality reliance through prompt perturbations \citep{QI2023103510}, stereotypical patterns in emotion attribution \citep{plaza2024angry}, and latent value structures \citep{kang-etal-2025-values}.
More directly related to our methodology, \citet{tak-etal-2025-mechanistic} probe theory-derived cognitive appraisal dimensions and intervene on their representations to influence emotion inference.
Other studies examine the decodability and steerability of self- and personality-related constructs \citep{ju2025probing} and ideological positions \citep{zhang-2025-probing}.
These findings motivate theory-guided probing and interventions for LLM social reasoning; however, to the best of our knowledge, prior work has not specifically probed attribution dimensions underlying responsibility and blame judgments.

\section{Theoretical Foundation}

\noindent \textbf{Responsibility and blame judgments.}
Responsibility judgment assesses the extent to which an agent is accountable for an event, considering causal involvement, controllability, and mitigating factors such as coercion or lack of foreknowledge.
Blame judgment further assesses whether the agent deserves moral criticism, taking justifications and excuses into account.
Responsibility thus presupposes causality but is not reducible to it; similarly, responsibility is a prerequisite for blame but does not necessarily entail blameworthiness \citep{shaver1985, weiner1995judgments}.

\noindent \textbf{Attribution theories and dimensions.}
Classic attribution theories characterize responsibility and blame judgments through structured assessments of attribution dimensions, which capture agents' causal involvement, mental states, and situational constraints \citep{weiner1995judgments, shaver1985, malle2014PMoB}; see Appendix~\ref{app:classical_attribution_theories} for detailed accounts and illustrations.
Drawing on these theories, we focus on five dimensions (Appendix~\ref{app:operational_definitions}):
\begin{enumerate} 
    \item \textit{Intention.} Whether the agent deliberately performed the focal action according to a goal or plan.
    Here, intention concerns the action itself, not the outcome it produces.
    \item \textit{Voluntariness.} Whether the agent acted freely rather than under external coercion, authority, or situational pressure, even without an explicit threat.
    \item \textit{Foreknowledge.} Whether the agent knew or, given their epistemic and cognitive conditions, should have foreseen the outcome.
    Foreknowledge typically increases attributed responsibility and blame \citep{shaver1985}.
    \item \textit{Controllability.} Whether the agent had the practical capacity to prevent the outcome, including the necessary means, skills, or opportunity.
    \item \textit{Obligation.} Whether the agent had a duty to prevent the negative outcome given the situation and social relationship, a key consideration in attributing responsibility or blame for unintentional outcomes.
\end{enumerate}

\section{Methodology}
\label{sec:methodology}

\subsection{Data Construction}
\label{sec:data_c}

\begin{figure}[t]
    \centering
    \includegraphics[width=\linewidth]{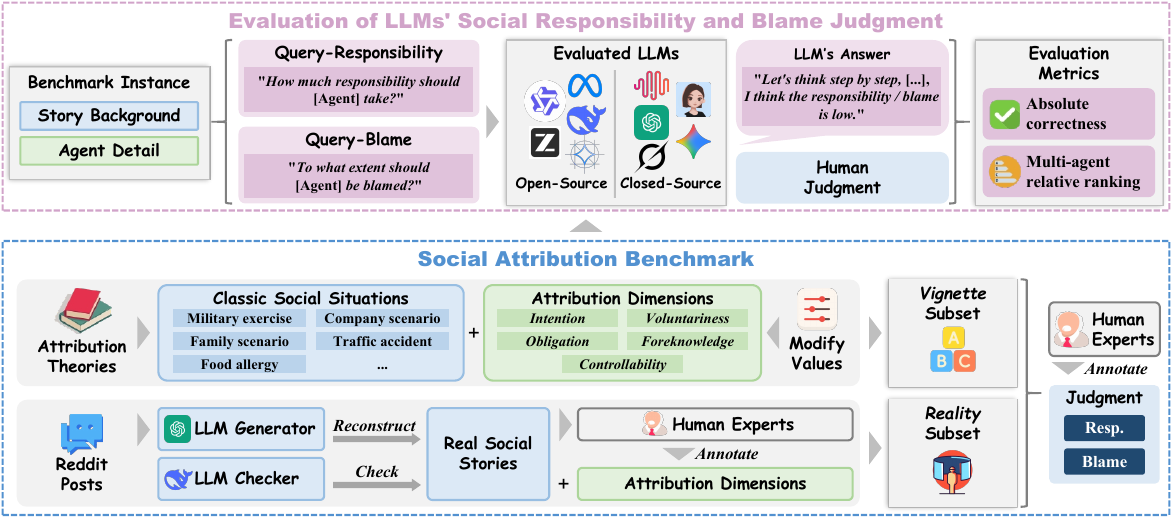}
    \caption{Benchmark construction and evaluation of LLM social attribution.}
    \label{fig:teaser_1}
\end{figure}

To address the lack of a dedicated data resource for systematically studying LLM social attribution, we first construct a benchmark that pairs human responsibility/blame judgments with theory-derived attribution dimensions.
The benchmark comprises two complementary subsets, \textit{Vignette} and \textit{Reality}, supporting both behavioral evaluation and analysis of the representations underlying these judgments.
Figure~\ref{fig:teaser_1} summarizes construction and evaluation.

The \textit{Vignette} subset draws on classic scenarios from psychological attribution research, preserving their theoretical grounding while using structured narratives to precisely control attribution dimensions.
Following the vignette paradigm \citep{baguley2022vignette_statistical}, we construct 11 templates spanning eight scenario types and three inter-agent relation categories: general interactions, vicarious blame, and command--obedience chains.
Dimension values constrained by each role structure are fixed, while the remaining values are systematically varied.
To reduce surface shortcuts, we diversify narrative details and phrasing, randomize names and genders, and avoid explicit dimension labels.
Each scenario involves multiple agents, with a separate responsibility question and a separate blame question for each agent.
Appendix~\ref{app:vignette_construction} details the construction.
The \textit{Reality} subset captures realistic and ambiguous social interactions through narratives posted by real users, complementing the controlled \textit{Vignette} settings.
Following the story-reconstruction approach of \citet{forbes2020socialnorm}, we select Reddit AITA posts with clear negative outcomes and reconstruct them into concise narratives.
LLM-assisted preprocessing and inspection remove irrelevant or harmful content and identifying information, replacing identities with fictional names.
For each scenario, we select a single target agent and ask separate responsibility and blame questions about that agent.
Appendices~\ref{app:reality_preprocessing} and~\ref{app:prompts_llm_annnotation} specify the models, processing steps, and prompts.
Both subsets are constructed in Chinese and English.

\par\smallskip
\noindent\fbox{%
\begin{minipage}{\dimexpr\linewidth-2\fboxsep-2\fboxrule\relax}
\textbf{Data Example.}
During an artillery exercise, a general orders a strike through a platoon leader and a soldier.
Both candidate buildings, A and B, are abandoned.
The general, unaware of workers cleaning inside A, allows the platoon leader to choose either building.
The platoon leader knows about the workers but orders a strike on A.
The soldier also knows about them.
The strike injures the workers.

\smallskip
\textbf{Question.}
To what extent should the general (platoon leader/soldier) be held responsible (be blamed)?
Choose from [high, medium, low, no].
\end{minipage}%
}
\par\smallskip

Each question $i$ concerns one agent, scenario, language, and task $m\in\{\textit{responsibility},\textit{blame}\}$, with judgment $a_{i,m}\in\mathcal{A} =\{\texttt{no},\texttt{low},\texttt{medium},\texttt{high}\}$.
The dimension set $\mathcal{K}$ comprises intention, voluntariness, foreknowledge, controllability, and obligation.
For each dimension $k\in\mathcal{K}$, $y_{i,k}\in\{0,1,\varnothing\}$ denotes a binary value or an unspecified/inapplicable label.
Dimension values are set during \textit{Vignette} construction and human-annotated for \textit{Reality}.
Two psychology experts independently annotate judgments, treating languages and tasks separately and viewing only scenarios and questions, with dimension labels hidden.
Before filtering, the annotators achieve 86.8\% exact agreement overall and a mean Cohen's $\kappa$ of 0.826 across the two tasks and languages, supporting the consistency of the annotations.
We retain only questions for which both annotators assign the same judgment label, yielding 7,639 questions.
See Appendices~\ref{app:annotation_details}, \ref{app:annotation_agreement}, \ref{app:data_statistics_of_filtering}, and~\ref{app:dataset_statistics} for annotation and dataset details.

To address RQ1, we use two metrics to quantify agreement between LLM and human responsibility and blame judgments.
Outputs are canonicalized to $\mathcal{A}$ (Appendix~\ref{app:answer_extraction}).
Accuracy is the proportion of valid predictions that exactly match human labels, measuring agreement on the absolute level of responsibility or blame assigned to an agent.
Mapping $(\texttt{no},\texttt{low},\texttt{medium},\texttt{high})$ to $(0,1,2,3)$ through $r$, we also compute Spearman correlation between predicted and human agent rankings within each \textit{Vignette} scenario, assigning average ranks to ties.
This metric measures agreement on which agents are more responsible or blameworthy relative to others, even when the predicted judgment levels differ from human labels.
We report the mean correlation across scenarios.
Invalid predictions are excluded from accuracy, and scenarios containing them are excluded from correlation (Appendix~\ref{app:evaluation_metrics}).

\subsection{Latent Space Dimension Probing}
\label{sec:latent_s_d}

\begin{figure}[t]
    \centering
    \includegraphics[width=\linewidth]{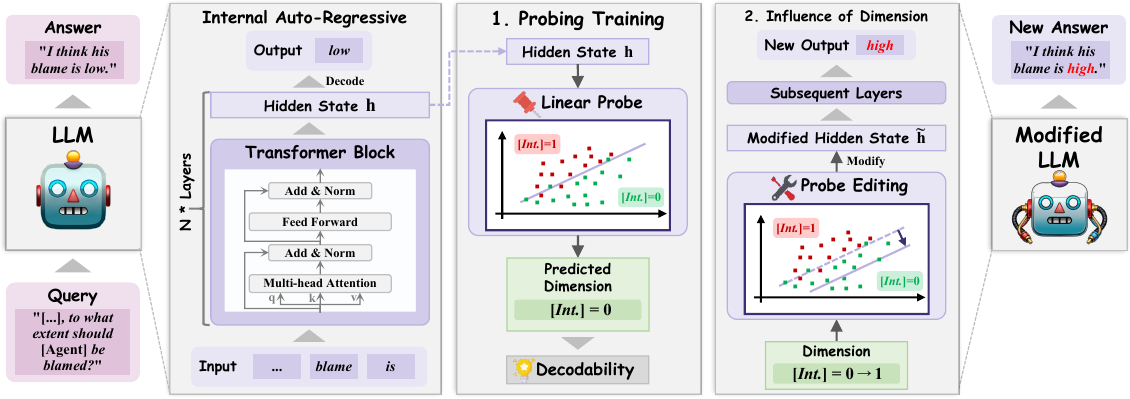}
    \caption{ Attribution-dimension probing through layerwise decodability analysis and directional interventions on judgments.
    }
    \label{fig:teaser_2}
\end{figure}

We examine whether attribution dimensions are decodable from LLM hidden states and whether interventions along their representations affect judgments in theory-predicted directions (Figure~\ref{fig:teaser_2}).

\subsubsection{Decodability of Attribution Dimensions}
\label{sec:decodability_o_a}
To address RQ2, we test whether theory-derived attribution dimensions can be linearly decoded from LLM hidden states.
For each prompt, we parse the frozen LLM's judgment and locate its label-token prediction step $\mathrm{Idx}_i^\star$.
At each probed layer $l$, we extract $z_i^{(l)}=h_{i,\mathrm{Idx}_i^\star}^{(l)}\in\mathbb{R}^d$, the state used to predict that token (Appendix~\ref{app:hidden_state_extraction}).
For each dimension and layer, a logistic probe estimates
\begin{equation}
p_k^{(l)}(y=1\mid z)
=
\sigma\!\left(w_k^{(l)\top}z+b_k^{(l)}\right),
\label{eq:dimension_probe}
\end{equation}
where $\sigma$ is the sigmoid.
We fit probes on training folds using class-balanced cross-entropy with $L_2$ regularization; training and evaluation require available hidden states and non-missing dimension labels (Appendix~\ref{app:probe_training}).

Probes are trained separately by language and subset to avoid train--test overlap between translations.
\textit{Vignette} uses eight-fold leave-one-scenario-type-out cross-validation: each fold tests one type and trains on the other seven, grouping all occurrences of that type across agent relations.
This tests decoding on unseen scenario types and reduces performance inflation from shared event backgrounds and templates.
\textit{Reality} uses 10-fold random cross-validation (Appendix~\ref{app:probe_splits}).
Within each language--subset run, we summarize \textit{decodability} by the unweighted mean of fold balanced accuracies, $(\mathrm{Recall}_0+\mathrm{Recall}_1)/2$.
Decodability measures recovery of binary dimension labels; benchmark accuracy concerns responsibility/blame judgments.
We report decodability as a percentage, with 50\% indicating chance.

\subsubsection{Influence of Attribution Dimensions on Judgment}
\label{sec:degree_o_d}

To address RQ3, we investigate whether the influence of latent attribution-dimension representations on LLMs' final responsibility and blame judgments is consistent with the patterns of human social attribution described by attribution theory \citep{shaver1985,mao2012modeling-JAIR,malle2014PMoB}.
To assess this consistency, we intervene along each dimension's representation and compare the resulting judgment changes with theory-derived directional expectations (Appendix~\ref{app:theoretical_positioning}).

First, for each training fold and selected layer $l\in\mathcal{L}_{\mathrm{sel}}$, let $W_{-k}^{(l)}$ collect the weights of other available dimension probes.
To reduce shared linear components, we residualize the target weight and normalize the residual:
\begin{equation}
\begin{aligned}
\hat v_k^{(l)}
&=
w_k^{(l)}
-
W_{-k}^{(l)}
\left(
W_{-k}^{(l)\top}W_{-k}^{(l)}+\gamma I
\right)^{-1}
W_{-k}^{(l)\top}w_k^{(l)},\\
v_k^{(l)}
&=
\frac{\hat v_k^{(l)}}{\|\hat v_k^{(l)}\|_2}.
\end{aligned}
\label{eq:dimension_direction}
\end{equation}
Here $\gamma>0$ regularizes residualization.
Directions are derived only from training-fold probes; degenerate residuals are skipped.
The positive direction increases probe evidence for $y_{i,k}=1$.

Second, the unedited output $\hat a_{i,m}$ fixes the feasible signs for all comparisons: $\mathcal{E}_i=\{+1\}$ for \texttt{no}, $\{-1\}$ for \texttt{high}, and $\{-1,+1\}$ otherwise.
A single-use hook edits its label-prediction step $\mathrm{Idx}_i^\star$ at layer $l$:
\begin{equation}
\tilde h_{i,\mathrm{Idx}_i^\star}^{(l)}
=
h_{i,\mathrm{Idx}_i^\star}^{(l)}
+\eta\alpha v_k^{(l)},
\qquad
\eta\in\mathcal{E}_i,\quad \alpha>0.
\label{eq:dimension_intervention}
\end{equation}

We then generate exactly one token and obtain comparison predictions by restricting the argmax to $\mathcal{A}$.
Besides the unedited baseline, we use two intervention controls: a random unit direction independent of labels and probe weights, and a direction obtained from a shuffled-label probe using the same direction-construction procedure.
Controls match the layer, prediction step, strength, sign, and perturbation norm, testing whether effects depend on the learned dimension direction.

In these negative-outcome scenarios, we hypothesize that strengthening an attribution-dimension representation tends to increase responsibility or blame, while weakening it tends to decrease the judgment.
For each setting $(m,l,k,\alpha)$, let $\hat a_{i,\eta}$ be the probe-directed prediction and $\hat a_{i,\eta}^{c}$ the comparison prediction, where $c\in\{\mathrm{base},\mathrm{rand},\mathrm{shuf}\}$ and $\hat a_{i,\eta}^{\mathrm{base}}=\hat a_{i,m}$.
Using $r$, we compute
\begin{equation}
\Delta_{i,\eta}^{c}
=
\eta\left[
r(\hat a_{i,\eta})-r(\hat a_{i,\eta}^{c})
\right].
\label{eq:directional_difference}
\end{equation}
Positive values indicate theory-aligned shifts relative to $c$: higher judgments under positive interventions or lower judgments under negative ones.

For $N$ held-out questions, define one binary outcome per question and comparison:
\begin{equation}
s_i^c=\mathbb{I}\!\left[\min_{\eta\in\mathcal{E}_i}\Delta_{i,\eta}^c>0\right],
\qquad S_c=\sum_{i=1}^{N}s_i^c.
\label{eq:joint_directional_success}
\end{equation}
Every feasible sign must strictly align; a tie or opposite shift on any tested sign counts as failure.
For success probability $\pi_c$, we test $H_0:\pi_c\leq1/2$ against $H_1:\pi_c>1/2$ using a one-sided exact binomial test:
\begin{equation}
p_c=\Pr[B\geq S_c]
=\sum_{j=S_c}^{N}\binom{N}{j}2^{-N},
\qquad
B\sim\mathrm{Binom}(N,1/2).
\label{eq:directional_test}
\end{equation}
We report ``Tests passed'' if $N>0$ and $p_c<0.01$ for all three comparisons, and ``Tests not passed'' otherwise.

%\subsection{Supplementary Experiments}
%\label{sec:supplementary_experiments}

%The Appendix provides additional analyses of evaluation reliability and the robustness of our probing results.
%These include valid-output rates and performance breakdowns by scenario type and attribution-dimension value.
%We assess decodability using shuffled-label, random-projection, and lexical controls (Appendix~\ref{app:probe_training}).
%We compare both decodability and intervention effects at the last input token with those at the label-token prediction step.
%We also examine sensitivity to intervention strength through the average signed change in responsibility and blame judgments (Appendix~\ref{app:intervention_strength}).

\section{Experiments}

\subsection{Experimental Setup}
\label{sec:experimental_setup}

We evaluate 32 LLMs: nine closed-source LLMs from the Claude, Doubao Seed, Gemini, GPT, Grok, Kimi, MiMo, and Step families, and 23 open-source LLMs from the Llama, Phi, GPT-oss, DeepSeek-R1, Qwen, Gemma, GLM, and InternLM families.
The open-source models span 0.5--120B parameters; Appendix~\ref{app:evaluated_models} provides the full model list and inference settings.
We evaluate each LLM with four prompt strategies: Direct (answer without reasoning), CoT (reason step by step), Self-reflection (review and revise an initial judgment), and Theory-guided (analyze the five attribution dimensions before judging).
For comparison, we evaluate five non-LLM models: Random Stratified (sample labels according to their frequencies), Majority Class (predict the most frequent label), Heuristic Rule (score textual keywords), Attribution-Theory Rule (apply theory-based rules to annotated dimensions), and mBERT Classifier (predict labels with fine-tuned multilingual BERT).
Performance is measured by accuracy and Spearman's $\rho$, as defined in Section~\ref{sec:methodology}.
Prompt templates and non-LLM model details are provided in Appendices~\ref{app:prompt_templates} and~\ref{app:non_llm_models}.

To understand how attribution dimensions are represented and influence LLM judgments, we examine their decodability in hidden states and whether their influences on judgments align with human attribution theory.
We use DeepSeek-R1-1.5B, Llama-3.2-1B, Llama-3.1-8B, Qwen3-4B, and Qwen3-8B as backbones with CoT prompts, following the methods in Section~\ref{sec:latent_s_d}.
Reported decodability plots recompute balanced accuracy from held-out predictions pooled across languages and subsets (Appendix~\ref{app:probe_splits}).
Hyperparameters, implementation, and testing details are provided in Appendices~\ref{app:intervention_implementation} and~\ref{app:directional_test}.

\begin{figure}[t]
    \centering
    \includegraphics[width=\linewidth]{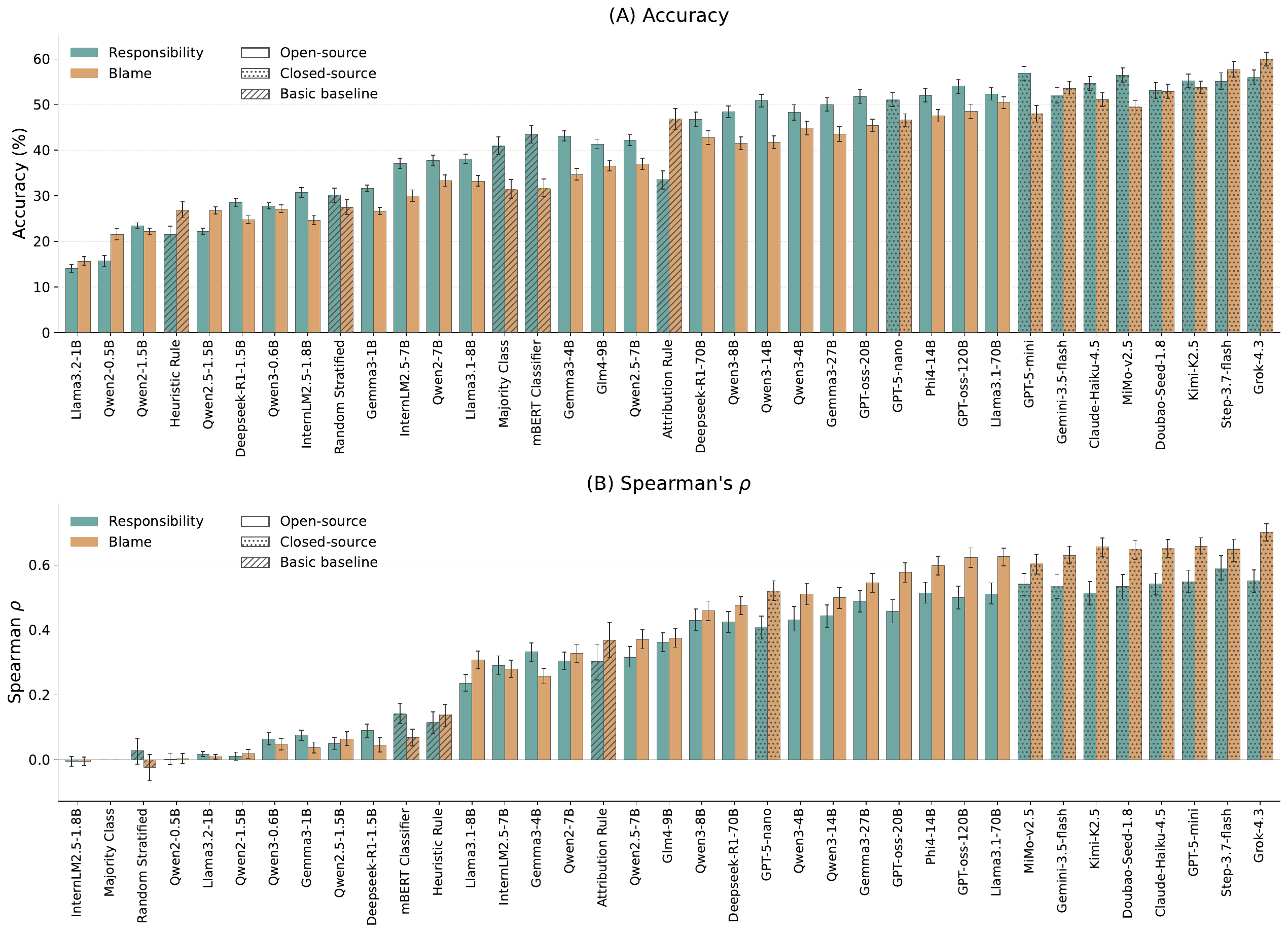}
    \caption{Accuracy and Spearman's $\rho$ across models, with 95\% confidence intervals.}
    \label{fig:overall_accuracy_spearman}
\end{figure}

\begin{figure}[t]
    \centering
    \includegraphics[width=\linewidth]{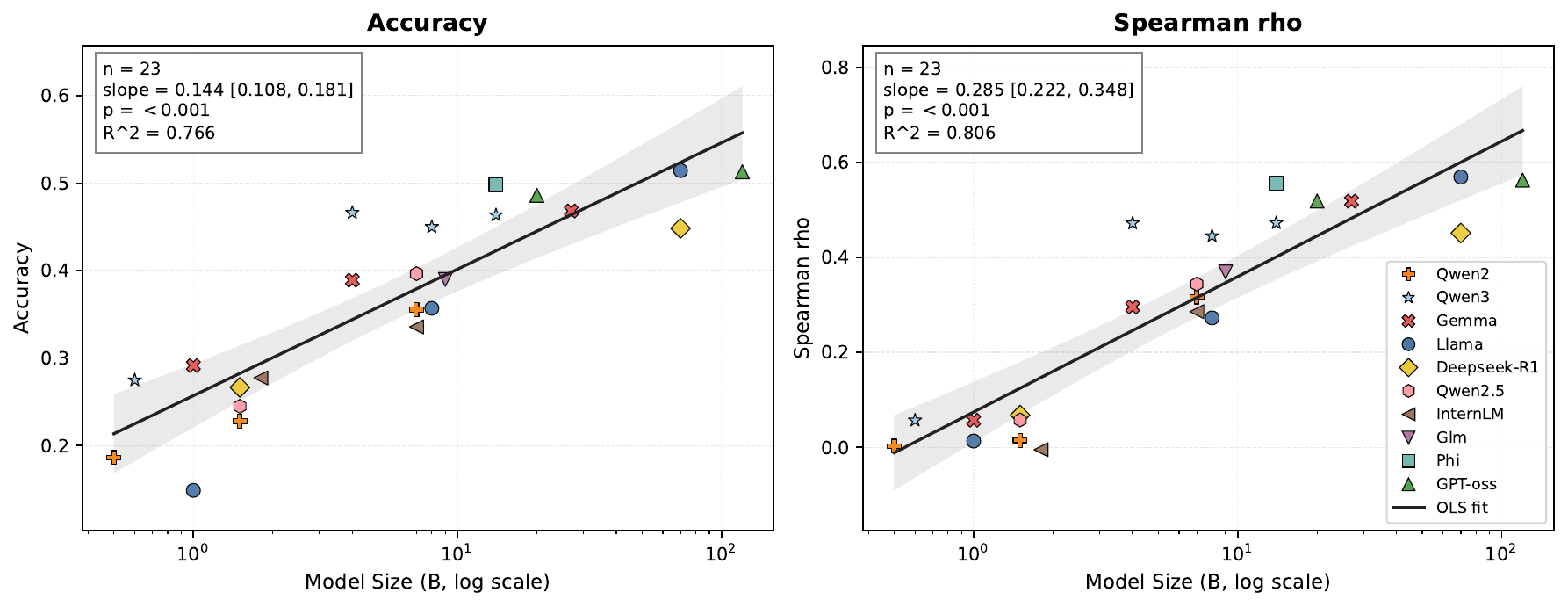}
    \caption{Model-size trends averaged over both tasks, with OLS fits and 95\% confidence bands.}
    \label{fig:model_size_trends}
\end{figure}

\begin{figure}[t]
    \centering
    \includegraphics[width=\linewidth]{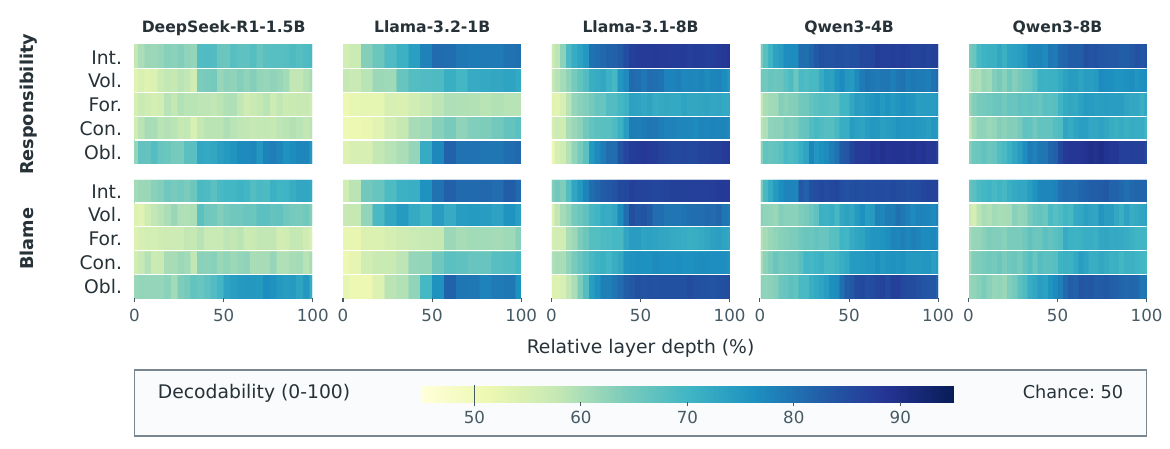}
    \caption{Layerwise dimension decodability. Panels separate models and tasks; rows denote dimensions, and darker colors indicate higher decodability (chance: 50\%).}
    \label{fig:dimension_decodability}
\end{figure}

\subsection{Performance Evaluation on LLM Social Attribution}
\label{sec:performance_evaluation}

Figure~\ref{fig:overall_accuracy_spearman} compares agreement with human judgments across models, averaging each LLM's results over the four prompting strategies.
Grok-4.3 achieves the highest scores on both metrics when averaged across responsibility and blame, with 58.00\% accuracy and $\rho=0.627$.

For most models, accuracy is higher on responsibility than on blame, whereas Spearman's $\rho$ shows the opposite pattern.
It is because assigning a blame level requires finer judgments of moral severity and mitigating circumstances: models may preserve relative differences in blame among agents while differing from humans on the precise level assigned.

The supervised mBERT classifier achieves accuracy comparable to selected 4--8B LLMs: 43.47\% on responsibility, versus 43.13\% for Gemma3-4B, and 31.60\% on blame, versus 33.23\% for Llama3.1-8B (Appendix Table~\ref{tab:basic_baselines}).
Its limited accuracy may reflect difficulty learning agent-specific roles and obligations from the available supervision.

Across the 23 open-source models, both accuracy and ranking agreement increase with log-scale parameter count (Figure~\ref{fig:model_size_trends}).
The fitted slopes are positive for both metrics ($p<0.001$), with $R^2=0.766$ for accuracy and $R^2=0.806$ for $\rho$.
However, differences in model families and training preclude attributing this trend to parameter count alone.

These results answer RQ1 by showing measurable but incomplete agreement with human judgments, with higher agreement associated with larger open-source models.

\subsection{Decodability of Attribution Dimensions in LLM Hidden States}
\label{sec:decodability_results}

Figure~\ref{fig:dimension_decodability} shows the decodability of attribution dimensions at the label-token prediction step across five models, both judgment tasks, and normalized layer depths.

Attribution dimensions are consistently decodable from hidden states, with most middle and later layers scoring above chance.
Intention and obligation show the highest decodability, approaching 90\% in Llama-3.1-8B and the Qwen models.
Foreknowledge is weaker, particularly in DeepSeek-R1-1.5B and Llama-3.2-1B, while voluntariness and controllability vary more across models.
Responsibility and blame show similar patterns in dimension strength and layerwise variation.

Decodability generally increases from early to middle layers and remains high across later layers.
Some dimensions plateau or decline slightly near the output, with peak decodability often occurring before the final layer.
This pattern is consistent with attribution information becoming easier to read out as the model integrates the scenario context.

Sanity checks further support these findings (see Appendix Figures~\ref{fig:app-step1-sanity-deepseek-r1-15b}--\ref{fig:app-step1-sanity-qwen3-8b}).
Across all five models, the original probes outperform the shuffled-label control, which stays near chance, and generally outperform the 64-dimensional random-projection control.
For Llama-3.1-8B, Qwen3-4B, and Qwen3-8B, they also exceed the scenario-text TF-IDF control across most layers.

For RQ2, these results provide evidence that the evaluated LLMs encode linearly decodable information about attribution dimensions, with strength varying across dimensions, models, and layers.

\subsection{Influence of Attribution Dimensions on Judgment}
\label{sec:influence_results}

\begin{figure}[t]
    \centering
    \includegraphics[width=\linewidth]{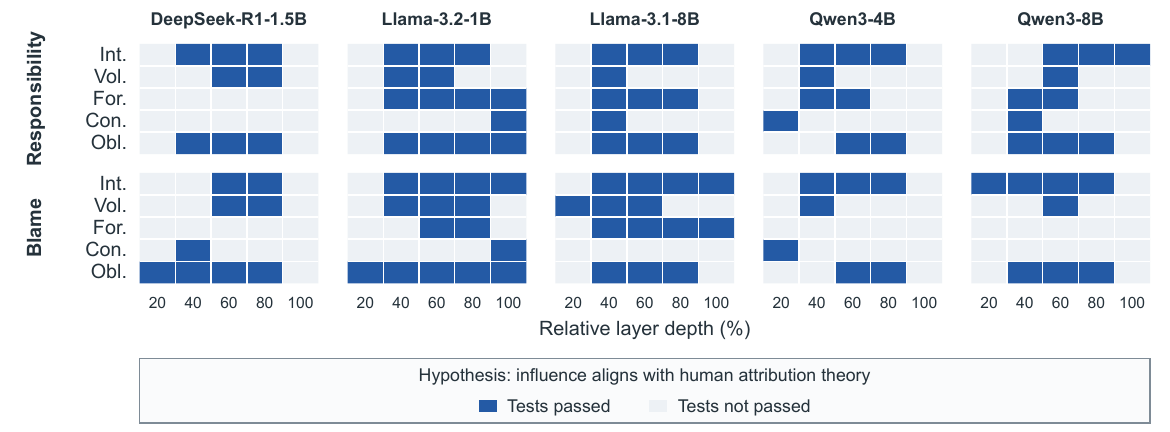}
    \caption{Layerwise influence of attribution dimensions. Blue cells pass all three question-level directional-success tests ($p<0.01$); gray cells do not. Rows denote dimensions; 100\% denotes the final layer.}
    \label{fig:dimension_influence}
\end{figure}

Figure~\ref{fig:dimension_influence} shows whether interventions along attribution-dimension representations shift responsibility and blame judgments in the directions predicted by attribution theory.
Intention and obligation show the most consistent effects: at 60\% and 80\% relative depth, both dimensions satisfy the directional tests in all five models and both tasks.
These effects hold against the unedited baseline and both intervention controls.
This pattern also matches the high decodability of intention and obligation observed in the preceding analysis.

Theory-aligned effects are concentrated in the middle and later layers, particularly between 40\% and 80\% relative depth.
The 60\% depth has the broadest support, with 33 of 50 settings passing the tests, compared with six at 20\% depth and nine at the final layer.
Llama-3.2-1B retains several significant effects at the final layer, while the other models show fewer.

The remaining dimensions show more variation across models and tasks.
Foreknowledge produces significant effects on both tasks in the Llama models, while the Qwen models show significant effects only on responsibility.
Voluntariness passes the tests in all five models at selected intermediate layers, including both tasks at 60\% and 80\% depth in DeepSeek-R1-1.5B.
Controllability has the fewest significant results, appearing at isolated layers.

For RQ3, the intervention results support theory-aligned influences most consistently for intention and obligation, while alignment for other dimensions depends on the model, task, and layer.

\subsection{Supplementary Experimental Results}
\label{sec:supplementary_results}

The appendices report full LLM and non-LLM results (\ref{app:full_llm_results}, \ref{app:full_baseline_results}), valid-output rates (\ref{app:valid_outputs}), scenario- and dimension-level breakdowns (\ref{app:error_breakdown}), decodability sanity checks (\ref{app:decodability_sanity}), earlier-token decodability and influence (\ref{app:early_token_decodability}, \ref{app:early_token_influence}), and intervention-strength sensitivity (\ref{app:intervention_strength}).

\section{Conclusion}
\label{sec:conclusion}

To understand social attribution as a fundamental aspect of social reasoning, we present the first systematic investigation of LLM responsibility and blame judgments.
Our benchmark comprises a \textit{Vignette} subset of classic attribution scenarios and a \textit{Reality} subset of real-world social narratives.
We evaluate 32 LLMs and develop a framework to probe the representations underlying their judgments.
Closed-source models generally outperform open-source models, whose agreement with human judgments is positively associated with model size.
Across five probed models, attribution dimensions are decodable from hidden states, especially intention and obligation.
Interventions on these dimensions produce the most consistent shifts predicted by attribution theory, especially in middle and later layers.
Effects of other dimensions vary across models, tasks, and layers.
Together, these findings connect LLM social attribution behavior with internal representations of factors identified in human attribution theory.

\bibliography{iclr2027_conference}
\bibliographystyle{iclr2027_conference}

\appendix
% Insert after the manuscript's existing \appendix command.
% Requires booktabs, tabularx, graphicx, and subcaption.
% Retain the existing appendices labeled
% app:prompts_llm_annnotation and app:annotation_details.

\input{appendix}

\end{document}

%% file: math_commands.tex
\usepackage{amsmath,amsfonts,bm}

\def\eqref#1{equation~\ref{#1}}
\def\1{\bm{1}}

\DeclareMathAlphabet{\mathsfit}{\encodingdefault}{\sfdefault}{m}{sl}
\SetMathAlphabet{\mathsfit}{bold}{\encodingdefault}{\sfdefault}{bx}{n}

%% file: old_intro.tex
\section{Introduction}
\label{sec:introduction}

In recent years, large language models (LLMs) have demonstrated strong capabilities in tasks ranging from natural language understanding and generation, interactive conversations, to complex problem solving \citep{achiam2023gpt}. With the continuous progress of advanced AI technologies, LLMs have shown broader impacts in the social realm and been integrated into sociotechnical applications, such as social simulation \citep{park2023gsocial_simulate}, human-AI interactions \citep{hadan2025responsible}, social learning \citep{goindani2020social}, and normative reasoning \citep{yuan2024socialnorm}. In such settings, inaccurate responsibility or blame attribution can lead to concrete harms, for example, in legal and social service assistance, incorrect attribution may mislead responsibility judgments; in mental health and healthcare settings, incorrect blame may lead to victim-blaming or unsafe advice \citep{dahl2024large_legal}; and in content moderation or community governance, incorrect blame may cause victims to be treated as violators \citep{moore2025expressing}. A critical topic underlying these diverse applications is the reliance on \textit{social attribution} \citep{gratch2006modeling, malle2022attribution}, the social reasoning process that centers on attributing the occurrence of external events to the causes and reasons of agents' social behaviors, and assigning responsibility, blame, or credit to the involved agents in rich social interactions. Such social inference reflects how agents make sense of and act on the social world around them, and lies at the heart of social intelligence.

Social attribution is a fundamental field of research in social psychology and social cognition, and has been well-studied. For decades, researchers have established \textit{Attribution Theory} \citep{shaver1985, weiner1995judgments, malle2014PMoB} as the dominant theoretical framework. In these theoretical models, the central theme involves the judgments of social cause, responsibility and blame/credit. Such judgments are not driven by outcomes alone, but systematically shaped by a set of key attributional dimensions, such as \textit{intention}, \textit{voluntariness}, \textit{foreknowledge}, \textit{controllability}, and \textit{obligation}, that characterize agents' mental states and the nature of their involvement in social events. 
In computational research on social attribution, previous work has built computational frameworks to model this judgment process grounded in attribution theory \citep{mao2011IEEE-IS, mao2012modeling-JAIR}. More recently, several works propose to leverage LLMs primarily for normative or legal reasoning \citep{yuan2024socialnorm, guha2023legalbench}. Despite the importance of social attribution in theoretical foundations and broader applications, it is still underexplored in the computing/information science community, and especially in LLM social reasoning. 

In this paper, we focus on responsibility and blame attributions, examining current LLMs' performances in making responsibility and blame judgments, as well as their internal mechanisms underlying such judgments. 
Specifically, to explore current LLMs' performances in making responsibility and blame judgments, we raise the first research question: \textit{How accurate are current LLMs in attributing responsibility and blame compared to human judgments?} Furthermore, given the judgment process and the intermediate cognitive variables (i.e., attribution dimensions) clearly identified by attribution theory, it remains unclear whether a similar process exists in LLM social reasoning and to what extent LLM social attribution resembles that of human judgments in the internal social and cognitive mechanisms described in attribution theory. Therefore, we raise deeper research questions: \textit{Are social attribution dimensions identified in human theory actually represented in LLMs' internal hidden states?} and further: \textit{Do the influences of LLMs' dimension representations on judgments align with human social attribution theory?}

To answer these research questions, we conduct one of the first systematic studies of social attribution in LLM reasoning. Guided by attribution theory, we construct a bilingual social attribution benchmark consisting of two complementary subsets. One subset is built  from classic scenarios in attribution theory research based on vignette paradigm, enabling controlled manipulation of attribution dimensions across different inter-agent relations and social situations. The other subset is constructed from real-world social narratives to introduce more realistic and ambiguous social attribution scenarios. After expert annotation and filtering, the benchmark contains 7,639 responsibility/blame judgment questions.
On this basis, we evaluate 32 representative LLMs, including 9 closed-source and 23 open-source models, and compare them with 5 basic non-LLM baselines. This evaluation addresses RQ1 by measuring both exact agreement with human responsibility/blame labels and relative agreement in multi-agent attribution rankings. 

To answer the second and third research questions of investigating the internal mechanisms in LLM social attribution, we leverage a dimension-level probing framework to examine the existence of dimension representations in LLM hidden states and their influences on the final judgments. In mechanistic interpretability for LLMs, a rich toolbox of well-established methods has been developed \citep{huben2023sparse_sae, zhang2025personalized_pca, zhang-2025-probing}. Among them, the \textit{probing} technique is typically adopted in studying dimension-based cognitive or psychological representations within LLMs, mainly because it enables a quantitative alignment between theory-derived dimensional variables and LLM hidden states \citep{tak-etal-2025-mechanistic, kang-etal-2025-values, zhang-2025-probing}. In our probing-based framework, to examine the existence of dimension representations, we first train linear probes on the hidden states, leveraging the annotations of the key attribution dimensions from our constructed benchmark. We then conduct an empirical analysis across layers and dimensions to investigate the representation strength of each dimension at different latent regions. To further examine the influences of dimension representations on responsibility and blame judgments, we directionally vary the LLM hidden states along specific dimension representations to quantify the extent to which the final judgments are affected, and conduct rigorous hypothesis testing to verify the consistency of these impacts with the human judgment patterns indicated in attribution theory.

Our research contributions and findings are summarized as follows:

\begin{enumerate}
    \item We focus on the fundamental research problem of LLM social attribution with wide applications, and present one of the first systematic studies on LLM responsibility and blame judgments as well as the internal mechanisms underlying such judgments.
    \item We construct a bilingual social attribution benchmark consisting of a controlled vignette subset and a realistic social scenario subset, and conduct a comprehensive empirical evaluation on 32 representative LLMs and 5 classical baselines.
    \item We develop a latent space dimension probing methodology that employs a probe-based technique to quantify the decodability of attribution dimensions and dimension-level influences, and on this basis, investigate the existence of dimensional representations in LLM hidden states and their influence mechanisms.
    \item Our empirical findings demonstrate that specific attribution dimensions are represented within the latent space of LLMs, and their influences on LLM final judgments align with those of human social attribution theory.
\end{enumerate}

%% file: appendix.tex
% Appendix content only.
\clearpage
\begingroup
\raggedbottom
\renewcommand{\topfraction}{0.95}
\renewcommand{\bottomfraction}{0.85}
\renewcommand{\textfraction}{0.07}
\renewcommand{\floatpagefraction}{0.65}
\input{appendix_sections/theoretical_background}

\clearpage
\input{appendix_sections/dataset_construction}

\clearpage
\input{appendix_sections/benchmark_protocol}

\clearpage
\input{appendix_sections/probing_details}

\clearpage
\input{appendix_sections/benchmark_results}

\clearpage
\input{appendix_sections/decodability_results}

\clearpage
\input{appendix_sections/intervention_results}

\clearpage
\input{appendix_sections/discussion_limitations}
\clearpage
\endgroup

%% file: appendix_sections/theoretical_background.tex
\section{Theoretical Background and Operationalization}
\label{app:attribution_theories}

\subsection{Classical Attribution Theories}
\label{app:classical_attribution_theories}

Attribution theories describe how perceivers connect an agent's actions to an outcome and assess the agent's responsibility and blameworthiness. These judgments are related but distinct: causal involvement does not by itself establish responsibility, and responsibility does not by itself establish blame. Assessments of control, mental states, coercion, and mitigating circumstances help explain why agents involved in the same negative event can receive different judgments \citep{shaver1985,weiner1995judgments}. This distinction motivates our separate responsibility and blame tasks, as well as annotations of intermediate attribution dimensions rather than final judgments alone.

\begin{figure}[!htbp]
    \centering
    \begin{subfigure}[t]{0.32\linewidth}
        \centering
        \includegraphics[width=\linewidth]{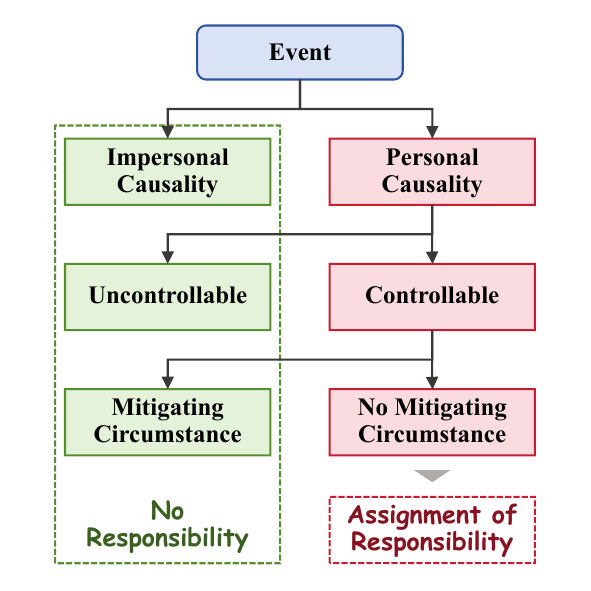}
        \caption{Weiner's responsibility model.}
        \label{fig:weiner}
    \end{subfigure}
    \hfill
    \begin{subfigure}[t]{0.32\linewidth}
        \centering
        \includegraphics[width=\linewidth]{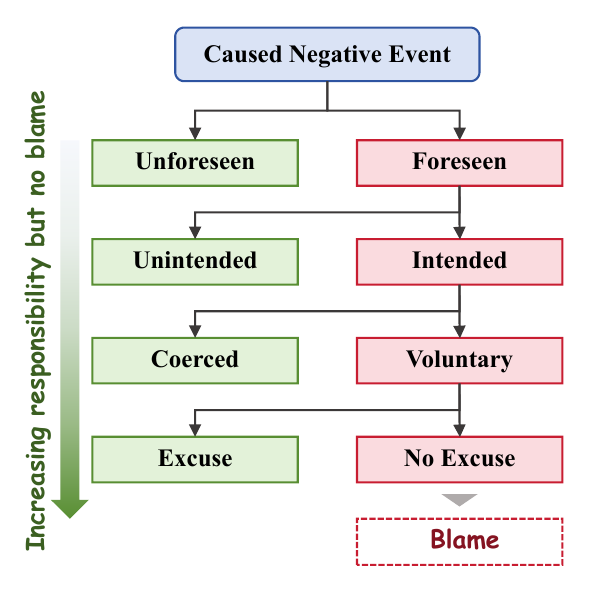}
        \caption{Shaver's blame model.}
        \label{fig:shaver}
    \end{subfigure}
    \hfill
    \begin{subfigure}[t]{0.32\linewidth}
        \centering
        \includegraphics[width=\linewidth]{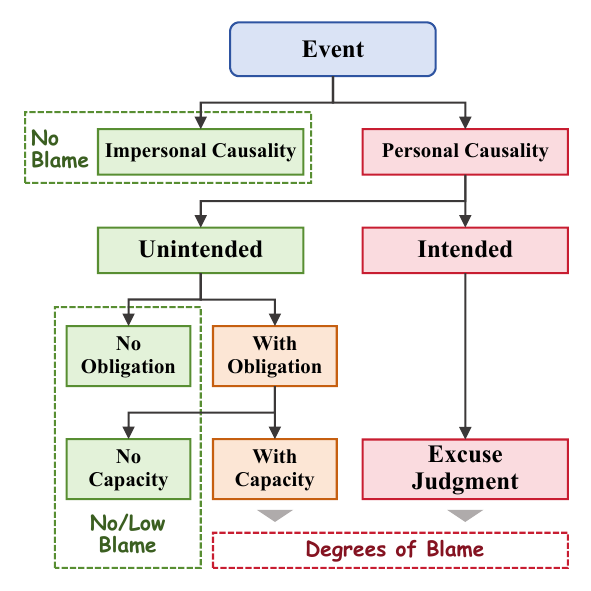}
        \caption{Malle et al.'s path model.}
        \label{fig:pmob}
    \end{subfigure}
    \caption{Classic attribution theories organize responsibility and blame judgments around assessments of causality, mental states, control, and mitigating circumstances \citep{weiner1995judgments,shaver1985,malle2014PMoB}. The models provide complementary theoretical foundations for the dimensions investigated here; they are not assumed to describe the computational architecture of an LLM.}
    \label{fig:attribution_theories_triptych}
\end{figure}

\paragraph{Weiner's model of responsibility judgment.}
Weiner characterizes causal explanations through separable dimensions: the \emph{locus of causality}, distinguishing internal from external causes; \emph{stability}, describing whether a cause persists over time; and \emph{controllability}, concerning whether the agent could have acted otherwise \citep{weiner1985attributional,weiner1986}. In the responsibility model (Figure~\ref{fig:weiner}), perceivers first assess whether an outcome reflects personal or situational causality. When personal causality is implicated, they evaluate controllability and consider mitigating circumstances before assigning responsibility \citep{weiner1995judgments}. The model therefore distinguishes an agent's causal contribution from the extent to which the agent can reasonably be held responsible for it.

\paragraph{Shaver's model of blame assignment.}
Shaver describes blame for a negative consequence as a structured assessment of responsibility-relevant variables \citep{shaver1985,shaver1986causality}. As illustrated in Figure~\ref{fig:shaver}, the assessment begins by distinguishing human agency from impersonal causes. When human agency is involved, subsequent assessments concern whether the agent foresaw the outcome, intended it, and acted voluntarily rather than under coercion. Justifications and excuses can further mitigate blame. These distinctions explain why similar outcomes can support different judgments depending on the agent's knowledge, intentions, and freedom of action. They also motivate treating responsibility and blame as separate targets: mitigating information can reduce blame even when responsibility remains.

\paragraph{Malle et al.'s Path Model of Blame.}
The Path Model of Blame (PMoB) specifies when different information becomes relevant to blame judgments \citep{malle2014PMoB}. After assessing causality, perceivers evaluate intentionality, which separates two subsequent paths (Figure~\ref{fig:pmob}). For intentional conduct, reasons and their justificatory force are central. For unintentional conduct, preventability becomes central, including whether the agent had an obligation and the capacity to prevent the outcome. New information can revise an earlier assessment and redirect the judgment along the other path. This conditional structure is particularly relevant to our inclusion of supervisory obligations and command--obedience relations: a social role can affect responsibility and blame even when the agent did not directly intend the negative outcome.

\subsection{Operational Definitions of Attribution Dimensions}
\label{app:operational_definitions}

We select five dimensions that are recurrent in attribution accounts and can be represented explicitly in controlled scenarios or annotated in reconstructed narratives. The operationalizations support comparisons across scenarios while preserving distinctions that would be obscured by a single overall judgment. They do not reproduce every construct or dependency in any one theory. In particular, causality provides part of the scenario context, whereas the five annotated dimensions describe mental states, constraints, capacities, and obligations associated with a target agent.

\paragraph{Intention: deliberate action rather than intended outcome.}
We operationalize intention at the level of the focal action: whether the agent deliberately performed it according to a goal or plan. This is \emph{action-intention}, which need not imply a purpose to bring about the eventual negative outcome. Some attribution accounts use intention in the latter, outcome-directed sense \citep{shaver1985,shaver1986causality,malle2014PMoB}. Maintaining the distinction allows the benchmark to vary deliberate action and foreknowledge separately. Their conjunction represents acting while foreseeing a consequence, without equating foresight with a desire to produce that consequence \citep{mao2011IEEE-IS}.

\paragraph{Voluntariness: constraints on behavioral choice.}
Voluntariness concerns whether the agent's choice was substantially constrained by coercion, authority, or situational pressure. An explicit threat is not required: compliance with an authority can also constrain the choice. This dimension therefore separates freely chosen actions from actions performed under orders or pressure \citep{shaver1985,mao2011IEEE-IS}. It is distinct from practical controllability. Following the distinction used in \citet{mao2011IEEE-IS}, information about control remains relevant when voluntariness is absent, because constrained choice alone does not specify which alternatives were practically available to the agent.

\paragraph{Foreknowledge: epistemic access to the consequence.}
Foreknowledge concerns whether the agent knew that the action could lead to the eventual outcome or, given the agent's epistemic and cognitive conditions, should have been able to foresee it. The assessment thus depends on information available at the time of acting, rather than knowledge acquired after the event. Foreknowledge can increase attributed responsibility and blame even when the negative outcome was not the agent's goal \citep{shaver1985}.

\paragraph{Controllability: practical capacity to prevent the outcome.}
We restrict controllability to the agent's physical or practical ability to prevent the outcome, including relevant means, skills, and opportunities. In PMoB, preventability can also involve the capacity to foresee the outcome \citep{malle2014PMoB}. We represent that epistemic component through foreknowledge, following the analytical separation of knowledge and control in our benchmark. This choice allows probing to distinguish knowing about a risk from having the practical capacity to avert it.

\paragraph{Obligation: a duty to prevent harm.}
Obligation concerns a duty of care, supervision, or prevention arising from the social situation and the agent's role or relationship. It is particularly relevant to unintentional outcomes and omissions, where an agent may bear responsibility or blame despite not performing the action that directly produced the harm \citep{malle2014PMoB}. Unlike broader uses of obligation that also include a subordinate's duty to obey a superior \citep{mao2011IEEE-IS}, our operationalization concerns the duty to prevent the negative outcome. This boundary keeps supervisory responsibility distinct from obedience within a command relationship.

The binary values indicate the presence or absence of each operationalized factor; $\varnothing$ denotes an unspecified or inapplicable value, as defined in the main text. The definitions guide the construction of \textit{Vignette} scenarios and the annotation of \textit{Reality} narratives. The annotation materials and instructions are provided in Appendix~\ref{app:annotation_details}.

\subsection{Theoretical Positioning and Directional Expectations}
\label{app:theoretical_positioning}

Human attribution theory supplies a principled reference for selecting variables, constructing data, and specifying hypotheses about LLM judgments. It does not require assuming that LLMs instantiate human cognitive processes. This distinction is central to the study design: the benchmark measures agreement with human judgments, the probes measure whether theory-derived dimensions can be recovered from hidden states, and the interventions examine whether directions associated with those dimensions affect outputs in the expected way. Together, these analyses connect behavioral agreement to internal representations while allowing the model's patterns to differ from the theoretical reference.

In the negative-outcome scenarios considered here, intention, voluntariness, foreknowledge, practical control, and obligation provide reasons for assigning greater responsibility or blame when other relevant circumstances are comparable \citep{shaver1985,weiner1995judgments,mao2012modeling-JAIR,malle2014PMoB}. We use this shared directional pattern to formulate an empirical expectation: strengthening the representation associated with a factor tends to increase the judgment, whereas weakening it tends to decrease the judgment. The positive probe direction corresponds to greater evidence for the factor's presence under our operationalization.

These expectations are conditional tendencies, not universal additive rules. The classic models distinguish intentional and unintentional conduct and allow justifications, excuses, and social relationships to change the relevance of a factor. Deliberately performing an action, for example, does not by itself establish an intention to cause harm or settle blameworthiness. Our directional tests therefore assess aggregate consistency with the stated hypothesis within the benchmark and intervention setting; they do not require every individual scenario to follow the same monotonic relation. The inferential limits of this evidence are discussed in Appendix~\ref{app:interpretation_limits}.

%% file: appendix_sections/dataset_construction.tex
\section{Dataset Construction and Annotation}
\label{app:data_construction}

The two subsets serve complementary purposes.
\textit{Vignette} provides controlled scenarios in which attribution
factors can be varied while preserving a coherent social context.
\textit{Reality} adds reconstructed everyday conflicts with less
explicit and more ambiguous attribution information.
Together, they support both evaluation against human judgments and
probing of the dimensions underlying those judgments.
This section details how the narratives, dimension labels, and
judgment labels are obtained, and how agreement-based filtering
changes the resulting sample.

\subsection{Vignette Construction}
\label{app:vignette_construction}

The vignette method uses short hypothetical situations to vary
explanatory factors systematically while retaining contextual detail
\citep{finch1987vignette,aguinis2014best_vignette,baguley2022vignette_statistical}.
This makes it useful for relating LLM judgments to the factors
studied in human social attribution.
We construct scenarios around negative outcomes using canonical
situations from attribution research
\citep{mao2011IEEE-IS,mao2012modeling-JAIR,malle2014PMoB}.
Each template specifies agents, their relations, and attribution
dimensions constrained by their roles.
An \emph{inter-agent relation} describes how agents are socially connected; a \emph{scenario type} describes the event setting.
The three relation categories are:
\begin{itemize}
    \item \textbf{General}: agents have no predefined social ties,
    or their ties do not entail responsibility or blame for
    another agent's conduct.
    \item \textbf{Vicarious blame}: agents occupy asymmetric social
    roles, such as parent--child or advisor--student, in which one
    agent has a supervisory obligation and may be held responsible
    for another agent's behavior \citep{malle2014PMoB}.
    \item \textbf{Commanding chain}: agents are linked by
    command--obedience relations
    \citep{mao2011IEEE-IS,mao2012modeling-JAIR}.
\end{itemize}

We generate instances by varying the remaining dimension values
subject to these role constraints.
For example, a parent--child relation introduces a supervisory
obligation for the parent without requiring the parent and child
to share the same intention.
The template therefore constrains which combinations are meaningful;
it does not require all five dimensions to vary independently for
every agent.
Table~\ref{tab:scenarios_details_across_three_relation_types} lists
the 11 relation-specific templates, their sources, event descriptions, fixed and
variable dimensions, agent counts, and instance counts.
These templates cover eight scenario types: Traffic Accident, Food Allergy,
Company, Drowning, Laboratory, Family, Gunman, and Military Strike.
Company occurs in all three relation categories, and Laboratory occurs in
General and Vicarious Blame; each remains one scenario type across relations.
These eight types define the held-out groups in the probing evaluation
(Appendix~\ref{app:probe_splits}).

To reduce the possibility that later probes exploit recurring names
or sentence patterns instead of attribution-related information,
we manually diversify the semantic details, syntactic structures,
and narrative phrasing of the instances.
Character names and genders are randomly generated for each instance.
The descriptions also avoid trivially revealing the attribution
variables through literal labels.
These construction choices reduce repeated surface cues;
the decodability controls separately examine how much information
can still be recovered from surface text
(Appendix~\ref{app:probe_training}).

A \emph{scenario instance} is one completed multi-agent narrative;
an \emph{agent--scenario pair} selects one of its agents;
a \emph{question} additionally specifies a judgment task and language.
Construction yields 320 general, 80 vicarious-blame, and 72
commanding-chain scenario instances per language.
Querying every agent produces
\[
320\times3+80\times2+72\times3=1{,}336
\]
agent--scenario pairs per language.
Scenarios are first constructed in Chinese and translated into
English item by item.
Each pair is instantiated into separate responsibility and blame
questions; the resulting pre-filtering question counts are given
in Appendix~\ref{app:annotation_details}.

\begin{table}[!htbp]
\centering
\caption{The 11 Vignette templates span eight scenario types and three inter-agent relation categories. Construction settings precede
annotation filtering. ``Agents Number'' gives the number of agents
in each scenario, and ``\#Instances'' gives the number of scenario
instances per language. Fixed and variable dimensions are subject
to agent-specific role constraints.}
\label{tab:scenarios_details_across_three_relation_types}
\includegraphics[width=\linewidth]{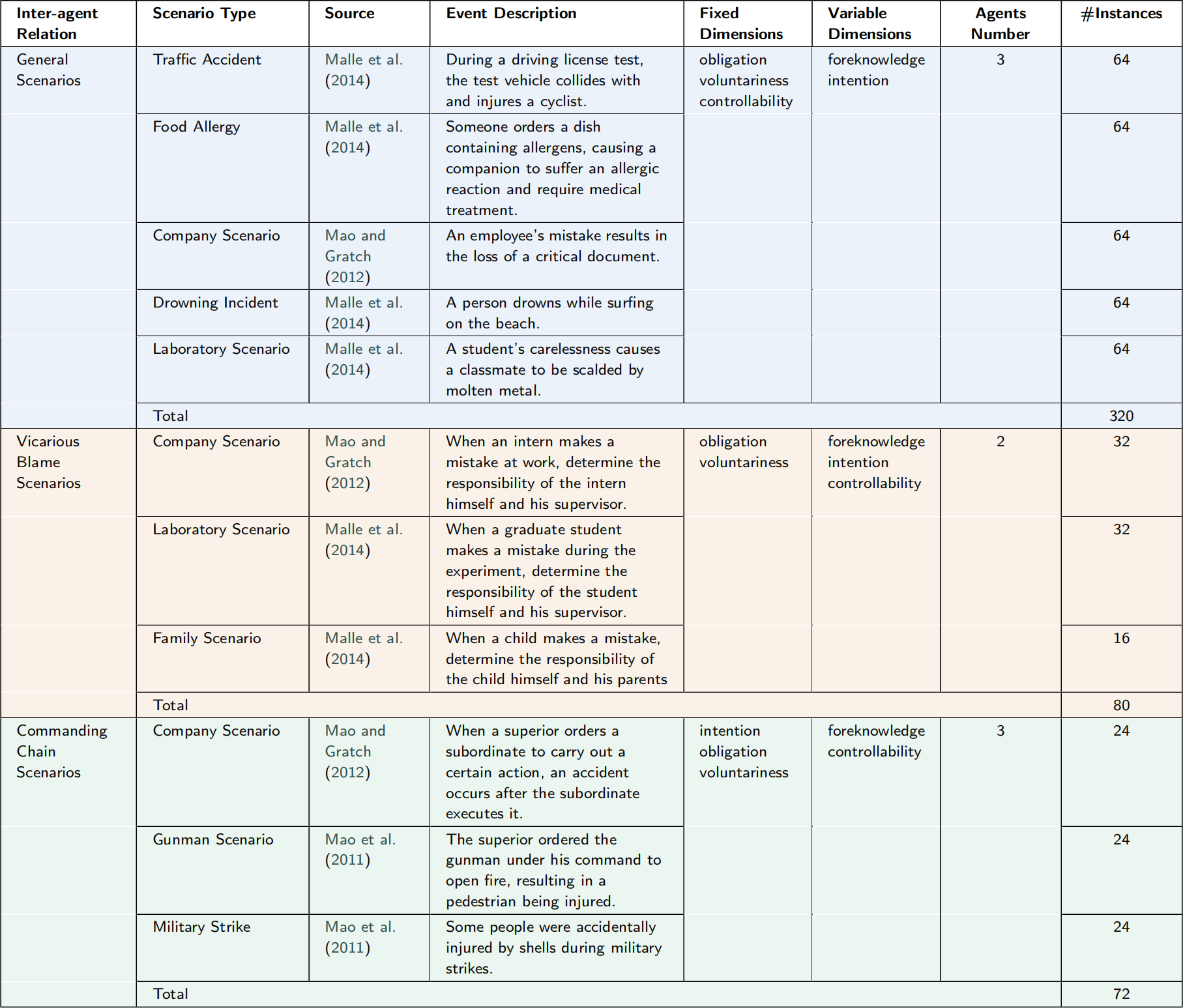}
\end{table}

\subsection{Reality Preprocessing}
\label{app:reality_preprocessing}

Controlled vignettes make the attribution factors explicit by design,
but cannot capture all the ambiguity of everyday social narratives.
The \textit{Reality} subset complements them with interpersonal
conflicts drawn from Reddit's AITA community,\footnote{
\url{https://www.reddit.com/r/AmItheAsshole/}}
where users describe conflicts and ask for judgments about the
parties involved.
These narratives provide agents, actions, constraints, and outcomes
in naturally occurring social contexts.

We select posts with clear negative outcomes and apply story
reconstruction \citep{forbes2020socialnorm} to preserve their core
events while removing irrelevant slang, dialogue, harmful or biased
content, personal information, and metadata.
Real identities are replaced with fictional names.
The inclusion criteria require a self-contained narrative with at
least two agents whose actions, roles, or knowledge are relevant to
the negative outcome.
For each reconstructed scenario, a single target agent is selected
for separate responsibility and blame questions.

ChatGPT-5.4-mini\footnote{
\url{https://developers.openai.com/api/docs/models/gpt-5.4-mini}}
performs cleaning and preprocessing.
DeepSeek-V4-Pro\footnote{
\url{https://huggingface.co/deepseek-ai/DeepSeek-V4-Pro}}
then checks the reconstructed narratives and filters those
containing residual unwanted content.
The prompts are provided in
Appendix~\ref{app:prompts_llm_annnotation}.
English instances are translated into Chinese, yielding
864 agent--scenario pairs in each language before judgment annotation.
This narrative-quality filtering precedes the separate
judgment-agreement filtering in
Appendix~\ref{app:data_statistics_of_filtering}.

\subsection{Prompts for LLM-Assisted Data Construction}
\label{app:prompts_llm_annnotation}

LLMs assist with reconstruction, cleaning, and quality inspection;
they do not supply the final responsibility or blame labels.
Those judgments are assigned by human experts.
The reconstruction prompt takes an AITA post as input and returns
the scenario background and agent descriptions.
The inspection prompt specifies retention criteria and returns a
decision with its reason.
Both steps require valid JSON without additional prose.

\paragraph{Prompt for story reconstruction and cleaning.}
We used the following prompt to transform raw Reddit posts into concise benchmark-style social scenarios.

\begin{quote}
\small
You are helping construct a dataset for social attribution research. Given a Reddit AITA post, rewrite it as a concise third-person story.

Requirements:
(1) Keep only the core plot related to the interpersonal conflict and the negative outcome.
(2) Preserve the main agents, their actions, relevant knowledge, constraints, and consequences.
(3) Remove irrelevant metadata, voting information, comments, edits, usernames, hyperlinks, and platform-specific expressions.
(4) Remove slang, offensive wording, harmful or biased expressions, and private personal information.
(5) Replace real or specific personal names with fictional names.
(6) Do not add new facts, motives, or outcomes that are not supported by the original post.
(7) Write the output in a neutral and clear style.

Return only a valid JSON object. Do not include markdown, comments, explanations, or any text outside the JSON object.

The JSON object must contain the following fields:
\begin{itemize}
    \item \texttt{story\_background}: a concise third-person description of the core event and the negative outcome.
    \item \texttt{character\_details}: a list of short descriptions of the main agents. Each item should describe one agent's relevant actions, knowledge, constraints, or role.
\end{itemize}
\end{quote}

\paragraph{Prompt for quality inspection and filtering.}
We used the following prompt to check whether the reconstructed stories satisfied the requirements for inclusion in the \textit{Reality} subset.

\begin{quote}
\small
You are checking reconstructed social scenarios for a benchmark on responsibility and blame judgment. Given a reconstructed story, determine whether it should be retained.

A story should be retained only if:
(1) it contains a clear negative outcome;
(2) it contains at least two agents involved in the event;
(3) the actions, roles, or knowledge of the agents are relevant to responsibility or blame judgment;
(4) the story is understandable without the original Reddit post;
(5) the story does not contain private personal information, platform metadata, irrelevant comments, or harmful and biased expressions;
(6) the story does not introduce unsupported facts beyond the original narrative.

Return only a valid JSON object. Do not include markdown, comments, explanations, or any text outside the JSON object.

The JSON object must contain the following fields:
\begin{itemize}
    \item \texttt{decision}: either \texttt{Retain} or \texttt{Filter}.
    \item \texttt{reason}: a brief reason for the decision. If the decision is \texttt{Retain}, write a short confirmation. If the decision is \texttt{Filter}, state the main problem.
\end{itemize}
\end{quote}

\subsection{Human Annotation Protocol and Guidelines}
\label{app:annotation_details}

\paragraph{Annotation tasks and units.}
The annotation procedure has two stages.
For \textit{Reality}, human annotators assign values to the five
attribution dimensions for the target agent in each reconstructed
story: intention, voluntariness, foreknowledge, controllability,
and obligation.
Unlike \textit{Vignette}, these narratives have no pre-designed
dimension settings.
For both subsets, annotators then assign the final responsibility
and blame judgments.
Thus, \textit{Vignette} dimension values come from construction,
whereas its judgments are human-annotated;
\textit{Reality} has human annotations for both.

A judgment-annotation unit is one agent--scenario pair paired with
one task in one language.
Before judgment annotation and agreement filtering,
\textit{Vignette} contains 1,336 pairs per language and
\textit{Reality} contains 864.
Instantiating both languages and tasks yields
$1{,}336\times2\times2=5{,}344$ \textit{Vignette} questions
and $864\times2\times2=3{,}456$ \textit{Reality} questions.
Chinese and English versions are treated as separate samples,
and responsibility and blame are annotated independently using
\texttt{no}, \texttt{low}, \texttt{medium}, and \texttt{high}.
Separating the tasks allows the reference judgments to reflect
the distinction between bearing responsibility for an outcome
and deserving blame for it.

\paragraph{Annotators and blinding.}
We recruit two social-psychology experts of different genders,
both fluent in English and Chinese.
They are recruited independently, do not know each other before
the task, and do not participate in the initial dataset design.
They are not informed of the research hypotheses.
For each item, annotators see the finalized scenario background,
agent descriptions, target agent, question, and label options.
For judgment annotation, the question identifies whether
responsibility or blame is being assessed.

Annotators are not shown the construction scaffold, including
template identifiers, scenario categories, designed variable
settings, or expected rankings among agents.
For judgment annotation, attribution-dimension labels are hidden.
They also do not see model outputs, the other annotator's labels,
or raw and intermediate narrative versions, including the original
Reddit posts.
This separation is intended to make the reference judgments depend
on the same finalized narrative available to the evaluated models,
rather than on the authors' intended dimension manipulations or
expected model behavior.

\paragraph{Training, calibration, and independent annotation.}
Before formal annotation, both experts receive the same guideline,
including dimension definitions, the four-level judgment scales,
and examples of common boundary cases.
They first read it independently, then complete a pilot set drawn
from different scenario and task types.
The authors collect their questions and clarify label boundaries
in the guideline.
Calibration establishes a shared interpretation of the scales;
it is not used to direct annotators toward an expected judgment.

After calibration, the experts annotate the formal dataset
independently without discussing individual samples.
For responsibility and blame judgments, no consensus label is forced
when they disagree.
Agreement is computed on their independent labels before filtering,
and only matching judgments are retained in the final benchmark.
Appendices~\ref{app:annotation_agreement} and
\ref{app:data_statistics_of_filtering} report agreement and the
resulting selection, respectively.

\paragraph{Annotation guidelines.}
The instructions below specify the criteria presented to annotators.
Dimension labels use $\{1,0,\varnothing\}$, with the supplied
\textit{Reality} instructions reserving $\varnothing$ for information
that cannot be determined from the narrative.
Final judgments use a separate four-level ordinal scale.

\paragraph{Instructions for attribution-dimension annotation in the \textit{Reality} subset.}

\begin{quote}
\small
You will read a reconstructed real-world social scenario. For each target agent, please annotate the following five attribution dimensions: \textit{intention}, \textit{voluntariness}, \textit{foreknowledge}, \textit{controllability}, and \textit{obligation}.

For each dimension, please choose one value from $\{1,0,\varnothing\}$:
\begin{itemize}
    \item Use 1 if the dimension is present for the target agent.
    \item Use 0 if the dimension is clearly absent for the target agent.
    \item Use $\varnothing$ if the story does not provide enough information to determine the value.
\end{itemize}

Please judge each dimension only based on the information explicitly stated or reasonably implied in the story. Do not infer hidden facts, motives, or background information that are not supported by the text.

The five dimensions are defined as follows:
\begin{itemize}
    \item \textit{Intention}: Judge whether the target agent deliberately performed the focal action. This refers to the intention behind the action itself, not necessarily the intention to cause the final negative outcome.
    \item \textit{Voluntariness}: Judge whether the target agent acted voluntarily, rather than under coercion, strong pressure, or direct command.
    \item \textit{Foreknowledge}: Judge whether the target agent knew, or should reasonably have been able to foresee, that the action could lead to the negative outcome.
    \item \textit{Controllability}: Judge whether the target agent had the practical ability, opportunity, or means to prevent or avoid the negative outcome.
    \item \textit{Obligation}: Judge whether the target agent had a role-based, relational, or situational duty to prevent the negative outcome.
\end{itemize}
\end{quote}

\paragraph{Instructions for responsibility annotation.}

\begin{quote}
\small
You will read a social scenario, the description of a target agent, and a question asking how much responsibility the target agent should bear for the negative outcome. Please choose one label from \{\texttt{no}, \texttt{low}, \texttt{medium}, \texttt{high}\}.

When making the judgment, please consider whether the target agent contributed to the negative outcome, whether the agent could have acted otherwise, whether the agent had relevant knowledge or control, and whether the agent had a relevant role or duty in the event.

The four labels are defined as follows:
\begin{itemize}
    \item \texttt{high}: Choose this label if the target agent is a central source of the negative outcome and should clearly bear strong responsibility.
    \item \texttt{medium}: Choose this label if the target agent makes a substantial contribution to the negative outcome, but the degree of responsibility is limited by mitigating factors such as limited knowledge, limited control, or role constraints.
    \item \texttt{low}: Choose this label if the target agent is only weakly or indirectly connected to the negative outcome, or if there are strong mitigating factors.
    \item \texttt{no}: Choose this label if the target agent should not be held responsible based on the given scenario.
\end{itemize}
\end{quote}

\paragraph{Instructions for blame annotation.}

\begin{quote}
\small
You will read a social scenario, the description of a target agent, and a question asking how much the target agent should be blamed for the negative outcome. Please choose one label from \{\texttt{no}, \texttt{low}, \texttt{medium}, \texttt{high}\}.

When making the judgment, please consider not only whether the target agent contributed to the outcome, but also whether the agent acted intentionally, knowingly, voluntarily, or in violation of a reasonable expectation or obligation. Please also consider whether there are excuses or mitigating circumstances.

The four labels are defined as follows:
\begin{itemize}
    \item \texttt{high}: Choose this label if the target agent clearly deserves strong blame for the negative outcome.
    \item \texttt{medium}: Choose this label if the target agent deserves a moderate degree of blame, but the blame is reduced by mitigating factors.
    \item \texttt{low}: Choose this label if the target agent deserves only slight blame because the action, knowledge, control, or obligation is limited.
    \item \texttt{no}: Choose this label if the target agent should not be blamed based on the given scenario.
\end{itemize}

Please note that responsibility and blame are related but not identical. A person may be responsible for an outcome but receive lower blame if they lacked intention, foreknowledge, voluntariness, or sufficient control. Please make the responsibility and blame judgments separately according to the specific question.
\end{quote}

\subsection{Inter-Annotator Agreement}
\label{app:annotation_agreement}

We measure agreement on responsibility and blame judgments before
agreement-based filtering.
Exact agreement across all 8,800 questions is 86.8\%.
We also report Cohen's $\kappa$ \citep{cohen1960kappa1},
linearly weighted $\kappa_{\mathrm{lin}}$
\citep{cohen1968weightedkappa}, and quadratically weighted
$\kappa_{\mathrm{quad}}$
\citep{fleiss1973quadratically_weighted_kappa}.
Unweighted $\kappa$ adjusts exact agreement for agreement expected
from the annotators' marginal label distributions.
The weighted measures additionally account for the ordered labels,
with quadratic weighting penalizing larger disagreements more heavily.
Reporting all three distinguishes disagreement about an exact level
from disagreement across more distant judgment levels.

Table~\ref{Inter_annotator_agreement} reports agreement by relation,
subset, task, and language.
The mean unweighted $\kappa$ across the four overall task--language
combinations is 0.826.
All reported coefficients are positive and significantly different
from zero under two-sided tests of $H_0:\kappa=0$ ($p<0.001$).
Agreement is lower for several vicarious-blame and \textit{Reality}
conditions than for general vignettes, consistent with the greater
judgment ambiguity in those settings.
These results support using the expert judgments as an evaluation
reference while retaining the distinction between reliable
agreement and universally shared judgments.
Only questions with matching judgment labels enter the final dataset.

\begin{table}[!htbp]
\centering
\fontsize{9}{11}\selectfont
\setlength{\tabcolsep}{4pt}
\caption{
Inter-annotator agreement ($\times100$).
Resp.\ denotes responsibility.
All coefficients are significantly different from zero
($p<0.001$).
}
\label{Inter_annotator_agreement}
\begin{tabular}{@{}llrrrrrr@{}}
\toprule
Type & Task
& \multicolumn{2}{c}{$\kappa$}
& \multicolumn{2}{c}{$\kappa_{\mathrm{lin}}$}
& \multicolumn{2}{c}{$\kappa_{\mathrm{quad}}$} \\
\cmidrule(lr){3-4}
\cmidrule(lr){5-6}
\cmidrule(lr){7-8}
& & EN & ZH & EN & ZH & EN & ZH \\
\midrule
General & Resp.
& 90.95 & 91.53 & 93.89 & 94.38 & 96.50 & 96.88 \\
& Blame
& 84.72 & 85.42 & 90.52 & 91.03 & 95.01 & 95.35 \\
Vicarious blame & Resp.
& 77.18 & 81.69 & 79.85 & 83.63 & 83.67 & 86.50 \\
& Blame
& 74.49 & 78.88 & 81.93 & 85.51 & 88.84 & 91.57 \\
Commanding chain & Resp.
& 80.86 & 84.86 & 85.95 & 88.85 & 90.97 & 92.76 \\
& Blame
& 83.53 & 86.02 & 90.14 & 91.63 & 95.03 & 95.78 \\
Reality & Resp.
& 75.14 & 74.65 & 84.22 & 84.06 & 91.70 & 91.77 \\
& Blame
& 76.42 & 75.89 & 86.11 & 85.93 & 93.12 & 93.17 \\
\midrule
Overall & Resp.
& 83.33 & 84.06 & 88.87 & 89.44 & 93.75 & 94.17 \\
& Blame
& 81.13 & 81.80 & 88.59 & 89.11 & 94.15 & 94.53 \\
\bottomrule
\end{tabular}
\end{table}

\subsection{Agreement-Based Filtering and Distribution Changes}
\label{app:data_statistics_of_filtering}

Agreement-based filtering removes questions for which the two experts
assign different responsibility or blame labels.
Its purpose is to provide a common reference label for exact-match
evaluation without forcing agreement on ambiguous cases.
Because this selection could alter the evaluated sample, we examine
its effect on subset composition, relation categories, dimension
values, and judgment-label frequencies.
All counts in this subsection refer to task-specific questions and
pool English and Chinese versions.

Table~\ref{tab:filtering_language_balance} gives the retained and
removed counts.
Table~\ref{tab:vignette_filtering_by_scenario_type} compares the three
\textit{Vignette} relation categories, and
Table~\ref{tab:filtering_dimension_distribution} compares dimension
values among questions with available labels.
Table~\ref{tab:filtering_label_distribution} reports the before/after
judgment-label distributions.
These comparisons complement the agreement coefficients:
agreement measures annotation consistency, whereas the distribution
checks show which parts of the sample remain after applying the
retention rule.

Filtering removes 1,161 of 8,800 questions.
The \textit{Vignette} share increases from 60.7\% to 62.5\%, reflecting
the larger proportion of disagreements removed from \textit{Reality}.
The three vignette relation shares change by at most 1.1 percentage
points, and the dimension-value shares change by at most 2.4
percentage points in the rounded tables.
The broad composition is therefore similar before and after
filtering, although judgment-label frequencies do change.
In particular, the retained sample represents expert-agreed cases;
these marginal distribution checks do not recover the excluded
judgment disagreements.

\begin{table}[!htbp]
\fontsize{9}{11}\selectfont
\centering
\caption{Question counts before and after agreement-based filtering. Percentages give each subset's share within the Before or After column; both tasks and languages are pooled.}
\label{tab:filtering_language_balance}
\setlength{\tabcolsep}{10pt}
\begin{tabular}{lccc}
\toprule
\textbf{Subset} & \textbf{Before} & \textbf{After} & \textbf{Removed} \\
\midrule
Vignette & 5,344 (60.7\%) & 4,776 (62.5\%) & 568 \\
Reality  & 3,456 (39.3\%) & 2,863 (37.5\%) & 593 \\
\midrule
Total    & 8,800          & 7,639          & 1,161 \\
\bottomrule
\end{tabular}
\end{table}

% Requires: \usepackage{booktabs}
\begin{table}[!htbp]
\fontsize{9}{11}\selectfont
\centering
\caption{Distribution of the three \textit{Vignette} relation categories before and after filtering. Percentages use the \textit{Vignette} total at the corresponding stage.}
\label{tab:vignette_filtering_by_scenario_type}
\setlength{\tabcolsep}{12pt}
\begin{tabular}{lrr}
\toprule
\textbf{Agent relation} & \textbf{Before} & \textbf{After} \\
\midrule
General Scenarios & 3,840 (71.9\%) & 3,488 (73.0\%) \\
Vicarious Blame Scenarios & 640 (12.0\%) & 541 (11.3\%) \\
Commanding Chain Scenarios & 864 (16.2\%) & 747 (15.7\%) \\
\bottomrule
\end{tabular}
\end{table}

% Requires: \usepackage{booktabs}
\begin{table}[!htbp]
\fontsize{9}{11}\selectfont
\centering
\caption{Attribution-dimension distributions before and after filtering. Percentages use questions with non-missing labels for each dimension at the corresponding stage.}
\label{tab:filtering_dimension_distribution}
\setlength{\tabcolsep}{8pt}
\begin{tabular}{lrrrr}
\toprule
\textbf{Dimension} & \multicolumn{2}{c}{\textbf{Before}} & \multicolumn{2}{c}{\textbf{After}} \\
\cmidrule(lr){2-3}\cmidrule(lr){4-5}
 & \textbf{Value = 1} & \textbf{Value = 0} & \textbf{Value = 1} & \textbf{Value = 0} \\
\midrule
Intention        & 6,097 (69.3\%) & 2,703 (30.7\%) & 5,405 (70.8\%) & 2,234 (29.2\%) \\
Voluntariness    & 3,164 (67.7\%) & 1,509 (32.3\%) & 2,845 (69.2\%) & 1,267 (30.8\%) \\
Foreknowledge    & 4,572 (52.0\%) & 4,228 (48.0\%) & 3,935 (51.5\%) & 3,704 (48.5\%) \\
Controllability  & 5,462 (62.1\%) & 3,338 (37.9\%) & 4,709 (61.6\%) & 2,930 (38.4\%) \\
Obligation       & 3,087 (50.8\%) & 2,993 (49.2\%) & 2,740 (53.2\%) & 2,413 (46.8\%) \\
\bottomrule
\end{tabular}
\end{table}

\begin{table}[!htbp]
\fontsize{9}{11}\selectfont
\centering
\caption{Judgment-label distributions before and after agreement-based filtering, pooled across English and Chinese. Percentages use the row total.}
\label{tab:filtering_label_distribution}
\setlength{\tabcolsep}{3pt}
\begin{tabular}{lllccccc}
\toprule
\textbf{Subset} & \textbf{Task} & \textbf{Stage} & \textbf{High} & \textbf{Medium} & \textbf{Low} & \textbf{No} & \textbf{Total} \\
\midrule
Vignette & Responsibility & Before & 961 (36.0\%) & 919 (34.4\%) & 598 (22.4\%) & 194 (7.3\%) & 2,672 \\
 &  & After & 921 (37.8\%) & 791 (32.4\%) & 539 (22.1\%) & 188 (7.7\%) & 2,439 \\
 & Blame & Before & 583 (21.8\%) & 733 (27.4\%) & 744 (27.8\%) & 612 (22.9\%) & 2,672 \\
 &  & After & 556 (23.8\%) & 550 (23.5\%) & 639 (27.3\%) & 592 (25.3\%) & 2,337 \\
\midrule
Reality & Responsibility & Before & 815 (47.2\%) & 528 (30.6\%) & 189 (10.9\%) & 196 (11.3\%) & 1,728 \\
 &  & After & 730 (51.2\%) & 369 (25.9\%) & 146 (10.2\%) & 181 (12.7\%) & 1,426 \\
 & Blame & Before & 684 (39.6\%) & 496 (28.7\%) & 232 (13.4\%) & 316 (18.3\%) & 1,728 \\
 &  & After & 634 (44.1\%) & 321 (22.3\%) & 183 (12.7\%) & 299 (20.8\%) & 1,437 \\
\bottomrule
\end{tabular}
\end{table}
\subsection{Final Dataset Statistics}
\label{app:dataset_statistics}
\label{sec:dataset_statistics}

\begin{figure}[!htbp]
\centering
\includegraphics[width=\linewidth]{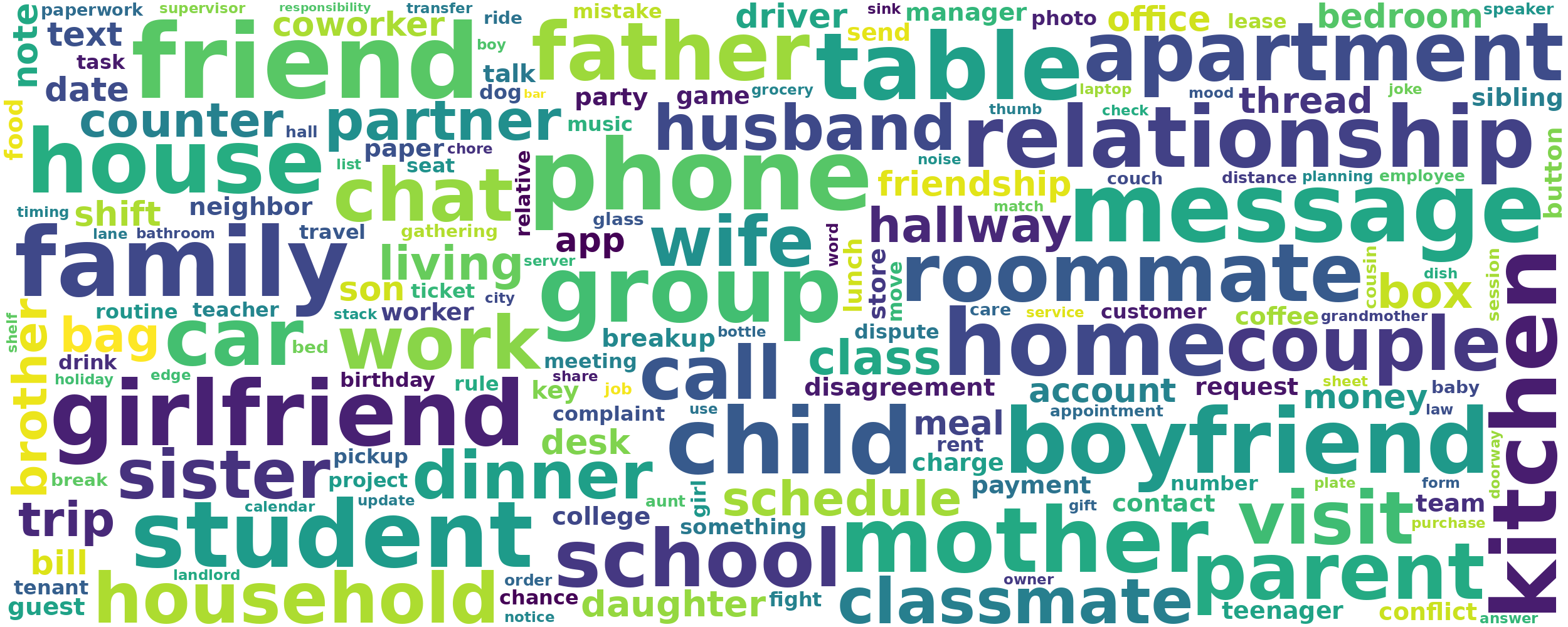}
\caption{High-frequency content words in the English \textit{Reality} narratives.}
\label{fig:wordcloud-en}
\end{figure}

\begin{figure}[!htbp]
\centering
\includegraphics[width=\linewidth]{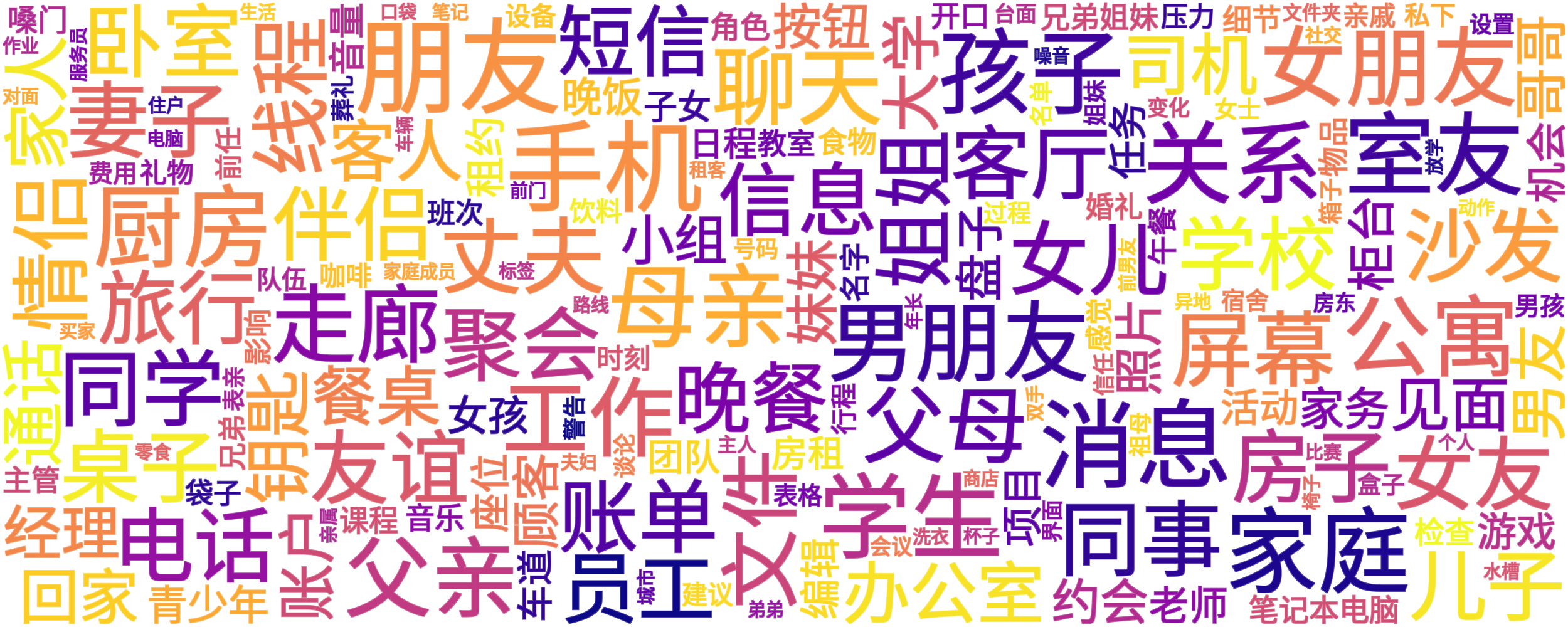}
\caption{High-frequency content words in the Chinese \textit{Reality} narratives.}
\label{fig:wordcloud-zh}
\end{figure}

The final dataset contains 7,639 questions:
4,776 from \textit{Vignette} and 2,863 from \textit{Reality},
with 3,837 English and 3,802 Chinese questions.
Table~\ref{tab:dataset_label_balance} reports language and judgment
distributions.
\mbox{Table~\ref{tab:dataset_dimension_balance}} reports dimension-value
distributions, excluding unspecified or inapplicable labels.
Table~\ref{tab:dataset_scenario_balance} reports the retained
\textit{Vignette} questions by agent relation and scenario type.
The lower denominators for voluntariness and obligation reflect unspecified or inapplicable labels; these are excluded from the corresponding dimension percentages rather than counted as zero.
Figures~\ref{fig:wordcloud-en} and~\ref{fig:wordcloud-zh} visualize frequent content
words in the English and Chinese \textit{Reality} narratives, respectively.

\begin{table}[!htbp]
\centering
\fontsize{9}{11}\selectfont
\setlength{\tabcolsep}{4pt}
\renewcommand{\arraystretch}{1.15}
\caption{
Language and judgment distributions.
Each cell reports a count and its percentage within the row.
Resp.\ denotes responsibility.
}
\label{tab:dataset_label_balance}
\begin{tabular}{@{}llrrrrrr@{}}
\toprule
Subset & Task
& \multicolumn{2}{c}{Language}
& \multicolumn{4}{c}{Judgment} \\
\cmidrule(lr){3-4}
\cmidrule(lr){5-8}
& & EN & ZH & High & Medium & Low & No \\
\midrule
Vignette & Resp.
& \shortstack{1,226\\(50.3\%)}
& \shortstack{1,213\\(49.7\%)}
& \shortstack{921\\(37.8\%)}
& \shortstack{791\\(32.4\%)}
& \shortstack{539\\(22.1\%)}
& \shortstack{188\\(7.7\%)} \\[2pt]
Vignette & Blame
& \shortstack{1,171\\(50.1\%)}
& \shortstack{1,166\\(49.9\%)}
& \shortstack{556\\(23.8\%)}
& \shortstack{550\\(23.5\%)}
& \shortstack{639\\(27.3\%)}
& \shortstack{592\\(25.3\%)} \\[2pt]
Reality & Resp.
& \shortstack{720\\(50.5\%)}
& \shortstack{706\\(49.5\%)}
& \shortstack{730\\(51.2\%)}
& \shortstack{369\\(25.9\%)}
& \shortstack{146\\(10.2\%)}
& \shortstack{181\\(12.7\%)} \\[2pt]
Reality & Blame
& \shortstack{720\\(50.1\%)}
& \shortstack{717\\(49.9\%)}
& \shortstack{634\\(44.1\%)}
& \shortstack{321\\(22.3\%)}
& \shortstack{183\\(12.7\%)}
& \shortstack{299\\(20.8\%)} \\[2pt]
\bottomrule
\end{tabular}
\end{table}

\clearpage
\begin{table}[!htbp]
\centering
\fontsize{9}{11}\selectfont
\setlength{\tabcolsep}{8pt}
\caption{
Attribution-dimension distributions.
Percentages are calculated within each dimension using only
questions with non-missing labels.
}
\label{tab:dataset_dimension_balance}
\begin{tabular}{@{}lrr@{}}
\toprule
Dimension & Value 1 & Value 0 \\
\midrule
Intention
& 5,405 (70.8\%) & 2,234 (29.2\%) \\
Voluntariness
& 2,845 (69.2\%) & 1,267 (30.8\%) \\
Foreknowledge
& 3,935 (51.5\%) & 3,704 (48.5\%) \\
Controllability
& 4,709 (61.6\%) & 2,930 (38.4\%) \\
Obligation
& 2,740 (53.2\%) & 2,413 (46.8\%) \\
\bottomrule
\end{tabular}
\end{table}

\begin{table}[!htbp]
\centering
\fontsize{9}{11}\selectfont
\setlength{\tabcolsep}{6pt}
\caption{
Retained \textit{Vignette} questions by agent relation and scenario type.
The 11 rows cover eight distinct scenario types, with shared types repeated under their respective relations.
Percentages use all 4,776 \textit{Vignette} questions as the denominator.
}
\label{tab:dataset_scenario_balance}
\begin{tabular}{@{}llrr@{}}
\toprule
Agent Relation & Scenario Type & Count & Percentage \\
\midrule
General & Traffic Accident & 755 & 15.8\% \\
General & Food Allergy & 744 & 15.6\% \\
General & Company Scenario & 759 & 15.9\% \\
General & Drowning Incident & 764 & 16.0\% \\
General & Laboratory Scenario & 466 & 9.8\% \\
\midrule
Vicarious Blame & Company Scenario & 248 & 5.2\% \\
Vicarious Blame & Laboratory Scenario & 179 & 3.7\% \\
Vicarious Blame & Family Scenario & 114 & 2.4\% \\
\midrule
Commanding Chain & Company Scenario & 252 & 5.3\% \\
Commanding Chain & Gunman Scenario & 257 & 5.4\% \\
Commanding Chain & Military Strike & 238 & 5.0\% \\
\bottomrule
\end{tabular}
\end{table}

\clearpage

%% file: appendix_sections/benchmark_protocol.tex
\section{Benchmark Evaluation Protocol}
\label{app:evaluation_protocol}

The behavioral evaluation measures agreement with human responsibility and blame judgments at two complementary levels: the label assigned to an individual agent and the relative ordering of agents within a scenario.
The protocol below specifies the evaluated models, elicitation strategies, answer extraction, metrics, and comparison methods.
Complete results are reported in Appendix~\ref{app:full_evaluation_results}.

\subsection{Evaluated Models and Inference Settings}
\label{app:evaluated_models}

We evaluate 32 LLMs, comprising nine models accessed through provider APIs and 23 locally deployed open-source models spanning 0.5--120B parameters.
Table~\ref{tab:List_of_LLMS} lists the model families, developers, versions, parameter scales, and links to their documentation.
The ``closed-source'' grouping used in the results follows the access mode in these experiments: Kimi, MiMo, and Step are included in this group because we used their official APIs, although their weights have also been publicly released.
The open-source group consists of instruction-tuned or chat models.
For the behavioral benchmark, these models are deployed through Ollama\footnote{\url{https://ollama.com/}}, with temperature set to 0 and random seed set to 0.
The hidden-state experiments use the separate backbone configuration described in Appendix~\ref{app:experimental_setup}.

Each model is evaluated on both languages and both judgment tasks under four prompting strategies.
Direct prompting elicits a judgment without an explicit reasoning trace; CoT requests a step-by-step analysis; Self-reflection asks the model to review an initial judgment; and Theory-guided prompting explicitly introduces the five attribution dimensions.
Comparing these strategies examines whether additional reasoning instructions, revision, or access to the theoretical scaffold changes agreement with human judgments.
The Theory-guided prompt supplies dimension names, not the human-annotated dimension values.

\input{figure_and_tables/llms}

\subsection{Prompt Templates}
\label{app:prompt_templates}
\label{app:prompt_template}

\input{appendix_sections/prompt_template}

\subsection{Answer Extraction and Invalid Outputs}
\label{app:answer_extraction}

We use a rule-based parser to map generated answers to
$\mathcal{A}=\{\texttt{no},\texttt{low},\texttt{medium},\texttt{high}\}$.
The parser first searches for all case-insensitive
\texttt{\textbackslash boxed\{\}} spans and scans them from last to first.
It returns the first span that can be mapped to a canonical label.
This order follows the requested output format and prevents an initial answer in a reasoning or self-reflection trace from taking precedence over a later final answer.
If no valid boxed answer is found, the parser falls back to the last standalone occurrence of a canonical label word in the complete output.
An output from which no valid label can be extracted is marked invalid.

Accuracy excludes individual questions with invalid predictions.
For the within-scenario ranking metric, any scenario containing an invalid prediction is excluded.
Consequently, valid-output coverage and agreement conditional on validity answer different questions: a high agreement score does not by itself establish that a model reliably produces an evaluable answer.
Appendix~\ref{app:valid_outputs} reports valid-output rates alongside the complete benchmark results.

\subsection{Evaluation Metrics and Statistical Reporting}
\label{app:evaluation_metrics}

\paragraph{Exact agreement.}
Let $\mathcal{I}_m$ contain the questions with valid predictions for task $m$.
Accuracy is
\begin{equation}
\mathrm{Acc}_m
=\frac{1}{|\mathcal{I}_m|}
\sum_{i\in\mathcal{I}_m}
\mathbb{1}[\hat a_{i,m}=a_{i,m}].
\end{equation}
It measures agreement on the absolute responsibility or blame level assigned to the target agent.
We report it as a percentage, rounded to two decimal places in the full result tables.

\paragraph{Within-scenario ranking agreement.}
For the multi-agent \textit{Vignette} subset, map the labels to ordinal scores with
$r(\texttt{no})=0$, $r(\texttt{low})=1$, $r(\texttt{medium})=2$, and $r(\texttt{high})=3$.
For scenario $s$, let $\mathcal{I}_{m,s}$ index its agent questions and write
\begin{equation}
u_{m,s}=(r(a_{i,m}))_{i\in\mathcal{I}_{m,s}},
\qquad
\hat u_{m,s}=(r(\hat a_{i,m}))_{i\in\mathcal{I}_{m,s}}.
\end{equation}
The Spearman correlation is
\begin{equation}
\rho_{m,s}=\mathrm{Corr}\!\left(
\mathrm{Rank}(u_{m,s}),\mathrm{Rank}(\hat u_{m,s})\right),
\end{equation}
with average ranks assigned to tied values.
We report the mean within-scenario correlation on its original scale, rounded to three decimal places in the full tables.
This metric can distinguish a model that shifts all judgment levels but preserves the ordering of agents from one that assigns responsibility or blame to the wrong agents.
It is restricted to \textit{Vignette} because each \textit{Reality} scenario has only one target agent.

\paragraph{Uncertainty and aggregation.}
The reported 95\% confidence intervals use 1,000 cluster-bootstrap resamples over scenario groups.
Scenario-level resampling accounts for the fact that several agent questions can concern the same event.
The full result tables separate the two tasks and the four prompt strategies.
The main model-comparison figure averages each model's scores over the four strategies, while the model-size analysis additionally averages over responsibility and blame.
Thus, the per-strategy tables retain variation that is hidden by the aggregate comparison.

\paragraph{Model-size analysis.}
For the 23 open-source models with identifiable parameter scales, we fit a separate ordinary least-squares regression for each metric, using the task-averaged agreement score as the response and $\log_{10}$ parameter count as the predictor.
The main figure reports fitted lines and 95\% confidence bands; the text reports the slope significance and $R^2$.
These regressions summarize an association across the evaluated models.
They do not isolate parameter count from model family, training data, architecture, or post-training differences.

\subsection{Non-LLM Baselines}
\label{app:non_llm_models}
\label{app:details_of_basic_baselines}

\input{appendix_sections/details_of_basic_baselines}

%% file: figure_and_tables/llms.tex
\begin{table}[!htbp]
\fontsize{7}{9}\selectfont
\caption{List of LLMs under test and their detailed information.}
\label{tab:List_of_LLMS}
\centering
\resizebox{\linewidth}{!}{%
\begin{tabular}{l l l l l}
\toprule
LLM Family & Developer & Open/Closed-source & Version & Size \\
\midrule
\href{https://docs.anthropic.com/en/docs/about-claude/models}{Claude} & Anthropic & Closed & \texttt{Claude-Haiku-4.5} & - \\
\href{https://www.volcengine.com/docs/82379/1330310}{Doubao Seed} & ByteDance & Closed & \texttt{Doubao-Seed-1.8} & - \\
\href{https://ai.google.dev/gemini-api/docs/models}{Gemini} & Google & Closed & \texttt{Gemini-3.5-flash} & - \\
\href{https://platform.openai.com/docs/models}{ChatGPT} & OpenAI & Closed & \texttt{GPT-5-mini}/\texttt{GPT-5-nano} & - \\
\href{https://docs.x.ai/docs/models}{Grok} & xAI & Closed & \texttt{Grok-4.3} & - \\
\href{https://platform.moonshot.ai/}{Kimi} & Moonshot AI & Closed & \texttt{Kimi-K2.5} & - \\
\href{https://github.com/XiaomiMiMo}{MiMo} & Xiaomi & Closed & \texttt{MiMo-v2.5} & - \\
\href{https://platform.stepfun.com/}{Step} & StepFun & Closed & \texttt{Step-3.7-flash} & - \\

\href{https://github.com/meta-llama/llama3}{Llama} & Meta & Open & \texttt{Llama3.1}/\texttt{Llama3.2} & 1B/8B/70B \\
\href{https://huggingface.co/microsoft/phi-4}{Phi} & Microsoft & Open & \texttt{Phi4} & 14B \\
\href{https://github.com/openai/gpt-oss}{GPT-oss} & OpenAI & Open & \texttt{GPT-oss} & 20B/120B \\
\href{https://huggingface.co/collections/deepseek-ai/deepseek-r1}{DeepSeek-R1} & DeepSeek & Open & \texttt{DeepSeek-R1} & 1.5B/70B\\
\href{https://huggingface.co/collections/Qwen/qwen3}{Qwen} & Alibaba & Open & \texttt{Qwen2}/\texttt{Qwen2.5}/\texttt{Qwen3} & 0.5B/0.6B/1.5B/4B/7B/8B/14B \\
\href{https://ai.google.dev/gemma/docs/core}{Gemma} & Google & Open & \texttt{Gemma3} & 1B/4B/27B\\
\href{https://huggingface.co/zai-org/glm-4-9b}{GLM} & Zhipu AI & Open & \texttt{GLM4} & 9B \\
\href{https://github.com/InternLM/InternLM}{InternLM} & Shanghai AI Laboratory & Open & \texttt{InternLM2.5} & 1.8B/7B \\

\bottomrule
\end{tabular}}
\end{table}

%% file: appendix_sections/prompt_template.tex
The English templates for the four strategies---Direct, CoT, Self-reflection, and Theory-guided---are reproduced below.
For each prompt, \texttt{[SCENARIO]} is replaced with the scenario text, \texttt{[QUESTION]} is replaced with the task-specific question, and \texttt{[AGENT]} is replaced with the target character.
The output label must be one of \texttt{high}, \texttt{medium}, \texttt{low}, or \texttt{no}.

\paragraph{Task questions.}
\begin{quote}
\small
\textbf{Responsibility question:}\\
In the scenario described above, to what extent should \texttt{[AGENT]} be held **responsible**?
Please choose one answer from [high, medium, low, no].

\textbf{Blame question:}\\
In the scenario described above, to what extent should \texttt{[AGENT]} be **blamed**?
Please choose one answer from [high, medium, low, no].
\end{quote}

\paragraph{Direct prompt.}
\begin{quote}
\small
\texttt{[SCENARIO]}

\texttt{[QUESTION]}

Answer directly.
Do not provide reasoning, analysis, explanation, or intermediate steps.
Output only the final answer in \texttt{\textbackslash boxed\{\}}.
The answer inside \texttt{\textbackslash boxed\{\}} must be exactly one of: high, medium, low, no.
\end{quote}

\paragraph{CoT prompt.}
\begin{quote}
\small
\texttt{[SCENARIO]}

\texttt{[QUESTION]}

Please analyze the scenario step by step before answering.
After your analysis, put your final answer directly in \texttt{\textbackslash boxed\{\}}.
The answer inside \texttt{\textbackslash boxed\{\}} must be exactly one of: high, medium, low, no.
\end{quote}

\paragraph{Self-reflection prompt.}
\begin{quote}
\small
\texttt{[SCENARIO]}

\texttt{[QUESTION]}

First give an initial answer with a brief reason, but do not put the initial answer in \texttt{\textbackslash boxed\{\}}.
Then reflect on whether your initial answer missed any important facts or should be revised.
After the reflection, put only the final updated answer directly in \texttt{\textbackslash boxed\{\}}.
The answer inside \texttt{\textbackslash boxed\{\}} must be exactly one of: high, medium, low, no.
\end{quote}

\paragraph{Theory-guided prompt.}
\begin{quote}
\small
\texttt{[SCENARIO]}

\texttt{[QUESTION]}

Before making the final judgment, analyze the case using these five dimensions: intention, voluntariness, foreknowledge, controllability, and obligation.
Then decide the final answer.
Put your final answer directly in \texttt{\textbackslash boxed\{\}}.
The answer inside \texttt{\textbackslash boxed\{\}} must be exactly one of: high, medium, low, no.
\end{quote}

%% file: appendix_sections/details_of_basic_baselines.tex
The five non-LLM baselines provide complementary reference points: label-frequency information, surface textual cues, an explicit attribution-theory rule, and supervised text classification.
Each predicts one of \texttt{no}, \texttt{low}, \texttt{medium}, or \texttt{high} for responsibility and blame and is evaluated with the metrics in Appendix~\ref{app:evaluation_metrics}.
Their information sources differ: in particular, the Attribution-Theory Rule receives annotated attribution dimensions, whereas the text-based methods must work from the narrative.
The results should be interpreted with this distinction in mind.

For each task--subset combination, mBERT uses five random splits with 90\% of the data for training and 10\% for testing.
For a fair comparison among the non-LLM baselines, the other four methods use the same test sets, and accuracy is computed from the pooled test predictions across the five runs.

\paragraph{Random Stratified.}
This baseline samples a label at random according to the empirical label distribution of the corresponding task and dataset subset.
It does not use the scenario text, attribution dimensions, or any learned parameters.
It measures how much performance can be obtained from label balance alone.

\paragraph{Majority Class.}
This baseline always predicts the most frequent label in the corresponding task and dataset subset.
Ties are broken by the fixed label order \texttt{no}, \texttt{low}, \texttt{medium}, \texttt{high}.
It provides a reference for the contribution of the dominant class to exact agreement.

\paragraph{Heuristic Rule.}
This baseline uses only surface cues from the scenario text and does not use the annotated attribution dimensions.
For each record, it first focuses on the sentence or paragraph mentioning the target agent when possible, and otherwise uses the full scenario.
It then matches English and Chinese keywords related to causal involvement, intention, foreknowledge, control, obligation, coercion, inability, accident, justification, and lack of knowledge.
Responsibility is scored mainly from causal involvement, control, obligation, foreknowledge, and intention, with penalties for coercion and mitigation.
Blame gives more weight to intention and foreknowledge, and applies stronger penalties for coercion, accident, justification, and lack of knowledge.
The final score is mapped to the four labels by fixed thresholds: scores at or below zero give \texttt{no}, one gives \texttt{low}, two or three give \texttt{medium}, and four or above gives \texttt{high}.
This baseline tests how much of the benchmark can be handled through explicit surface matching.

\paragraph{Attribution-Theory Rule.}
This baseline uses the annotated attribution dimensions rather than the raw scenario text.
It follows a Mao-and-Gratch-inspired rule over five binary dimensions \citep{mao2012modeling-JAIR}: intention, obligation, voluntariness, controllability, and foreknowledge.
Missing or not-applicable dimension values are treated as zero.
The responsibility rule uses controllability as a key prerequisite, reduces judgments for non-voluntary action, and assigns higher levels in conjunction with intention, foreknowledge, and obligation.
The blame rule uses the same dimensions but is more sensitive to lack of voluntariness and lack of foreknowledge, because these factors provide stronger mitigation for blame.
This comparison assesses the utility of explicitly supplied attribution factors, while also giving the rule a different input advantage from the LLMs and text classifiers.

\paragraph{mBERT Classifier.}
This baseline is a supervised, non-generative encoder model.
We fine-tune bert-base-multilingual-cased\footnote{\url{https://huggingface.co/google-bert/bert-base-multilingual-cased}} with a four-way classification head, using the scenario text as input and the gold attribution label as the target.
Responsibility and blame are trained as separate classification tasks.
Unlike the frequency and hand-written rule baselines, mBERT tests whether supervised learning from the available text--judgment pairs is sufficient to reproduce the annotated judgments.

Complete baseline results and their comparison with the LLMs are given in Appendix~\ref{app:full_evaluation_results}, Table~\ref{tab:basic_baselines}.

%% file: appendix_sections/probing_details.tex
\section{Probing and Intervention Details}
\label{app:experimental_setup}
\label{app:evaluation_probing}

The representation experiments test decodability and theory-aligned influence separately. The implementation below specifies the controls and token-position comparisons; Appendices~\ref{app:additional_decodability} and~\ref{app:additional_interventions} report their results.

\subsection{Backbones and Hidden-State Extraction}
\label{app:hidden_state_extraction}

We use five frozen backbones with CoT prompts: DeepSeek-R1-1.5B (28 Transformer layers), Llama-3.2-1B (16), Llama-3.1-8B (32), Qwen3-4B (36), and Qwen3-8B (36).
The main decodability analysis probes every Transformer layer.
For cross-model comparison, layer depth is normalized by each model's total number of layers; 100\% denotes the final layer.
The selected intervention layers are given in Table~\ref{tab:probing_backbone_layers}.

For each baseline response, we parse the final judgment using the canonical labels in Appendix~\ref{app:answer_extraction} and locate its label-token prediction step $\mathrm{Idx}_i^\star$.
At layer $l$, $z_i^{(l)}=h_{i,\mathrm{Idx}_i^\star}^{(l)}$ denotes the state used to predict the label token, as defined in Section~\ref{sec:decodability_o_a}.
This position is chosen because it follows the available scenario context and the model's generated reasoning and lies at the point where that information is converted into a judgment output.
Throughout these appendices, ``label-token prediction step'' refers to this prediction position, rather than a state obtained after feeding the generated label back into the model.

The last input token is a distinct, earlier position at the end of the prompt, before the generated CoT response.
Probing this position tests whether dimension information is already accessible before the final judgment stage.
Intervening there tests whether a local perturbation can propagate through subsequent autoregressive generation to affect the final answer.
These complementary comparisons are reported in Appendices~\ref{app:early_token_decodability} and~\ref{app:early_token_influence}.

\begin{table}[!htbp]
\centering
\fontsize{9}{11}\selectfont
\setlength{\tabcolsep}{5pt}
\caption{Backbones and intervention layers. Layer numbers use the one-based convention $l\in\{1,\ldots,L\}$. The four intermediate positions are $\lfloor0.2L\rfloor$, $\lfloor0.4L\rfloor$, $\lfloor0.6L\rfloor$, and $\lfloor0.8L\rfloor$; percentages are nominal depth selections.}
\label{tab:probing_backbone_layers}
\begin{tabular}{lrrrrrr}
\toprule
Backbone & $L$ & 20\% & 40\% & 60\% & 80\% & Final \\
\midrule
DeepSeek-R1-1.5B & 28 & 5 & 11 & 16 & 22 & 28 \\
Llama-3.2-1B & 16 & 3 & 6 & 9 & 12 & 16 \\
Llama-3.1-8B & 32 & 6 & 12 & 19 & 25 & 32 \\
Qwen3-4B & 36 & 7 & 14 & 21 & 28 & 36 \\
Qwen3-8B & 36 & 7 & 14 & 21 & 28 & 36 \\
\bottomrule
\end{tabular}
\end{table}

\subsection{Data Splits and Result Aggregation}
\label{app:probe_splits}

Probes are trained and evaluated separately for each language and each subset.
Keeping Chinese and English versions in separate runs prevents a translated version of a held-out scenario from appearing in that probe's training data.

\paragraph{Vignette: leave one scenario type out.}
The eight scenario types describe event settings, whereas the three inter-agent relation categories describe how agents are related (Appendix~\ref{app:vignette_construction}).
We use eight-fold leave-one-scenario-type-out cross-validation: each fold holds out all samples of one scenario type for testing and uses the other seven types for training.
Occurrences of the same type across relation categories belong to the same fold, including all their agent questions and narrative variants.
For example, a Company test fold includes Company templates from General, Vicarious Blame, and Commanding Chain; a Laboratory test fold includes both of its relation-specific templates in Table~\ref{tab:scenarios_details_across_three_relation_types}.
Each scenario type is held out once.
Withholding entire event settings limits the opportunity to memorize recurring scenario content or templates, reducing the risk of inflated decodability from such overlap.
The resulting evaluation tests whether attribution dimensions can be decoded in unseen scenario types, strengthening the evidence for generalization across event settings.

\paragraph{Reality: random cross-validation.}
We use 10-fold random cross-validation, training on nine folds and testing on the remaining fold in each round.
The same subset-specific splits are used for the original probes and decodability controls at both the label-token prediction step and the last input token.

\paragraph{Eligibility and aggregation.}

For dimension $k$, training and evaluation require a non-missing label $y_{i,k}\in\{0,1\}$ and an available hidden state at the probed location.
Questions with unspecified or inapplicable values are excluded for that dimension; they need not be excluded from other dimensions for which labels are available.
Consequently, the eligible sample set can differ across dimensions.
All probe fitting and direction construction use training-fold data; interventions are evaluated on held-out prompts.

For a fixed language--subset combination, model, task, dimension, and layer, we compute balanced accuracy on each held-out fold and take the unweighted mean over the eight \textit{Vignette} folds or the ten \textit{Reality} folds.
This is the summary for an individual language--subset run.
For the combined decodability plots, we instead collect the held-out predictions and labels from all folds of all four language--subset runs and recompute balanced accuracy on that pooled set.
Each eligible question contributes the prediction from its test fold.
Thus, the combined score is computed from pooled predictions, rather than by averaging the fold means or the four run-level scores; these aggregations can differ when fold sizes or class frequencies differ.
Probe training is repeated with seeds 0, 1, and 2, and the resulting scores are averaged across seeds.
Pooling predictions for reporting does not entail fitting a joint bilingual probe.
Balanced accuracy gives the two binary dimension values equal importance, which is necessary because their label frequencies differ.
It is distinct from benchmark accuracy, whose targets are the four human responsibility/blame labels.

\subsection{Probe Training and Decodability Controls}
\label{app:probe_training}

\paragraph{Training objective.}
For fold $f$, dimension $k$, and layer $l$, let
\begin{equation}
\mathcal{D}^{(f,l)}_{\mathrm{tr},k}
=\left\{i\in\mathcal{D}^{(f)}_{\mathrm{tr}}\mid
y_{i,k}\neq\varnothing,\ z_i^{(l)}\text{ available}\right\}.
\end{equation}
The logistic probe is fitted by minimizing class-balanced cross-entropy with $L_2$ regularization:
\begin{equation}
\min_{w,b}\;
\sum_{i\in\mathcal{D}^{(f,l)}_{\mathrm{tr},k}}
\beta_{y_{i,k}}\,
\mathrm{CE}\!\left(y_{i,k},\sigma(w^\top z_i^{(l)}+b)\right)
+\lambda\|w\|_2^2,
\end{equation}
where $\beta_y$ denotes class-balancing weights and
$\mathrm{CE}(y,p)=-y\log p-(1-y)\log(1-p)$.
The held-out set is defined by the same label and hidden-state eligibility conditions.
Decodability is its balanced accuracy, $(\mathrm{Recall}_0+\mathrm{Recall}_1)/2$, with 50\% denoting chance.

High probe accuracy alone does not establish that the model represents the intended construct independently of superficial text patterns \citep{hewitt2019designing,belinkov2022probing}.
The vignette construction therefore varies narrative phrasing and character identities and avoids explicit dimension labels; leaving entire scenario types out of training further reduces shortcuts based on recurring event settings.
We complement these design choices with three controls, using the same splits, class-balanced logistic classifier, and balanced-accuracy metric.

\paragraph{Shuffled-label control.}
For each dimension and training fold, we randomly permute the training labels while leaving hidden states and test labels unchanged.
This preserves the training-label counts but removes their original correspondence with the representations.
It tests whether held-out performance depends on that correspondence rather than on label balance or the probe-fitting procedure alone.

\paragraph{Random-projection control.}
At each layer, we replace the hidden state with $\tilde z_i^{(l)}=R_l^\top z_i^{(l)}$, where
$R_l\in\mathbb{R}^{d\times64}$ and $(R_l)_{ab}\sim\mathcal{N}(0,1/64)$.
The same random matrix is applied to the training and test states at that layer.
This control measures how much dimension information remains after compression to 64 features.
Because it retains a projection of the original representation, it is a compression comparison, not a label-independent chance baseline.

\paragraph{Lexical control.}
We replace hidden states with character 3--5-gram TF-IDF features extracted from the scenario text alone.
This control measures predictive information available from surface text without using the LLM's hidden representations or generated reasoning.
The original-probe and control curves in Appendix~\ref{app:decodability_sanity} therefore provide complementary evidence: shuffled labels check the label--representation association, while random projection and lexical features compare the full representation with restricted alternatives.

\subsection{Intervention Implementation and Hyperparameters}
\label{app:intervention_implementation}

\paragraph{Direction construction.}
At each selected layer and training fold, we derive a direction from the target dimension's probe weight.
Equation~\ref{eq:dimension_direction} removes components explained by the other available dimension-probe weights using regularized residualization, then normalizes the residual to unit length.
This reduces shared linear components among correlated dimensions; it does not establish complete independence among the underlying attribution factors.
Degenerate residuals are skipped.
The positive direction is oriented to increase probe evidence for $y_{i,k}=1$.
The directional expectations apply to the negative-outcome scenarios considered here, as discussed in Appendix~\ref{app:attribution_theories}.

\paragraph{Prediction-step intervention.}
For each held-out prompt, baseline generation identifies $\mathrm{Idx}_i^\star$.
A single-use forward hook adds $\eta\alpha v_k^{(l)}$ at that prediction step and selected layer, following Equation~\ref{eq:dimension_intervention}.
We generate exactly one token at the edited step and obtain the comparison prediction by restricting the argmax to the four canonical labels.
The hook is local to that step: the main intervention asks whether the decoded direction can influence the final judgment at the output decision stage.
The earlier-token experiment moves both probe training and the edit to the last input token and permits subsequent autoregressive generation, providing a separate test of effects initiated before this decision stage.

\paragraph{Matched controls.}
The unedited baseline establishes the direction of change in the judgment.
The random-direction control uses a unit vector independent of dimension labels and probe weights.
The shuffled-label control derives a direction from a training-fold probe fitted to shuffled labels, using the same direction-construction procedure.
Both intervention controls match the target intervention's layer, prediction step, strength, sign, and perturbation norm.
This matching tests whether the learned dimension direction has a stronger theory-aligned effect than a generic perturbation of the same size or a direction learned without the original label association.

\begin{table}[!htbp]
\centering
\fontsize{8}{10}\selectfont
\setlength{\tabcolsep}{5pt}
\caption{Settings shared across the probing and intervention analyses. Backbone-specific layer selections appear in Table~\ref{tab:probing_backbone_layers}.}
\label{tab:probing_intervention_settings}
\begin{tabular}{ll}
\toprule
Setting & Value or procedure \\
\midrule
Prompt strategy & CoT \\
Probe-training runs & Separate for each language and subset \\
Vignette cross-validation & 8 folds; one scenario type held out, seven used for training \\
Reality cross-validation & 10 random folds; one held out, nine used for training \\
Probe-training seeds & 0, 1, 2 \\
Decodability score & Balanced accuracy; chance 50\% \\
Random-projection dimension & 64 \\
Lexical features & Scenario-text character 3--5-gram TF-IDF \\
Direction norm & Unit $L_2$ norm \\
Intervention signs & $+1$ for unedited \texttt{no}; $-1$ for \texttt{high}; both otherwise \\
Intervention strengths & $\alpha\in\{1,2,4,8,16\}$ \\
Main intervention strength & $\alpha=8$, the pre-specified representative setting \\
Test unit & One binary outcome per question and comparison \\
Success criterion & Every feasible sign strictly aligns; ties count as failures \\
Directional test & One-sided exact binomial test; null threshold $\pi_c=1/2$ \\
Passing criterion & $N>0$ and $p_c<0.01$ for all three comparisons \\
\bottomrule
\end{tabular}
\end{table}

The full strength grid characterizes how the average signed judgment change varies with perturbation size.
Appendix~\ref{app:intervention_strength} reports this descriptive sensitivity analysis using the continuous measure $D(\alpha)$, averaged equally across the evaluated layer--dimension settings.

\subsection{Statistical Testing and Reporting Details}
\label{app:directional_test}

For each setting $(m,l,k,\alpha)$, the unedited output fixes $\mathcal{E}_i$ for all three comparisons, with one sign at the endpoints and both signs otherwise (Table~\ref{tab:probing_intervention_settings}).
Equation~\ref{eq:joint_directional_success} assigns $s_i^c=1$ only when every sign in $\mathcal{E}_i$ has $\Delta_{i,\eta}^c>0$; a tie or opposite shift on any tested sign gives $s_i^c=0$.
Each question, including endpoint cases, contributes one binary outcome per comparison and remains in $N$.

For $S_c=\sum_{i=1}^{N}s_i^c$, we apply the one-sided exact binomial test in Equation~\ref{eq:directional_test} to $H_0:\pi_c\leq1/2$ against $H_1:\pi_c>1/2$.
The threshold $1/2$ tests whether more than half of the questions satisfy their applicable directional criterion.
The binomial model assumes independent question-level indicators, without requiring independence between intervention outcomes within questions that have two feasible signs.
The reported $p$-values do not account for dependence among questions derived from the same scenario.
A setting is reported as ``Tests passed'' (blue in the heatmaps) only when $N>0$ and all three comparisons satisfy $p_c<0.01$.
The same criterion is applied separately against the unedited baseline, random-direction output, and shuffled-label output.

``Tests not passed'' (gray) denotes $N=0$ or insufficient evidence in at least one comparison; it does not establish the absence of a dimension or its influence.
The binary tests assess directional success, while $D(\alpha)$ (Appendix~\ref{app:intervention_strength}) averages signed changes over the same feasible signs before averaging across questions.

%% file: appendix_sections/benchmark_results.tex
\section{Full Benchmark Results}
\label{app:full_evaluation_results}

The aggregate comparisons in Section~\ref{sec:performance_evaluation} summarize agreement with human judgments, but can conceal differences between prompting strategies, tasks, and scenario groups. We therefore report the complete results and complementary breakdowns below. These analyses establish the behavioral setting for the representation experiments: successful recovery of attribution dimensions should not be confused with complete agreement between model outputs and human judgments.

\subsection{Full LLM Results across Prompting Strategies}
\label{app:full_llm_results}

Tables~\ref{tab:accuracy_full} and~\ref{tab:spearman_vignette} report all 32 LLMs under the four prompting strategies. Accuracy measures exact agreement with the four-level human judgments across both subsets; Spearman's $\rho$ measures within-scenario agent-ranking agreement on \textit{Vignette}. The two metrics distinguish calibration of judgment levels from preservation of relative responsibility or blame. In the tables, N denotes the Direct prompt, SR denotes Self-reflection, and TG denotes Theory-guided prompting; Avg. averages over the four strategies.

\input{figure_and_tables/benchmark_acc}
\input{figure_and_tables/benchmark_rho}

\paragraph{Task and metric differences.}
GPT-5-mini has the highest prompt-averaged responsibility accuracy (56.87\%), whereas Step-3.7-flash has the highest responsibility rank agreement ($\rho=0.588$). Grok-4.3 leads on both blame accuracy (60.03\%) and blame rank agreement ($\rho=0.702$). Across all 32 LLMs, responsibility accuracy exceeds blame accuracy on average (43.10\% versus 39.82\%), but rank agreement is higher for blame (0.400 versus 0.348). Most models follow these task-specific patterns. Thus, matching the absolute judgment category and preserving the ordering of agents are empirically distinct aspects of social attribution. Models can rank agents' blame reasonably well while disagreeing with the precise severity assigned by human annotators.

\paragraph{Prompting strategy.}
Averaging over all models and both tasks, accuracy is 40.45\% for Direct, 43.13\% for CoT, 40.33\% for Self-reflection, and 41.93\% for Theory-guided prompting; the corresponding correlations are 0.373, 0.397, 0.337, and 0.387. CoT also has the highest mean score within each task. This comparison motivates examining attribution representations during CoT-based judgments: explicit reasoning improves behavioral agreement on average, while leaving substantial disagreement to investigate. Theory-guided prompting also improves the aggregate scores over Direct, but explicitly asking for the five dimensions does not guarantee the best judgment. Self-reflection does not improve agreement on average. These are aggregate comparisons, with individual model--strategy exceptions visible in the tables.

\paragraph{Model families and generations.}
The closed-source/API group averages 54.50\% responsibility accuracy and 52.59\% blame accuracy, compared with 38.64\% and 34.82\% for the open-source group; the corresponding correlations are 0.529/0.636 and 0.276/0.307. Family-level comparisons further show why parameter count alone is insufficient. At 8B parameters, Qwen3-8B exceeds Llama-3.1-8B on both tasks and metrics. Within the Qwen family, the 7B--8B sequence Qwen2-7B, Qwen2.5-7B, and Qwen3-8B improves in responsibility accuracy from 37.77\% to 42.26\% to 48.43\%, and in blame accuracy from 33.31\% to 37.04\% to 41.56\%. Rank agreement improves along the same sequence. Differences in training and model generation therefore remain relevant when interpreting the model-size association in Figure~\ref{fig:model_size_trends}.

\clearpage
\subsection{Full Non-LLM Baseline Results}
\label{app:full_baseline_results}

The five baselines test how much agreement can be obtained from label frequencies, surface textual cues, explicit attribution dimensions, or supervised text classification. Their implementations are given in Appendix~\ref{app:non_llm_models}; Table~\ref{tab:basic_baselines} reports both tasks and metrics.

\input{figure_and_tables/benchmark_basic_model}

mBERT has the highest responsibility accuracy among the baselines (43.47\%), while the Attribution-Theory Rule has the highest responsibility rank agreement (0.303). On blame, the Attribution-Theory Rule leads on both accuracy (46.91\%) and correlation (0.368). This contrast motivates distinguishing textual prediction from reasoning over attribution-relevant factors. However, the rule baseline receives annotated dimension values, whereas mBERT and the LLMs must infer relevant information from text; the comparison therefore does not isolate model architecture or establish that a particular reasoning mechanism is necessary. The strongest LLMs exceed these baselines, but the limited absolute agreement across methods reinforces the need to examine intermediate attribution representations.

\clearpage
\subsection{Valid-Output Rates}
\label{app:valid_outputs}
\label{app:llm_valid}

Because the evaluation excludes invalid predictions, output validity is needed to interpret the coverage of the reported agreement scores. A model that seldom produces an extractable answer should not be assessed from its agreement score alone. The parser and task-specific exclusion rules are described in Appendix~\ref{app:answer_extraction}.

\input{appendix_sections/valid_statistic}

\clearpage
\subsection{Performance Breakdowns}
\label{app:error_breakdown}

The benchmark combines controlled relations and realistic narratives, with multiple values of the attribution dimensions. The following breakdowns check whether aggregate performance conceals systematic variation across scenario groups and dimension values. They complement the latent-space experiments by identifying behavioral differences that a single overall score cannot reveal.

\input{appendix_sections/error_breakdown}

%% file: figure_and_tables/benchmark_acc.tex
% =============================================================================
% llm acc
% =============================================================================
\begin{table}[!htbp]
\caption{Accuracy of 32 representative LLMs on our constructed social attribution benchmark.}
\label{tab:accuracy_full}
\centering
\fontsize{6}{7.2}\selectfont
\setlength{\tabcolsep}{1.5pt}
\begin{minipage}[t]{0.49\linewidth}
\vspace{0pt}
\centering
\textbf{(a) Responsibility judgment}\par\smallskip
\begin{tabular}{lcccccc}
\toprule
 & \multicolumn{4}{c}{Prompt Strategy} & & \\
\cmidrule(lr){2-5}
Model & N & CoT & SR & TG & Avg. & 95\% CI \\
\midrule
\multicolumn{7}{c}{Closed-source Models} \\
\midrule
\rowcolor{teal!7}\texttt{Claude-Haiku-4.5} & 55.71 & 52.01 & 54.75 & 56.09 & 54.64 & [53.16, 56.16] \\
\texttt{Doubao-Seed-1.8} & 52.70 & 52.91 & 52.39 & 54.51 & 53.13 & [51.34, 54.81] \\
\rowcolor{teal!7}\texttt{Gemini-3.5-flash} & 50.09 & 52.19 & 52.82 & 52.73 & 51.96 & [50.35, 53.77] \\
\texttt{GPT-5-mini} & \textbf{57.18} & 57.67 & 55.96 & 56.66 & \textbf{56.87} & [55.36, 58.38] \\
\texttt{GPT-5-nano} & 53.04 & 50.79 & 47.35 & 53.04 & 51.05 & [49.64, 52.66] \\
\rowcolor{teal!7}\texttt{Grok-4.3} & 56.09 & \underline{57.75} & \underline{56.09} & 53.95 & 55.97 & [54.33, 57.58] \\
\texttt{Kimi-K2.5} & 54.49 & 55.73 & 52.86 & \underline{57.77} & 55.21 & [53.70, 56.74] \\
\rowcolor{teal!7}\texttt{MiMo-v2.5} & \underline{56.77} & 55.55 & 54.83 & \textbf{58.82} & \underline{56.49} & [54.92, 58.03] \\
\texttt{Step-3.7-flash} & 53.91 & \textbf{58.47} & \textbf{56.72} & 51.52 & 55.15 & [53.27, 57.00] \\
\midrule
\multicolumn{7}{c}{Open-source Models} \\
\midrule
\rowcolor{teal!7}\texttt{Deepseek-R1-70B} & 47.24 & 49.86 & 46.42 & 43.80 & 46.83 & [45.28, 48.41] \\
\rowcolor{teal!7}\texttt{Deepseek-R1-1.5B} & 27.48 & 28.56 & 27.61 & 30.51 & 28.54 & [27.63, 29.38] \\
\texttt{Gemma3-27B} & 51.98 & 52.70 & 48.62 & 46.88 & 50.05 & [48.63, 51.54] \\
\texttt{Gemma3-4B} & 41.63 & 47.87 & 42.23 & 40.80 & 43.13 & [42.02, 44.28] \\
\texttt{Gemma3-1B} & 16.07 & 39.07 & 34.75 & 36.61 & 31.62 & [30.88, 32.38] \\
\rowcolor{teal!7}\texttt{Glm4-9B} & 39.38 & 47.76 & 41.37 & 37.05 & 41.39 & [40.43, 42.41] \\
\texttt{GPT-oss-120B} & 54.33 & 53.89 & 53.45 & 54.64 & 54.08 & [52.46, 55.57] \\
\texttt{GPT-oss-20B} & 52.57 & 52.52 & 51.69 & 50.40 & 51.80 & [50.22, 53.42] \\
\rowcolor{teal!7}\texttt{InternLM2.5-7B} & 36.12 & 41.16 & 35.89 & 35.24 & 37.10 & [35.99, 38.28] \\
\rowcolor{teal!7}\texttt{InternLM2.5-1.8B} & 25.59 & 33.76 & 33.45 & 30.38 & 30.80 & [29.70, 31.79] \\
\texttt{Llama3.1-70B} & 49.91 & 55.94 & 51.10 & 52.65 & 52.40 & [50.98, 53.82] \\
\texttt{Llama3.1-8B} & 36.07 & 39.90 & 38.55 & 38.06 & 38.14 & [37.07, 39.13] \\
\texttt{Llama3.2-1B} & 17.72 & 14.85 & 9.99 & 13.74 & 14.08 & [13.19, 14.89] \\
\rowcolor{teal!7}\texttt{Phi4-14B} & 49.42 & 56.40 & 52.24 & 50.09 & 52.04 & [50.58, 53.51] \\
\texttt{Qwen2-7B} & 36.82 & 40.60 & 38.84 & 34.83 & 37.77 & [36.57, 38.92] \\
\texttt{Qwen2-1.5B} & 19.66 & 15.58 & 31.57 & 26.78 & 23.40 & [22.76, 24.06] \\
\texttt{Qwen2-0.5B} & 16.02 & 14.26 & 16.48 & 16.14 & 15.72 & [14.50, 16.91] \\
\rowcolor{teal!7}\texttt{Qwen2.5-7B} & 42.90 & 49.70 & 34.75 & 41.68 & 42.26 & [41.02, 43.43] \\
\rowcolor{teal!7}\texttt{Qwen2.5-1.5B} & 10.97 & 31.38 & 25.92 & 20.54 & 22.21 & [21.53, 22.88] \\
\texttt{Qwen3-14B} & 50.43 & 53.22 & 45.30 & 54.49 & 50.86 & [49.50, 52.27] \\
\texttt{Qwen3-8B} & 46.70 & 50.87 & 43.10 & 53.07 & 48.43 & [47.12, 49.74] \\
\texttt{Qwen3-4B} & 48.23 & 48.77 & 47.58 & 48.95 & 48.38 & [46.63, 49.98] \\
\texttt{Qwen3-0.6B} & 27.04 & 26.34 & 26.29 & 31.57 & 27.81 & [27.09, 28.51] \\
\bottomrule
\end{tabular}
\end{minipage}
\hfill
\begin{minipage}[t]{0.49\linewidth}
\vspace{0pt}
\centering
\textbf{(b) Blame judgment}\par\smallskip
\begin{tabular}{lcccccc}
\toprule
 & \multicolumn{4}{c}{Prompt Strategy} & & \\
\cmidrule(lr){2-5}
Model & N & CoT & SR & TG & Avg. & 95\% CI \\
\midrule
\multicolumn{7}{c}{Closed-source Models} \\
\midrule
\rowcolor{teal!7}\texttt{Claude-Haiku-4.5} & 52.23 & 48.97 & 48.99 & 54.27 & 51.11 & [49.60, 52.58] \\
\texttt{Doubao-Seed-1.8} & 49.92 & 52.65 & 52.17 & 57.10 & 52.96 & [51.32, 54.48] \\
\rowcolor{teal!7}\texttt{Gemini-3.5-flash} & 46.79 & 55.09 & 55.72 & 56.65 & 53.56 & [52.11, 55.11] \\
\texttt{GPT-5-mini} & 51.19 & 48.65 & 46.77 & 45.34 & 47.99 & [46.20, 49.84] \\
\texttt{GPT-5-nano} & 49.84 & 46.85 & 45.84 & 44.12 & 46.66 & [45.17, 48.00] \\
\rowcolor{teal!7}\texttt{Grok-4.3} & \textbf{60.31} & \textbf{60.60} & \textbf{60.25} & \textbf{58.96} & \textbf{60.03} & [58.53, 61.49] \\
\texttt{Kimi-K2.5} & 51.14 & 55.80 & 49.68 & \underline{58.45} & 53.77 & [52.45, 55.17] \\
\rowcolor{teal!7}\texttt{MiMo-v2.5} & 50.66 & 49.63 & 46.34 & 51.41 & 49.51 & [48.00, 50.87] \\
\texttt{Step-3.7-flash} & \underline{55.81} & \underline{60.60} & \underline{58.45} & 56.00 & \underline{57.72} & [56.06, 59.50] \\
\midrule
\multicolumn{7}{c}{Open-source Models} \\
\midrule
\rowcolor{teal!7}\texttt{Deepseek-R1-70B} & 40.70 & 46.79 & 41.84 & 41.97 & 42.83 & [41.23, 44.23] \\
\rowcolor{teal!7}\texttt{Deepseek-R1-1.5B} & 24.64 & 25.13 & 24.40 & 24.83 & 24.75 & [23.88, 25.64] \\
\texttt{Gemma3-27B} & 45.57 & 45.23 & 43.14 & 40.36 & 43.57 & [41.89, 45.15] \\
\texttt{Gemma3-4B} & 34.50 & 36.46 & 32.75 & 34.98 & 34.67 & [33.47, 36.02] \\
\texttt{Gemma3-1B} & 23.16 & 28.43 & 26.10 & 29.01 & 26.68 & [25.90, 27.49] \\
\rowcolor{teal!7}\texttt{Glm4-9B} & 32.33 & 42.53 & 32.86 & 38.58 & 36.57 & [35.46, 37.74] \\
\texttt{GPT-oss-120B} & 48.68 & 49.71 & 47.83 & 47.85 & 48.52 & [46.96, 50.11] \\
\texttt{GPT-oss-20B} & 45.28 & 45.57 & 45.07 & 45.71 & 45.41 & [44.09, 46.80] \\
\rowcolor{teal!7}\texttt{InternLM2.5-7B} & 30.47 & 32.43 & 28.35 & 28.86 & 30.03 & [28.79, 31.30] \\
\rowcolor{teal!7}\texttt{InternLM2.5-1.8B} & 25.09 & 26.34 & 21.97 & 25.25 & 24.66 & [23.65, 25.73] \\
\texttt{Llama3.1-70B} & 48.46 & 53.29 & 47.09 & 52.99 & 50.46 & [49.16, 51.74] \\
\texttt{Llama3.1-8B} & 30.15 & 36.22 & 34.18 & 32.35 & 33.23 & [32.15, 34.44] \\
\texttt{Llama3.2-1B} & 21.78 & 11.63 & 10.73 & 18.57 & 15.68 & [14.77, 16.64] \\
\rowcolor{teal!7}\texttt{Phi4-14B} & 48.20 & 52.70 & 43.35 & 46.03 & 47.57 & [46.20, 48.96] \\
\texttt{Qwen2-7B} & 31.90 & 36.22 & 30.45 & 34.66 & 33.31 & [32.09, 34.54] \\
\texttt{Qwen2-1.5B} & 22.97 & 17.33 & 25.30 & 23.13 & 22.18 & [21.44, 22.87] \\
\texttt{Qwen2-0.5B} & 21.52 & 21.36 & 22.36 & 20.88 & 21.53 & [20.31, 22.84] \\
\rowcolor{teal!7}\texttt{Qwen2.5-7B} & 34.82 & 43.96 & 30.63 & 38.74 & 37.04 & [35.82, 38.24] \\
\rowcolor{teal!7}\texttt{Qwen2.5-1.5B} & 24.06 & 29.09 & 27.11 & 26.79 & 26.76 & [26.01, 27.57] \\
\texttt{Qwen3-14B} & 40.20 & 47.38 & 36.78 & 43.00 & 41.84 & [40.39, 43.12] \\
\texttt{Qwen3-8B} & 41.02 & 44.57 & 34.74 & 45.92 & 41.56 & [40.15, 42.96] \\
\texttt{Qwen3-4B} & 44.78 & 44.49 & 44.20 & 45.95 & 44.85 & [43.38, 46.40] \\
\texttt{Qwen3-0.6B} & 26.29 & 26.71 & 25.01 & 30.55 & 27.14 & [26.36, 27.99] \\
\bottomrule
\end{tabular}
\end{minipage}
\par\smallskip
\par{\scriptsize * The best results are highlighted in \textbf{bold} and the second-best results are \underline{underlined}.
Accuracy is reported as a percentage and rounded to two decimal places.
The 95\% confidence intervals are for Avg. and are cluster-bootstrapped over scenario groups.
Strategy abbreviations: N = Direct, CoT = chain-of-thought, SR = Self-reflection, TG = Theory-guided.}
\end{table}

%% file: figure_and_tables/benchmark_rho.tex
% =============================================================================
% llm rho
% =============================================================================
\begin{table}[!htbp]
\caption{Spearman $\rho$ of 32 representative LLMs on our constructed social attribution benchmark.}
\label{tab:spearman_vignette}
\centering
\fontsize{6}{7.2}\selectfont
\setlength{\tabcolsep}{1.5pt}
\begin{minipage}[t]{0.49\linewidth}
\vspace{0pt}
\centering
\textbf{(a) Responsibility judgment}\par\smallskip
\begin{tabular}{lcccccc}
\toprule
 & \multicolumn{4}{c}{Prompt Strategy} & & \\
\cmidrule(lr){2-5}
Model & N & CoT & SR & TG & Avg. & 95\% CI \\
\midrule
\multicolumn{7}{c}{Closed-source Models} \\
\midrule
\rowcolor{teal!7}\texttt{Claude-Haiku-4.5} & 0.554 & 0.534 & 0.538 & 0.544 & 0.542 & [0.509, 0.575] \\
\texttt{Doubao-Seed-1.8} & 0.535 & 0.540 & 0.530 & 0.532 & 0.534 & [0.495, 0.571] \\
\rowcolor{teal!7}\texttt{Gemini-3.5-flash} & 0.519 & 0.554 & 0.521 & 0.541 & 0.534 & [0.497, 0.570] \\
\texttt{GPT-5-mini} & \underline{0.556} & \underline{0.567} & 0.538 & 0.534 & 0.549 & [0.515, 0.584] \\
\texttt{GPT-5-nano} & 0.435 & 0.411 & 0.344 & 0.440 & 0.407 & [0.372, 0.443] \\
\rowcolor{teal!7}\texttt{Grok-4.3} & 0.545 & 0.566 & \underline{0.547} & \underline{0.550} & \underline{0.552} & [0.515, 0.585] \\
\texttt{Kimi-K2.5} & 0.513 & 0.543 & 0.488 & 0.515 & 0.515 & [0.478, 0.549] \\
\rowcolor{teal!7}\texttt{MiMo-v2.5} & 0.539 & 0.546 & 0.535 & 0.547 & 0.542 & [0.506, 0.575] \\
\texttt{Step-3.7-flash} & 0.550 & \textbf{0.595} & \textbf{0.655} & \textbf{0.554} & \textbf{0.588} & [0.554, 0.629] \\
\midrule
\multicolumn{7}{c}{Open-source Models} \\
\midrule
\rowcolor{teal!7}\texttt{Deepseek-R1-70B} & 0.464 & 0.476 & 0.430 & 0.331 & 0.426 & [0.393, 0.457] \\
\rowcolor{teal!7}\texttt{Deepseek-R1-1.5B} & 0.063 & 0.107 & 0.057 & 0.133 & 0.090 & [0.069, 0.110] \\
\texttt{Gemma3-27B} & 0.506 & 0.539 & 0.457 & 0.457 & 0.490 & [0.455, 0.521] \\
\texttt{Gemma3-4B} & 0.344 & 0.392 & 0.233 & 0.363 & 0.333 & [0.302, 0.360] \\
\texttt{Gemma3-1B} & 0.072 & 0.107 & 0.041 & 0.085 & 0.076 & [0.060, 0.091] \\
\rowcolor{teal!7}\texttt{Glm4-9B} & 0.426 & 0.415 & 0.193 & 0.414 & 0.362 & [0.333, 0.392] \\
\texttt{GPT-oss-120B} & 0.515 & 0.500 & 0.488 & 0.501 & 0.501 & [0.464, 0.535] \\
\texttt{GPT-oss-20B} & 0.462 & 0.455 & 0.466 & 0.453 & 0.459 & [0.422, 0.494] \\
\rowcolor{teal!7}\texttt{InternLM2.5-7B} & 0.310 & 0.318 & 0.228 & 0.306 & 0.290 & [0.262, 0.320] \\
\rowcolor{teal!7}\texttt{InternLM2.5-1.8B} & 0.044 & 0.002 & 0.030 & -0.096 & -0.005 & [-0.020, 0.010] \\
\texttt{Llama3.1-70B} & \textbf{0.558} & 0.521 & 0.457 & 0.509 & 0.511 & [0.480, 0.546] \\
\texttt{Llama3.1-8B} & 0.293 & 0.306 & 0.154 & 0.194 & 0.237 & [0.212, 0.263] \\
\texttt{Llama3.2-1B} & 0.000 & 0.020 & 0.035 & 0.015 & 0.017 & [0.009, 0.025] \\
\rowcolor{teal!7}\texttt{Phi4-14B} & 0.484 & 0.551 & 0.516 & 0.506 & 0.514 & [0.483, 0.547] \\
\texttt{Qwen2-7B} & 0.331 & 0.301 & 0.271 & 0.316 & 0.305 & [0.279, 0.332] \\
\texttt{Qwen2-1.5B} & 0.079 & 0.006 & -0.027 & -0.015 & 0.011 & [-0.001, 0.023] \\
\texttt{Qwen2-0.5B} & 0.037 & -0.010 & -0.012 & -0.006 & 0.002 & [-0.015, 0.020] \\
\rowcolor{teal!7}\texttt{Qwen2.5-7B} & 0.324 & 0.378 & 0.200 & 0.364 & 0.316 & [0.286, 0.349] \\
\rowcolor{teal!7}\texttt{Qwen2.5-1.5B} & -0.046 & 0.102 & 0.047 & 0.098 & 0.050 & [0.031, 0.069] \\
\texttt{Qwen3-14B} & 0.427 & 0.464 & 0.410 & 0.478 & 0.445 & [0.408, 0.477] \\
\texttt{Qwen3-8B} & 0.362 & 0.446 & 0.420 & 0.494 & 0.430 & [0.397, 0.465] \\
\texttt{Qwen3-4B} & 0.437 & 0.440 & 0.414 & 0.434 & 0.432 & [0.397, 0.472] \\
\texttt{Qwen3-0.6B} & -0.023 & 0.125 & 0.001 & 0.157 & 0.065 & [0.046, 0.085] \\
\bottomrule
\end{tabular}
\end{minipage}
\hfill
\begin{minipage}[t]{0.49\linewidth}
\vspace{0pt}
\centering
\textbf{(b) Blame judgment}\par\smallskip
\begin{tabular}{lcccccc}
\toprule
 & \multicolumn{4}{c}{Prompt Strategy} & & \\
\cmidrule(lr){2-5}
Model & N & CoT & SR & TG & Avg. & 95\% CI \\
\midrule
\multicolumn{7}{c}{Closed-source Models} \\
\midrule
\rowcolor{teal!7}\texttt{Claude-Haiku-4.5} & 0.666 & 0.649 & 0.622 & 0.668 & 0.651 & [0.623, 0.679] \\
\texttt{Doubao-Seed-1.8} & 0.635 & 0.655 & 0.632 & 0.670 & 0.648 & [0.618, 0.677] \\
\rowcolor{teal!7}\texttt{Gemini-3.5-flash} & 0.488 & 0.676 & 0.668 & 0.690 & 0.631 & [0.604, 0.658] \\
\texttt{GPT-5-mini} & \underline{0.672} & 0.649 & 0.654 & 0.658 & \underline{0.658} & [0.631, 0.685] \\
\texttt{GPT-5-nano} & 0.532 & 0.535 & 0.497 & 0.518 & 0.521 & [0.492, 0.552] \\
\rowcolor{teal!7}\texttt{Grok-4.3} & \textbf{0.695} & \textbf{0.695} & \textbf{0.700} & \textbf{0.718} & \textbf{0.702} & [0.675, 0.727] \\
\texttt{Kimi-K2.5} & 0.637 & \underline{0.694} & 0.604 & \underline{0.693} & 0.657 & [0.627, 0.684] \\
\rowcolor{teal!7}\texttt{MiMo-v2.5} & 0.615 & 0.626 & 0.584 & 0.591 & 0.604 & [0.573, 0.634] \\
\texttt{Step-3.7-flash} & 0.669 & 0.668 & \underline{0.677} & 0.584 & 0.650 & [0.611, 0.680] \\
\midrule
\multicolumn{7}{c}{Open-source Models} \\
\midrule
\rowcolor{teal!7}\texttt{Deepseek-R1-70B} & 0.503 & 0.549 & 0.453 & 0.402 & 0.477 & [0.448, 0.504] \\
\rowcolor{teal!7}\texttt{Deepseek-R1-1.5B} & 0.051 & 0.010 & 0.035 & 0.087 & 0.046 & [0.024, 0.068] \\
\texttt{Gemma3-27B} & 0.565 & 0.582 & 0.497 & 0.542 & 0.546 & [0.516, 0.575] \\
\texttt{Gemma3-4B} & 0.239 & 0.317 & 0.098 & 0.377 & 0.258 & [0.234, 0.282] \\
\texttt{Gemma3-1B} & 0.043 & 0.012 & 0.054 & 0.043 & 0.038 & [0.021, 0.054] \\
\rowcolor{teal!7}\texttt{Glm4-9B} & 0.357 & 0.472 & 0.194 & 0.479 & 0.376 & [0.347, 0.404] \\
\texttt{GPT-oss-120B} & 0.633 & 0.625 & 0.619 & 0.620 & 0.624 & [0.593, 0.653] \\
\texttt{GPT-oss-20B} & 0.554 & 0.586 & 0.568 & 0.604 & 0.578 & [0.547, 0.607] \\
\rowcolor{teal!7}\texttt{InternLM2.5-7B} & 0.329 & 0.321 & 0.175 & 0.297 & 0.281 & [0.254, 0.307] \\
\rowcolor{teal!7}\texttt{InternLM2.5-1.8B} & 0.023 & -0.012 & -0.026 & -0.004 & -0.005 & [-0.018, 0.008] \\
\texttt{Llama3.1-70B} & 0.652 & 0.660 & 0.535 & 0.660 & 0.627 & [0.598, 0.653] \\
\texttt{Llama3.1-8B} & 0.331 & 0.408 & 0.170 & 0.322 & 0.308 & [0.281, 0.335] \\
\texttt{Llama3.2-1B} & 0.000 & 0.011 & 0.003 & 0.022 & 0.009 & [0.002, 0.016] \\
\rowcolor{teal!7}\texttt{Phi4-14B} & 0.598 & 0.627 & 0.561 & 0.610 & 0.599 & [0.569, 0.626] \\
\texttt{Qwen2-7B} & 0.348 & 0.358 & 0.222 & 0.384 & 0.328 & [0.300, 0.355] \\
\texttt{Qwen2-1.5B} & 0.072 & 0.022 & -0.002 & -0.018 & 0.019 & [0.004, 0.032] \\
\texttt{Qwen2-0.5B} & 0.036 & -0.027 & -0.003 & 0.005 & 0.003 & [-0.012, 0.019] \\
\rowcolor{teal!7}\texttt{Qwen2.5-7B} & 0.347 & 0.470 & 0.236 & 0.431 & 0.371 & [0.342, 0.401] \\
\rowcolor{teal!7}\texttt{Qwen2.5-1.5B} & 0.044 & 0.060 & 0.089 & 0.066 & 0.065 & [0.044, 0.087] \\
\texttt{Qwen3-14B} & 0.464 & 0.562 & 0.428 & 0.545 & 0.500 & [0.467, 0.530] \\
\texttt{Qwen3-8B} & 0.348 & 0.518 & 0.401 & 0.571 & 0.460 & [0.428, 0.489] \\
\texttt{Qwen3-4B} & 0.512 & 0.497 & 0.490 & 0.549 & 0.512 & [0.478, 0.544] \\
\texttt{Qwen3-0.6B} & 0.002 & 0.136 & -0.048 & 0.105 & 0.049 & [0.030, 0.066] \\
\bottomrule
\end{tabular}
\end{minipage}
\par\smallskip
\par{\scriptsize * The best results are highlighted in \textbf{bold} and the second-best results are \underline{underlined}.
Spearman's $\rho$ is reported on its original $[-1,1]$ scale and rounded to three decimal places.
The 95\% confidence intervals are for Avg. and are cluster-bootstrapped over scenario groups.
Strategy abbreviations: N = Direct, CoT = chain-of-thought, SR = Self-reflection, TG = Theory-guided.}
\end{table}

%% file: figure_and_tables/benchmark_basic_model.tex
% =============================================================================
% basic models
% =============================================================================
\begin{table}[!htbp]
\caption{Social attribution performance of basic baselines.}
\label{tab:basic_baselines}
\centering
\fontsize{6}{7.2}\selectfont
\setlength{\tabcolsep}{1.5pt}
\begin{minipage}[t]{0.49\linewidth}
\vspace{0pt}
\centering
\textbf{(a) Responsibility judgment}\par\smallskip
\begin{tabular}{lcccc}
\toprule
Baseline & Acc. & 95\% CI & $\rho$ & 95\% CI \\
\midrule
\texttt{Random Stratified} & 30.16 & [28.62, 31.68] & 0.028 & [-0.013, 0.068] \\
\texttt{Majority Class} & \underline{40.93} & [38.87, 42.99] & 0.000 & [0.000, 0.000] \\
\texttt{Heuristic Rule} & 21.51 & [19.82, 23.16] & 0.115 & [0.078, 0.149] \\
\texttt{Attribution Rule} & 33.52 & [31.52, 35.43] & \textbf{0.303} & [0.246, 0.353] \\
\texttt{mBERT Classifier} & \textbf{43.47} & [41.43, 45.32] & \underline{0.141} & [0.112, 0.170] \\
\bottomrule
\end{tabular}
\end{minipage}
\hfill
\begin{minipage}[t]{0.49\linewidth}
\vspace{0pt}
\centering
\textbf{(b) Blame judgment}\par\smallskip
\begin{tabular}{lcccc}
\toprule
Baseline & Acc. & 95\% CI & $\rho$ & 95\% CI \\
\midrule
\texttt{Random Stratified} & 27.48 & [25.89, 29.09] & -0.023 & [-0.063, 0.015] \\
\texttt{Majority Class} & 31.37 & [29.20, 33.39] & 0.000 & [0.000, 0.000] \\
\texttt{Heuristic Rule} & 26.89 & [25.07, 28.87] & \underline{0.138} & [0.106, 0.169] \\
\texttt{Attribution Rule} & \textbf{46.91} & [44.61, 49.39] & \textbf{0.368} & [0.314, 0.421] \\
\texttt{mBERT Classifier} & \underline{31.60} & [29.66, 33.60] & 0.069 & [0.042, 0.098] \\
\bottomrule
\end{tabular}
\end{minipage}
\par\smallskip
\par{\scriptsize The best results are highlighted in \textbf{bold} and the second-best results are \underline{underlined}.}
\end{table}

%% file: appendix_sections/valid_statistic.tex
We report the proportion of valid LLM outputs for each model, task, and prompt strategy.
An output is considered valid if the evaluator can extract one canonical answer label from \{\texttt{no}, \texttt{low}, \texttt{medium}, \texttt{high}\}.
Across all evaluated LLM outputs, 98.49\% are valid, and 1.51\% are invalid.
These invalid cases are excluded from the calculation of metrics.

\begin{table}[!htbp]
\caption{Valid output rates of evaluated LLMs by task and prompt strategy.}
\label{tab:valid_llm_outputs}
\centering
\fontsize{6}{7.2}\selectfont
\setlength{\tabcolsep}{1.5pt}
\begin{minipage}[t]{0.49\linewidth}
\vspace{0pt}
\centering
\textbf{(a) Responsibility judgment}\par\smallskip
\begin{tabular}{lccccc}
\toprule
 & \multicolumn{4}{c}{Prompt Strategy} & \\
\cmidrule(lr){2-5}
Model & N & CoT & SR & TG & Avg. \\
\midrule
\multicolumn{6}{c}{Closed-source Models} \\
\midrule
\rowcolor{teal!7}\texttt{Claude-Haiku-4.5} & 100.00 & 100.00 & 100.00 & 100.00 & 100.00 \\
\texttt{Doubao-Seed-1.8} & 100.00 & 100.00 & 100.00 & 100.00 & 100.00 \\
\rowcolor{teal!7}\texttt{Gemini-3.5-flash} & 99.90 & 100.00 & 100.00 & 100.00 & 99.97 \\
\texttt{GPT-5-mini} & 100.00 & 100.00 & 100.00 & 100.00 & 100.00 \\
\texttt{GPT-5-nano} & 100.00 & 99.12 & 92.94 & 98.50 & 97.64 \\
\rowcolor{teal!7}\texttt{Grok-4.3} & 100.00 & 100.00 & 100.00 & 100.00 & 100.00 \\
\texttt{Kimi-K2.5} & 100.00 & 100.00 & 100.00 & 100.00 & 100.00 \\
\rowcolor{teal!7}\texttt{MiMo-v2.5} & 100.00 & 100.00 & 100.00 & 100.00 & 100.00 \\
\texttt{Step-3.7-flash} & 100.00 & 97.87 & 99.92 & 96.54 & 98.58 \\
\midrule
\multicolumn{6}{c}{Open-source Models} \\
\midrule
\rowcolor{teal!7}\texttt{Deepseek-R1-70B} & 99.90 & 97.98 & 96.07 & 89.50 & 95.86 \\
\rowcolor{teal!7}\texttt{Deepseek-R1-1.5B} & 95.78 & 97.26 & 97.88 & 98.68 & 97.40 \\
\texttt{Gemma3-27B} & 100.00 & 100.00 & 100.00 & 100.00 & 100.00 \\
\texttt{Gemma3-4B} & 100.00 & 100.00 & 100.00 & 100.00 & 100.00 \\
\texttt{Gemma3-1B} & 100.00 & 99.92 & 89.99 & 99.79 & 97.43 \\
\rowcolor{teal!7}\texttt{Glm4-9B} & 100.00 & 100.00 & 100.00 & 100.00 & 100.00 \\
\texttt{GPT-oss-120B} & 99.97 & 100.00 & 99.97 & 100.00 & 99.99 \\
\texttt{GPT-oss-20B} & 100.00 & 99.84 & 99.87 & 99.97 & 99.92 \\
\rowcolor{teal!7}\texttt{InternLM2.5-7B} & 100.00 & 99.25 & 97.15 & 99.04 & 98.86 \\
\rowcolor{teal!7}\texttt{InternLM2.5-1.8B} & 100.00 & 91.38 & 89.52 & 90.74 & 92.91 \\
\texttt{Llama3.1-70B} & 100.00 & 99.33 & 97.57 & 99.79 & 99.17 \\
\texttt{Llama3.1-8B} & 100.00 & 87.83 & 86.27 & 91.85 & 91.49 \\
\texttt{Llama3.2-1B} & 100.00 & 91.45 & 93.79 & 86.94 & 93.05 \\
\rowcolor{teal!7}\texttt{Phi4-14B} & 100.00 & 100.00 & 100.00 & 99.97 & 99.99 \\
\texttt{Qwen2-7B} & 100.00 & 94.05 & 100.00 & 95.21 & 97.32 \\
\texttt{Qwen2-1.5B} & 100.00 & 66.91 & 99.09 & 96.58 & 90.65 \\
\texttt{Qwen2-0.5B} & 100.00 & 96.66 & 99.59 & 99.79 & 99.01 \\
\rowcolor{teal!7}\texttt{Qwen2.5-7B} & 100.00 & 100.00 & 99.84 & 100.00 & 99.96 \\
\rowcolor{teal!7}\texttt{Qwen2.5-1.5B} & 100.00 & 94.31 & 99.92 & 99.35 & 98.40 \\
\texttt{Qwen3-14B} & 100.00 & 100.00 & 100.00 & 100.00 & 100.00 \\
\texttt{Qwen3-8B} & 100.00 & 100.00 & 100.00 & 100.00 & 100.00 \\
\texttt{Qwen3-4B} & 100.00 & 100.00 & 100.00 & 100.00 & 100.00 \\
\texttt{Qwen3-0.6B} & 100.00 & 100.00 & 100.00 & 100.00 & 100.00 \\
\bottomrule
\end{tabular}
\end{minipage}
\hfill
\begin{minipage}[t]{0.49\linewidth}
\vspace{0pt}
\centering
\textbf{(b) Blame judgment}\par\smallskip
\begin{tabular}{lccccc}
\toprule
 & \multicolumn{4}{c}{Prompt Strategy} & \\
\cmidrule(lr){2-5}
Model & N & CoT & SR & TG & Avg. \\
\midrule
\multicolumn{6}{c}{Closed-source Models} \\
\midrule
\rowcolor{teal!7}\texttt{Claude-Haiku-4.5} & 100.00 & 100.00 & 100.00 & 100.00 & 100.00 \\
\texttt{Doubao-Seed-1.8} & 99.97 & 100.00 & 100.00 & 100.00 & 99.99 \\
\rowcolor{teal!7}\texttt{Gemini-3.5-flash} & 98.99 & 100.00 & 100.00 & 100.00 & 99.75 \\
\texttt{GPT-5-mini} & 100.00 & 100.00 & 100.00 & 100.00 & 100.00 \\
\texttt{GPT-5-nano} & 100.00 & 99.84 & 97.11 & 99.02 & 98.99 \\
\rowcolor{teal!7}\texttt{Grok-4.3} & 100.00 & 100.00 & 100.00 & 100.00 & 100.00 \\
\texttt{Kimi-K2.5} & 100.00 & 100.00 & 100.00 & 100.00 & 100.00 \\
\rowcolor{teal!7}\texttt{MiMo-v2.5} & 100.00 & 100.00 & 100.00 & 100.00 & 100.00 \\
\texttt{Step-3.7-flash} & 99.97 & 97.57 & 99.91 & 97.59 & 98.76 \\
\midrule
\multicolumn{6}{c}{Open-source Models} \\
\midrule
\rowcolor{teal!7}\texttt{Deepseek-R1-70B} & 99.84 & 99.39 & 97.43 & 90.99 & 96.91 \\
\rowcolor{teal!7}\texttt{Deepseek-R1-1.5B} & 96.24 & 96.24 & 97.56 & 97.83 & 96.97 \\
\texttt{Gemma3-27B} & 100.00 & 100.00 & 100.00 & 100.00 & 100.00 \\
\texttt{Gemma3-4B} & 100.00 & 100.00 & 100.00 & 100.00 & 100.00 \\
\texttt{Gemma3-1B} & 100.00 & 99.97 & 98.81 & 99.87 & 99.66 \\
\rowcolor{teal!7}\texttt{Glm4-9B} & 100.00 & 100.00 & 100.00 & 100.00 & 100.00 \\
\texttt{GPT-oss-120B} & 99.92 & 100.00 & 99.97 & 100.00 & 99.97 \\
\texttt{GPT-oss-20B} & 99.97 & 99.92 & 99.92 & 99.97 & 99.95 \\
\rowcolor{teal!7}\texttt{InternLM2.5-7B} & 100.00 & 98.22 & 92.29 & 97.85 & 97.09 \\
\rowcolor{teal!7}\texttt{InternLM2.5-1.8B} & 100.00 & 87.60 & 81.61 & 89.35 & 89.64 \\
\texttt{Llama3.1-70B} & 100.00 & 99.50 & 97.51 & 99.97 & 99.24 \\
\texttt{Llama3.1-8B} & 100.00 & 97.96 & 96.18 & 93.22 & 96.84 \\
\texttt{Llama3.2-1B} & 100.00 & 83.26 & 84.59 & 91.69 & 89.89 \\
\rowcolor{teal!7}\texttt{Phi4-14B} & 100.00 & 100.00 & 99.89 & 100.00 & 99.97 \\
\texttt{Qwen2-7B} & 100.00 & 98.20 & 100.00 & 97.64 & 98.96 \\
\texttt{Qwen2-1.5B} & 100.00 & 77.66 & 99.97 & 99.68 & 94.33 \\
\texttt{Qwen2-0.5B} & 100.00 & 95.95 & 99.95 & 99.92 & 98.95 \\
\rowcolor{teal!7}\texttt{Qwen2.5-7B} & 100.00 & 100.00 & 99.79 & 100.00 & 99.95 \\
\rowcolor{teal!7}\texttt{Qwen2.5-1.5B} & 100.00 & 99.13 & 99.95 & 99.95 & 99.75 \\
\texttt{Qwen3-14B} & 100.00 & 100.00 & 100.00 & 100.00 & 100.00 \\
\texttt{Qwen3-8B} & 100.00 & 100.00 & 100.00 & 100.00 & 100.00 \\
\texttt{Qwen3-4B} & 100.00 & 100.00 & 100.00 & 100.00 & 100.00 \\
\texttt{Qwen3-0.6B} & 100.00 & 99.95 & 100.00 & 100.00 & 99.99 \\
\bottomrule
\end{tabular}
\end{minipage}
\par\smallskip
\par{\scriptsize * Values are valid-output rates in percentages, pooled over two subsets and over English and Chinese items.
Strategy abbreviations: N = Direct, CoT = chain-of-thought, SR = Self-reflection, TG = Theory-guided.}
\end{table}

%% file: appendix_sections/error_breakdown.tex
\paragraph{Agent relations and dataset subsets.}
Tables~\ref{tab:error_breakdown_scenario_type_accuracy} and~\ref{tab:error_breakdown_scenario_type_spearman} break down \textit{Vignette} results by agent relation; the accuracy table also includes \textit{Reality}. Responsibility accuracy is highest on \textit{Reality} for all nine API-accessed LLMs, but the ordering of these groups changes across tasks. For GPT-5-mini, responsibility accuracy is 61.89\% on \textit{Reality}, compared with 51.88\% on vicarious-blame scenarios and 46.16\% on command chains. For blame, its highest score among these groups is instead on vicarious-blame scenarios (66.60\%), compared with 49.39\% on \textit{Reality} and 44.68\% on command chains. These differences support retaining complementary groups rather than treating aggregate agreement as a uniform measure across social relations. ``General'' denotes interactions without the designated supervisory or command relation.

\begin{table}[!htbp]
\caption{Accuracy by inter-agent relation within \textit{Vignette}, with \textit{Reality} shown separately.}
\label{tab:error_breakdown_scenario_type_accuracy}
\centering
\fontsize{6}{7.2}\selectfont
\setlength{\tabcolsep}{1.5pt}
\begin{minipage}[t]{0.49\linewidth}
\vspace{0pt}
\centering
\textbf{(a) Responsibility judgment}\par\smallskip
\begin{tabular}{lcccc}
\toprule
 & \multicolumn{3}{c}{Vignette agent relation} & \\
\cmidrule(lr){2-4}
Model & General & Vicarious & Commanding & Reality \\
\midrule
\multicolumn{5}{c}{Closed-source Models} \\
\midrule
\rowcolor{teal!7}\texttt{Claude-Haiku-4.5} & 55.74 & 42.92 & 41.44 & 58.98 \\
\texttt{Doubao-Seed-1.8} & 48.83 & 44.62 & 39.15 & 63.83 \\
\rowcolor{teal!7}\texttt{Gemini-3.5-flash} & 46.60 & 48.84 & 43.60 & 61.46 \\
\texttt{GPT-5-mini} & \underline{55.87} & 51.88 & 46.16 & 61.89 \\
\texttt{GPT-5-nano} & 48.16 & 40.14 & 41.11 & 59.41 \\
\rowcolor{teal!7}\texttt{Grok-4.3} & 51.80 & 39.25 & \underline{47.98} & \textbf{66.55} \\
\texttt{Kimi-K2.5} & 54.65 & 48.75 & 41.37 & 60.78 \\
\rowcolor{teal!7}\texttt{MiMo-v2.5} & \textbf{56.43} & 51.34 & 43.41 & 60.98 \\
\texttt{Step-3.7-flash} & 50.32 & 37.76 & 37.81 & \underline{66.15} \\
\midrule
\multicolumn{5}{c}{Open-source Models} \\
\midrule
\rowcolor{teal!7}\texttt{Deepseek-R1-70B} & 48.17 & 31.36 & 38.75 & 50.28 \\
\rowcolor{teal!7}\texttt{Deepseek-R1-1.5B} & 28.25 & 26.08 & 32.08 & 28.47 \\
\texttt{Gemma3-27B} & 47.92 & 50.09 & 39.56 & 55.43 \\
\texttt{Gemma3-4B} & 40.20 & 47.13 & 37.13 & 47.58 \\
\texttt{Gemma3-1B} & 29.61 & 33.42 & 31.47 & 33.84 \\
\rowcolor{teal!7}\texttt{Glm4-9B} & 42.17 & 33.24 & 39.42 & 42.51 \\
\texttt{GPT-oss-120B} & 53.72 & 42.74 & \textbf{50.67} & 57.64 \\
\texttt{GPT-oss-20B} & 52.50 & 40.23 & 47.24 & 54.37 \\
\rowcolor{teal!7}\texttt{InternLM2.5-7B} & 38.16 & 40.68 & 34.30 & 35.80 \\
\rowcolor{teal!7}\texttt{InternLM2.5-1.8B} & 26.91 & 37.01 & 30.86 & 34.43 \\
\texttt{Llama3.1-70B} & 50.81 & 38.08 & 47.04 & 58.59 \\
\texttt{Llama3.1-8B} & 38.23 & 34.59 & 35.51 & 39.41 \\
\texttt{Llama3.2-1B} & 16.74 & 8.15 & 21.36 & 9.99 \\
\rowcolor{teal!7}\texttt{Phi4-14B} & 52.96 & 34.59 & 43.26 & 56.57 \\
\texttt{Qwen2-7B} & 39.55 & 32.44 & 39.62 & 36.10 \\
\texttt{Qwen2-1.5B} & 24.47 & 22.40 & 29.78 & 20.58 \\
\texttt{Qwen2-0.5B} & 20.16 & 3.67 & 16.11 & 12.41 \\
\rowcolor{teal!7}\texttt{Qwen2.5-7B} & 43.25 & 33.87 & 40.70 & 43.06 \\
\rowcolor{teal!7}\texttt{Qwen2.5-1.5B} & 24.11 & 18.91 & 18.73 & 21.37 \\
\texttt{Qwen3-14B} & 50.59 & 41.31 & 43.94 & 54.87 \\
\texttt{Qwen3-8B} & 46.17 & 47.13 & 37.87 & 54.28 \\
\texttt{Qwen3-4B} & 45.86 & 44.44 & 36.66 & 55.36 \\
\texttt{Qwen3-0.6B} & 25.14 & 28.67 & 28.98 & 30.68 \\
\midrule
\multicolumn{5}{c}{Basic Baselines} \\
\midrule
\texttt{Random Stratified} & 27.72 & 34.77 & 31.00 & 35.06 \\
\texttt{Majority Class} & 34.94 & \underline{60.22} & 34.50 & 51.19 \\
\texttt{Heuristic Rule} & 21.52 & 9.68 & 32.61 & 20.62 \\
\texttt{Attribution Rule} & 35.33 & 20.07 & 17.25 & 43.20 \\
\texttt{mBERT Classifier} & 35.72 & \textbf{60.93} & 35.31 & 60.45 \\
\bottomrule
\end{tabular}
\end{minipage}
\hfill
\begin{minipage}[t]{0.49\linewidth}
\vspace{0pt}
\centering
\textbf{(b) Blame judgment}\par\smallskip
\begin{tabular}{lcccc}
\toprule
 & \multicolumn{3}{c}{Vignette agent relation} & \\
\cmidrule(lr){2-4}
Model & General & Vicarious & Commanding & Reality \\
\midrule
\multicolumn{5}{c}{Closed-source Models} \\
\midrule
\rowcolor{teal!7}\texttt{Claude-Haiku-4.5} & 54.03 & 53.34 & 42.55 & 48.79 \\
\texttt{Doubao-Seed-1.8} & 52.56 & 52.77 & 47.07 & 55.01 \\
\rowcolor{teal!7}\texttt{Gemini-3.5-flash} & 52.65 & 48.57 & \underline{57.71} & 54.47 \\
\texttt{GPT-5-mini} & 44.66 & \textbf{66.60} & 44.68 & 49.39 \\
\texttt{GPT-5-nano} & 43.44 & 58.49 & 41.89 & 49.57 \\
\rowcolor{teal!7}\texttt{Grok-4.3} & \underline{61.48} & 46.09 & \textbf{62.03} & \underline{60.33} \\
\texttt{Kimi-K2.5} & 55.36 & 59.35 & 49.00 & 52.12 \\
\rowcolor{teal!7}\texttt{MiMo-v2.5} & 49.03 & 64.03 & 42.72 & 49.21 \\
\texttt{Step-3.7-flash} & \textbf{62.82} & 37.95 & 50.74 & 56.87 \\
\midrule
\multicolumn{5}{c}{Open-source Models} \\
\midrule
\rowcolor{teal!7}\texttt{Deepseek-R1-70B} & 47.68 & 44.94 & 37.17 & 38.19 \\
\rowcolor{teal!7}\texttt{Deepseek-R1-1.5B} & 25.08 & 26.53 & 24.00 & 24.23 \\
\texttt{Gemma3-27B} & 40.99 & 53.34 & 36.04 & 46.82 \\
\texttt{Gemma3-4B} & 28.56 & 36.26 & 33.38 & 41.95 \\
\texttt{Gemma3-1B} & 24.26 & 29.87 & 26.99 & 28.86 \\
\rowcolor{teal!7}\texttt{Glm4-9B} & 36.89 & 45.42 & 30.92 & 36.06 \\
\texttt{GPT-oss-120B} & 45.97 & \underline{64.98} & 44.68 & 49.53 \\
\texttt{GPT-oss-20B} & 46.81 & 57.06 & 41.09 & 42.76 \\
\rowcolor{teal!7}\texttt{InternLM2.5-7B} & 33.00 & 35.31 & 21.21 & 27.85 \\
\rowcolor{teal!7}\texttt{InternLM2.5-1.8B} & 19.85 & 26.53 & 28.66 & 28.97 \\
\texttt{Llama3.1-70B} & 54.30 & 53.05 & 43.22 & 47.34 \\
\texttt{Llama3.1-8B} & 34.09 & 37.69 & 28.52 & 32.62 \\
\texttt{Llama3.2-1B} & 17.98 & 18.99 & 21.08 & 10.94 \\
\rowcolor{teal!7}\texttt{Phi4-14B} & 52.60 & 50.48 & 39.49 & 43.20 \\
\texttt{Qwen2-7B} & 35.02 & 40.74 & 26.46 & 31.72 \\
\texttt{Qwen2-1.5B} & 23.15 & 28.15 & 25.60 & 19.07 \\
\texttt{Qwen2-0.5B} & 25.87 & 21.47 & 25.80 & 15.29 \\
\rowcolor{teal!7}\texttt{Qwen2.5-7B} & 40.05 & 43.23 & 26.46 & 35.11 \\
\rowcolor{teal!7}\texttt{Qwen2.5-1.5B} & 28.53 & 26.05 & 22.87 & 25.82 \\
\texttt{Qwen3-14B} & 41.10 & 50.38 & 35.84 & 42.73 \\
\texttt{Qwen3-8B} & 39.05 & 51.24 & 36.97 & 43.96 \\
\texttt{Qwen3-4B} & 46.72 & 42.94 & 43.28 & 43.41 \\
\texttt{Qwen3-0.6B} & 26.81 & 27.86 & 27.86 & 27.21 \\
\midrule
\multicolumn{5}{c}{Basic Baselines} \\
\midrule
\texttt{Random Stratified} & 27.19 & 22.90 & 23.94 & 29.71 \\
\texttt{Majority Class} & 26.02 & 29.01 & 32.18 & 44.12 \\
\texttt{Heuristic Rule} & 31.84 & 15.65 & 25.80 & 20.74 \\
\texttt{Attribution Rule} & 46.38 & 29.01 & 50.00 & 50.73 \\
\texttt{mBERT Classifier} & 19.25 & 25.95 & 31.38 & \textbf{62.92} \\
\bottomrule
\end{tabular}
\end{minipage}
\par\smallskip
\par{\fontsize{7}{9}\selectfont * The best results are highlighted in \textbf{bold} and the second-best results are \underline{underlined}.
LLM cells are averages over N, CoT, SR, and TG.
Accuracy is reported as a percentage and rounded to two decimal places.
Accuracy is reported separately for \textit{Vignette} agent relations and \textit{Reality}.}
\end{table}

\begin{table}[!htbp]
\caption{Spearman $\rho$ by inter-agent relation within \textit{Vignette}.}
\label{tab:error_breakdown_scenario_type_spearman}
\centering
\fontsize{6}{7.2}\selectfont
\setlength{\tabcolsep}{1.5pt}
\begin{minipage}[t]{0.49\linewidth}
\vspace{0pt}
\centering
\textbf{(a) Responsibility judgment}\par\smallskip
\begin{tabular}{lccc}
\toprule
 & \multicolumn{3}{c}{Vignette agent relation} \\
\cmidrule(lr){2-4}
Model & General & Vicarious & Commanding \\
\midrule
\multicolumn{4}{c}{Closed-source Models} \\
\midrule
\rowcolor{teal!7}\texttt{Claude-Haiku-4.5} & 0.661 & 0.256 & 0.258 \\
\texttt{Doubao-Seed-1.8} & 0.646 & 0.252 & 0.275 \\
\rowcolor{teal!7}\texttt{Gemini-3.5-flash} & 0.627 & \underline{0.324} & 0.299 \\
\texttt{GPT-5-mini} & 0.645 & \textbf{0.334} & 0.303 \\
\texttt{GPT-5-nano} & 0.496 & 0.176 & 0.211 \\
\rowcolor{teal!7}\texttt{Grok-4.3} & \underline{0.665} & 0.231 & \underline{0.320} \\
\texttt{Kimi-K2.5} & 0.637 & 0.210 & 0.232 \\
\rowcolor{teal!7}\texttt{MiMo-v2.5} & 0.664 & 0.210 & 0.280 \\
\texttt{Step-3.7-flash} & \textbf{0.676} & 0.237 & 0.318 \\
\midrule
\multicolumn{4}{c}{Open-source Models} \\
\midrule
\rowcolor{teal!7}\texttt{Deepseek-R1-70B} & 0.538 & 0.132 & 0.176 \\
\rowcolor{teal!7}\texttt{Deepseek-R1-1.5B} & 0.132 & -0.034 & 0.012 \\
\texttt{Gemma3-27B} & 0.612 & 0.141 & 0.242 \\
\texttt{Gemma3-4B} & 0.446 & 0.061 & 0.063 \\
\texttt{Gemma3-1B} & 0.105 & 0.000 & 0.012 \\
\rowcolor{teal!7}\texttt{Glm4-9B} & 0.481 & 0.011 & 0.134 \\
\texttt{GPT-oss-120B} & 0.609 & 0.147 & \textbf{0.322} \\
\texttt{GPT-oss-20B} & 0.555 & 0.143 & 0.300 \\
\rowcolor{teal!7}\texttt{InternLM2.5-7B} & 0.401 & 0.006 & 0.041 \\
\rowcolor{teal!7}\texttt{InternLM2.5-1.8B} & -0.022 & 0.055 & 0.017 \\
\texttt{Llama3.1-70B} & 0.617 & 0.258 & 0.258 \\
\texttt{Llama3.1-8B} & 0.325 & -0.021 & 0.065 \\
\texttt{Llama3.2-1B} & 0.019 & 0.006 & 0.020 \\
\rowcolor{teal!7}\texttt{Phi4-14B} & 0.639 & 0.158 & 0.262 \\
\texttt{Qwen2-7B} & 0.403 & 0.044 & 0.090 \\
\texttt{Qwen2-1.5B} & 0.011 & 0.027 & -0.006 \\
\texttt{Qwen2-0.5B} & 0.001 & -0.006 & 0.018 \\
\rowcolor{teal!7}\texttt{Qwen2.5-7B} & 0.426 & 0.019 & 0.085 \\
\rowcolor{teal!7}\texttt{Qwen2.5-1.5B} & 0.070 & 0.025 & -0.014 \\
\texttt{Qwen3-14B} & 0.564 & 0.078 & 0.229 \\
\texttt{Qwen3-8B} & 0.563 & 0.057 & 0.160 \\
\texttt{Qwen3-4B} & 0.523 & 0.197 & 0.224 \\
\texttt{Qwen3-0.6B} & 0.078 & -0.008 & 0.070 \\
\midrule
\multicolumn{4}{c}{Basic Baselines} \\
\midrule
\texttt{Random Stratified} & 0.017 & 0.025 & 0.049 \\
\texttt{Majority Class} & 0.000 & 0.000 & 0.000 \\
\texttt{Heuristic Rule} & 0.176 & -0.084 & 0.046 \\
\texttt{Attribution Rule} & 0.473 & -0.252 & 0.158 \\
\texttt{mBERT Classifier} & 0.186 & 0.025 & 0.038 \\
\bottomrule
\end{tabular}
\end{minipage}
\hfill
\begin{minipage}[t]{0.49\linewidth}
\vspace{0pt}
\centering
\textbf{(b) Blame judgment}\par\smallskip
\begin{tabular}{lccc}
\toprule
 & \multicolumn{3}{c}{Vignette agent relation} \\
\cmidrule(lr){2-4}
Model & General & Vicarious & Commanding \\
\midrule
\multicolumn{4}{c}{Closed-source Models} \\
\midrule
\rowcolor{teal!7}\texttt{Claude-Haiku-4.5} & \underline{0.726} & 0.450 & 0.506 \\
\texttt{Doubao-Seed-1.8} & 0.715 & 0.495 & 0.495 \\
\rowcolor{teal!7}\texttt{Gemini-3.5-flash} & 0.686 & 0.407 & \underline{0.579} \\
\texttt{GPT-5-mini} & 0.719 & \textbf{0.500} & 0.537 \\
\texttt{GPT-5-nano} & 0.557 & 0.423 & 0.452 \\
\rowcolor{teal!7}\texttt{Grok-4.3} & \textbf{0.757} & 0.484 & \textbf{0.651} \\
\texttt{Kimi-K2.5} & 0.715 & \underline{0.498} & 0.546 \\
\rowcolor{teal!7}\texttt{MiMo-v2.5} & 0.660 & 0.461 & 0.489 \\
\texttt{Step-3.7-flash} & 0.700 & 0.459 & 0.544 \\
\midrule
\multicolumn{4}{c}{Open-source Models} \\
\midrule
\rowcolor{teal!7}\texttt{Deepseek-R1-70B} & 0.544 & 0.314 & 0.331 \\
\rowcolor{teal!7}\texttt{Deepseek-R1-1.5B} & 0.068 & -0.018 & 0.003 \\
\texttt{Gemma3-27B} & 0.631 & 0.332 & 0.372 \\
\texttt{Gemma3-4B} & 0.325 & 0.102 & 0.109 \\
\texttt{Gemma3-1B} & 0.051 & -0.011 & 0.021 \\
\rowcolor{teal!7}\texttt{Glm4-9B} & 0.437 & 0.264 & 0.215 \\
\texttt{GPT-oss-120B} & 0.699 & 0.443 & 0.460 \\
\texttt{GPT-oss-20B} & 0.641 & 0.391 & 0.470 \\
\rowcolor{teal!7}\texttt{InternLM2.5-7B} & 0.355 & 0.166 & 0.071 \\
\rowcolor{teal!7}\texttt{InternLM2.5-1.8B} & -0.006 & 0.020 & -0.017 \\
\texttt{Llama3.1-70B} & 0.702 & 0.436 & 0.471 \\
\texttt{Llama3.1-8B} & 0.344 & 0.216 & 0.233 \\
\texttt{Llama3.2-1B} & 0.009 & 0.002 & 0.016 \\
\rowcolor{teal!7}\texttt{Phi4-14B} & 0.677 & 0.402 & 0.437 \\
\texttt{Qwen2-7B} & 0.398 & 0.191 & 0.154 \\
\texttt{Qwen2-1.5B} & 0.019 & 0.032 & 0.005 \\
\texttt{Qwen2-0.5B} & -0.001 & -0.011 & 0.027 \\
\rowcolor{teal!7}\texttt{Qwen2.5-7B} & 0.454 & 0.223 & 0.151 \\
\rowcolor{teal!7}\texttt{Qwen2.5-1.5B} & 0.089 & 0.018 & 0.005 \\
\texttt{Qwen3-14B} & 0.568 & 0.325 & 0.358 \\
\texttt{Qwen3-8B} & 0.537 & 0.309 & 0.263 \\
\texttt{Qwen3-4B} & 0.550 & 0.425 & 0.425 \\
\texttt{Qwen3-0.6B} & 0.057 & 0.009 & 0.046 \\
\midrule
\multicolumn{4}{c}{Basic Baselines} \\
\midrule
\texttt{Random Stratified} & -0.011 & -0.109 & -0.048 \\
\texttt{Majority Class} & 0.000 & 0.000 & 0.000 \\
\texttt{Heuristic Rule} & 0.156 & 0.027 & 0.219 \\
\texttt{Attribution Rule} & 0.506 & 0.182 & 0.211 \\
\texttt{mBERT Classifier} & 0.101 & -0.018 & 0.004 \\
\bottomrule
\end{tabular}
\end{minipage}
\par\smallskip
\par{\fontsize{7}{9}\selectfont * The best results are highlighted in \textbf{bold} and the second-best results are \underline{underlined}.
LLM cells are averages over N, CoT, SR, and TG.
Spearman's $\rho$ is reported on its original $[-1,1]$ scale and rounded to three decimal places.
Spearman's $\rho$ uses vignette data only.}
\end{table}

\clearpage
\paragraph{Attribution-dimension values.}
Table~\ref{tab:error_breakdown_dimension_accuracy} shows that accuracy also varies across dimension-value groups. For GPT-5-mini, responsibility accuracy is 62.82\% when intention is present versus 42.01\% when absent, and 65.78\% versus 47.29\% for foreknowledge. The pattern is not universal: Grok-4.3 has higher blame accuracy when intention is absent (61.89\%) than when present (59.23\%). These observational differences concern agreement within groups that can differ in scenario composition and human-label distributions. They do not measure a dimension's causal effect or show that a model necessarily assigns higher judgments when the dimension is present. This distinction motivates the controlled representation interventions in the main analysis.

\begin{table}[!htbp]
\caption{Accuracy of tested models by attribution dimension values.}
\label{tab:error_breakdown_dimension_accuracy}
\centering
\fontsize{6}{7.2}\selectfont
\setlength{\tabcolsep}{1.0pt}
\begin{tabular}{@{}lcccccccccc@{\hspace{5pt}}cccccccccc@{}}
\toprule
 & \multicolumn{10}{c}{\textbf{(a) Responsibility judgment}} & \multicolumn{10}{c}{\textbf{(b) Blame judgment}} \\
\cmidrule(lr){2-11}\cmidrule(lr){12-21}
 & \multicolumn{2}{c}{Intent.} & \multicolumn{2}{c}{Volunt.} & \multicolumn{2}{c}{Foreknow.} & \multicolumn{2}{c}{Control.} & \multicolumn{2}{c}{Obligation} & \multicolumn{2}{c}{Intent.} & \multicolumn{2}{c}{Volunt.} & \multicolumn{2}{c}{Foreknow.} & \multicolumn{2}{c}{Control.} & \multicolumn{2}{c}{Obligation} \\
\cmidrule(lr){2-3}\cmidrule(lr){4-5}\cmidrule(lr){6-7}\cmidrule(lr){8-9}\cmidrule(lr){10-11}\cmidrule(lr){12-13}\cmidrule(lr){14-15}\cmidrule(lr){16-17}\cmidrule(lr){18-19}\cmidrule(lr){20-21}
Model & 0 & 1 & 0 & 1 & 0 & 1 & 0 & 1 & 0 & 1 & 0 & 1 & 0 & 1 & 0 & 1 & 0 & 1 & 0 & 1 \\
\midrule
\multicolumn{21}{c}{Closed-source Models} \\
\midrule
\rowcolor{teal!7}\texttt{Claude-Haiku-4.5} & 44.55 & 58.68 & 53.42 & 66.64 & 46.43 & 62.27 & \textbf{48.44} & 58.47 & 56.89 & 55.00 & 45.58 & 53.46 & 49.73 & 54.69 & 43.45 & 58.55 & 46.02 & 54.62 & 56.23 & 51.69 \\
\texttt{Doubao-Seed-1.8} & 41.92 & 57.62 & 54.33 & 67.43 & 43.53 & 62.06 & 40.53 & 60.93 & 56.84 & 52.10 & 49.60 & 54.40 & \underline{57.48} & 57.03 & 45.75 & 59.83 & 47.69 & 56.26 & 55.51 & 51.88 \\
\rowcolor{teal!7}\texttt{Gemini-3.5-flash} & 41.10 & 56.30 & 52.86 & 63.98 & 40.59 & 62.52 & 41.40 & 58.49 & 54.46 & 53.86 & 49.98 & 55.09 & 54.00 & 58.12 & 50.96 & 56.04 & 50.15 & 55.70 & 55.45 & 54.17 \\
\texttt{GPT-5-mini} & 42.01 & \textbf{62.82} & \textbf{59.22} & \textbf{70.13} & 47.29 & \textbf{65.78} & \underline{48.07} & \underline{62.32} & 50.31 & \textbf{62.80} & 33.13 & 54.33 & 56.19 & 55.72 & 35.50 & 59.89 & 36.98 & 54.87 & 44.01 & 48.92 \\
\texttt{GPT-5-nano} & 44.23 & 53.79 & 56.04 & 60.17 & 44.88 & 56.79 & 44.86 & 54.89 & 53.79 & 53.19 & 35.25 & 51.53 & 52.86 & 54.33 & 34.36 & 58.39 & 37.09 & 52.65 & 45.62 & 48.85 \\
\rowcolor{teal!7}\texttt{Grok-4.3} & \textbf{49.50} & 58.56 & 56.60 & 66.63 & \textbf{48.43} & 62.98 & 46.63 & 61.75 & \textbf{61.35} & 55.34 & \underline{61.89} & \textbf{59.23} & \textbf{60.27} & \textbf{62.87} & \underline{59.11} & \underline{60.91} & \underline{58.49} & \textbf{60.99} & \textbf{61.86} & \textbf{59.36} \\
\texttt{Kimi-K2.5} & 42.58 & 60.27 & 55.96 & 68.38 & 44.88 & 64.82 & 46.14 & 60.83 & 57.11 & 54.99 & 44.86 & \underline{57.57} & 57.09 & 59.63 & 45.18 & \textbf{61.96} & 46.38 & \underline{58.39} & 53.79 & 54.91 \\
\rowcolor{teal!7}\texttt{MiMo-v2.5} & 44.64 & \underline{61.24} & 57.08 & 68.57 & 46.55 & \underline{65.73} & 46.74 & \textbf{62.53} & 57.71 & \underline{57.71} & 41.81 & 52.80 & 56.11 & 52.51 & 38.08 & 60.40 & 42.05 & 54.18 & 50.88 & 50.57 \\
\texttt{Step-3.7-flash} & \underline{48.48} & 57.80 & 55.30 & \underline{70.11} & 45.39 & 63.13 & 47.38 & 59.71 & \underline{58.95} & 53.31 & \textbf{64.15} & 54.73 & 54.62 & \underline{59.98} & 57.85 & 57.57 & \textbf{61.58} & 55.25 & 60.71 & \underline{56.87} \\
\midrule
\multicolumn{21}{c}{Open-source Models} \\
\midrule
\rowcolor{teal!7}\texttt{Deepseek-R1-70B} & 40.68 & 49.29 & 46.66 & 53.16 & 43.21 & 50.20 & 41.36 & 50.22 & 50.59 & 47.27 & 39.79 & 44.12 & 40.87 & 43.83 & 38.38 & 47.06 & 38.83 & 45.33 & 43.61 & 43.34 \\
\rowcolor{teal!7}\texttt{Deepseek-R1-1.5B} & 26.45 & 29.38 & 32.11 & 27.56 & 28.33 & 28.74 & 27.40 & 29.25 & 26.41 & 30.60 & 17.58 & 27.81 & 27.86 & 26.39 & 20.68 & 28.63 & 18.49 & 28.67 & 24.41 & 24.59 \\
\texttt{Gemma3-27B} & 34.43 & 56.30 & 54.97 & 63.23 & 43.15 & 56.45 & 41.34 & 55.44 & 41.25 & 56.96 & 21.74 & 52.89 & 52.66 & 53.13 & 27.84 & 58.58 & 29.29 & 52.51 & 39.38 & 46.48 \\
\texttt{Gemma3-4B} & 28.05 & 49.17 & 48.01 & 53.53 & 33.18 & 52.38 & 34.78 & 48.30 & 33.57 & 49.52 & 10.78 & 44.87 & 43.93 & 49.49 & 19.92 & 48.73 & 17.61 & 45.34 & 29.69 & 35.89 \\
\texttt{Gemma3-1B} & 23.67 & 34.81 & 35.73 & 35.72 & 27.60 & 35.36 & 24.88 & 35.80 & 26.78 & 34.86 & 17.16 & 30.74 & 31.82 & 30.26 & 20.32 & 32.74 & 18.84 & 31.58 & 24.20 & 28.55 \\
\rowcolor{teal!7}\texttt{Glm4-9B} & 36.76 & 43.24 & 45.35 & 44.33 & 39.59 & 43.06 & 38.95 & 42.90 & 41.45 & 43.26 & 26.79 & 40.75 & 42.75 & 39.79 & 28.77 & 44.01 & 30.56 & 40.33 & 34.20 & 38.11 \\
\texttt{GPT-oss-120B} & 43.71 & 58.23 & \underline{58.07} & 67.26 & \underline{47.66} & 60.05 & 48.04 & 57.82 & 53.09 & 52.16 & 37.69 & 53.14 & 54.51 & 54.52 & 40.00 & 56.64 & 37.35 & 55.50 & 46.02 & 51.57 \\
\texttt{GPT-oss-20B} & 44.41 & 54.76 & 54.97 & 60.89 & 45.31 & 57.83 & 47.01 & 54.77 & 49.34 & 54.08 & 40.50 & 47.50 & 50.98 & 46.58 & 37.77 & 52.69 & 40.03 & 48.77 & 45.91 & 46.90 \\
\rowcolor{teal!7}\texttt{InternLM2.5-7B} & 32.67 & 38.88 & 40.74 & 40.61 & 34.57 & 39.45 & 36.67 & 37.37 & 34.02 & 37.54 & 23.78 & 32.69 & 36.44 & 28.35 & 22.86 & 36.87 & 27.44 & 31.64 & 29.13 & 31.02 \\
\rowcolor{teal!7}\texttt{InternLM2.5-1.8B} & 17.94 & 35.94 & 35.17 & 38.08 & 23.99 & 37.12 & 21.04 & 36.83 & 25.68 & 33.90 & 10.47 & 30.72 & 26.41 & 35.09 & 17.59 & 31.41 & 11.98 & 32.59 & 22.40 & 25.22 \\
\texttt{Llama3.1-70B} & 43.03 & 56.15 & 55.17 & 60.68 & 46.04 & 58.31 & 44.38 & 57.36 & 55.33 & 55.02 & 47.23 & 51.83 & 49.29 & 53.22 & 45.94 & 54.76 & 47.78 & 52.13 & 50.98 & 52.72 \\
\texttt{Llama3.1-8B} & 33.28 & 40.09 & 41.77 & 43.39 & 35.14 & 40.94 & 35.89 & 39.54 & 34.96 & 39.57 & 27.64 & 35.61 & 38.48 & 34.20 & 26.63 & 39.52 & 31.44 & 34.35 & 30.62 & 36.27 \\
\texttt{Llama3.2-1B} & 16.40 & 13.14 & 12.36 & 10.94 & 17.90 & 10.52 & 16.98 & 12.27 & 14.63 & 13.68 & 17.09 & 15.08 & 12.42 & 12.96 & 19.01 & 12.50 & 15.17 & 16.00 & 15.61 & 15.01 \\
\rowcolor{teal!7}\texttt{Phi4-14B} & 46.13 & 54.40 & 53.62 & 60.24 & 46.08 & 57.58 & 47.80 & 54.66 & 56.09 & 51.33 & 46.01 & 48.23 & 49.69 & 47.87 & 41.45 & 53.40 & 46.32 & 48.35 & 48.58 & 48.02 \\
\texttt{Qwen2-7B} & 32.13 & 40.03 & 40.54 & 41.30 & 35.04 & 40.30 & 35.72 & 39.03 & 34.34 & 38.22 & 23.38 & 37.54 & 37.66 & 35.11 & 25.52 & 40.73 & 27.50 & 36.94 & 31.50 & 34.65 \\
\texttt{Qwen2-1.5B} & 22.76 & 23.65 & 24.80 & 21.72 & 24.76 & 22.13 & 24.02 & 23.01 & 21.97 & 23.12 & 19.13 & 23.49 & 21.75 & 21.27 & 21.95 & 22.41 & 19.18 & 24.06 & 19.64 & 21.27 \\
\texttt{Qwen2-0.5B} & 25.05 & 11.99 & 9.22 & 8.25 & 21.48 & 10.37 & 22.68 & 11.42 & 23.42 & 13.77 & 31.11 & 17.44 & 14.54 & 14.60 & 30.93 & 12.56 & 28.20 & 17.36 & 24.77 & 18.52 \\
\rowcolor{teal!7}\texttt{Qwen2.5-7B} & 34.57 & 45.34 & 46.98 & 48.15 & 36.86 & 47.28 & 40.61 & 43.28 & 39.86 & 43.15 & 28.01 & 40.89 & 41.77 & 38.36 & 29.33 & 44.38 & 32.33 & 39.98 & 36.06 & 39.16 \\
\rowcolor{teal!7}\texttt{Qwen2.5-1.5B} & 28.62 & 19.64 & 19.08 & 18.38 & 23.89 & 20.64 & 25.61 & 20.10 & 26.78 & 20.37 & 32.15 & 24.46 & 26.33 & 23.09 & 29.18 & 24.46 & 31.27 & 23.94 & 28.44 & 25.87 \\
\texttt{Qwen3-14B} & 36.81 & 56.49 & 54.29 & 61.58 & 43.39 & 57.80 & 41.83 & 56.45 & 50.02 & 52.95 & 32.26 & 45.93 & 49.29 & 45.48 & 30.66 & 52.50 & 35.59 & 45.75 & 39.96 & 44.57 \\
\texttt{Qwen3-8B} & 35.63 & 53.56 & 51.55 & 59.99 & 38.65 & 57.53 & 39.24 & 54.13 & 44.55 & 52.10 & 27.21 & 47.68 & 49.02 & 47.54 & 28.98 & 53.56 & 27.74 & 50.20 & 39.77 & 42.69 \\
\texttt{Qwen3-4B} & 42.13 & 50.89 & 49.32 & 58.31 & 37.37 & 58.62 & 40.05 & 53.54 & 49.51 & 50.06 & 45.11 & 44.74 & 46.20 & 49.93 & 39.02 & 50.41 & 43.53 & 45.68 & 45.24 & 42.69 \\
\texttt{Qwen3-0.6B} & 28.42 & 27.56 & 29.85 & 27.55 & 27.78 & 27.83 & 26.89 & 28.37 & 29.10 & 29.03 & 27.44 & 27.01 & 24.80 & 29.81 & 27.59 & 26.71 & 25.65 & 28.07 & 27.56 & 27.13 \\
\midrule
\multicolumn{21}{c}{Basic Baselines} \\
\midrule
\texttt{Random Stratified} & 20.27 & 35.65 & 34.66 & 38.86 & 27.66 & 34.60 & 25.10 & 35.06 & 26.15 & 34.56 & 23.12 & 29.41 & 27.27 & 30.95 & 27.74 & 27.33 & 25.62 & 28.73 & 28.67 & 28.79 \\
\texttt{Majority Class} & 14.75 & 53.91 & 49.28 & 62.91 & 24.87 & 59.31 & 20.70 & 56.35 & 25.82 & 51.03 & 12.31 & 42.87 & 36.52 & 49.60 & 34.69 & 32.82 & 22.45 & 40.78 & 37.89 & 38.53 \\
\texttt{Heuristic Rule} & 33.39 & 16.59 & 21.14 & 17.60 & 24.06 & 18.92 & 28.69 & 16.88 & 28.85 & 15.81 & 50.66 & 15.31 & 22.10 & 11.87 & 36.21 & 16.05 & 46.69 & 12.88 & 33.61 & 25.32 \\
\texttt{Attribution Rule} & 32.49 & 36.56 & 5.09 & 55.71 & 42.05 & 29.21 & 25.78 & 41.35 & 53.28 & 22.97 & 58.90 & 42.19 & 25.86 & 57.90 & \textbf{60.80} & 34.21 & 49.31 & 45.87 & \underline{61.19} & 37.16 \\
\texttt{mBERT Classifier} & 24.36 & 50.59 & 44.59 & 57.14 & 30.85 & 55.65 & 27.74 & 54.71 & 31.91 & 47.98 & 27.66 & 33.18 & 31.17 & 39.75 & 29.14 & 34.03 & 20.13 & 39.83 & 45.61 & 30.77 \\
\bottomrule
\end{tabular}
\par\smallskip
\par{\fontsize{7}{9}\selectfont * The best results are highlighted in \textbf{bold} and the second-best results are \underline{underlined}.
LLM cells are averages over N, CoT, SR, and TG.
Accuracy is reported as a percentage and rounded to two decimal places.
Accuracy pools vignette and reality data; questions with unspecified or inapplicable dimension labels are excluded from the corresponding 0/1 columns.}
\end{table}

%% file: appendix_sections/decodability_results.tex
\section{Additional Decodability Analyses}
\label{app:additional_decodability}

The main decodability analysis asks whether theory-derived dimensions can be recovered from hidden states. Above-chance accuracy alone does not establish what information a probe uses or whether that information is specific to the final judgment stage. We therefore examine controls for alternative predictive signals and compare two token positions.

\subsection{Decodability Sanity Checks}
\label{app:decodability_sanity}
\label{app:sanity_check_step1}

\input{appendix_sections/sanity_check_step1}

\clearpage
\subsection{Decodability at the Last Input Token}
\label{app:early_token_decodability}
\label{app:early_token_step1}

\input{appendix_sections/early_token_step1}

%% file: appendix_sections/sanity_check_step1.tex
The controls target three distinct concerns about the probing measurement. Shuffling training labels tests dependence on the original dimension--representation association; random projection tests how much decodability is retained after compressing the representation to 64 features; and scenario-text TF-IDF measures the information available from lexical surface cues. All use the same data splits, classifier family, and balanced-accuracy metric as the main probes (Appendix~\ref{app:probe_training}): eight folds holding out one scenario type at a time for \textit{Vignette}, and ten random folds for \textit{Reality}. Together with diversified vignette wording, holding out entire scenario types reduces opportunities to exploit familiar event settings; the controls further assess dependence on the tested representation and surface features.

\begin{figure}[!htbp]
  \centering
  \includegraphics[width=1.0\linewidth]{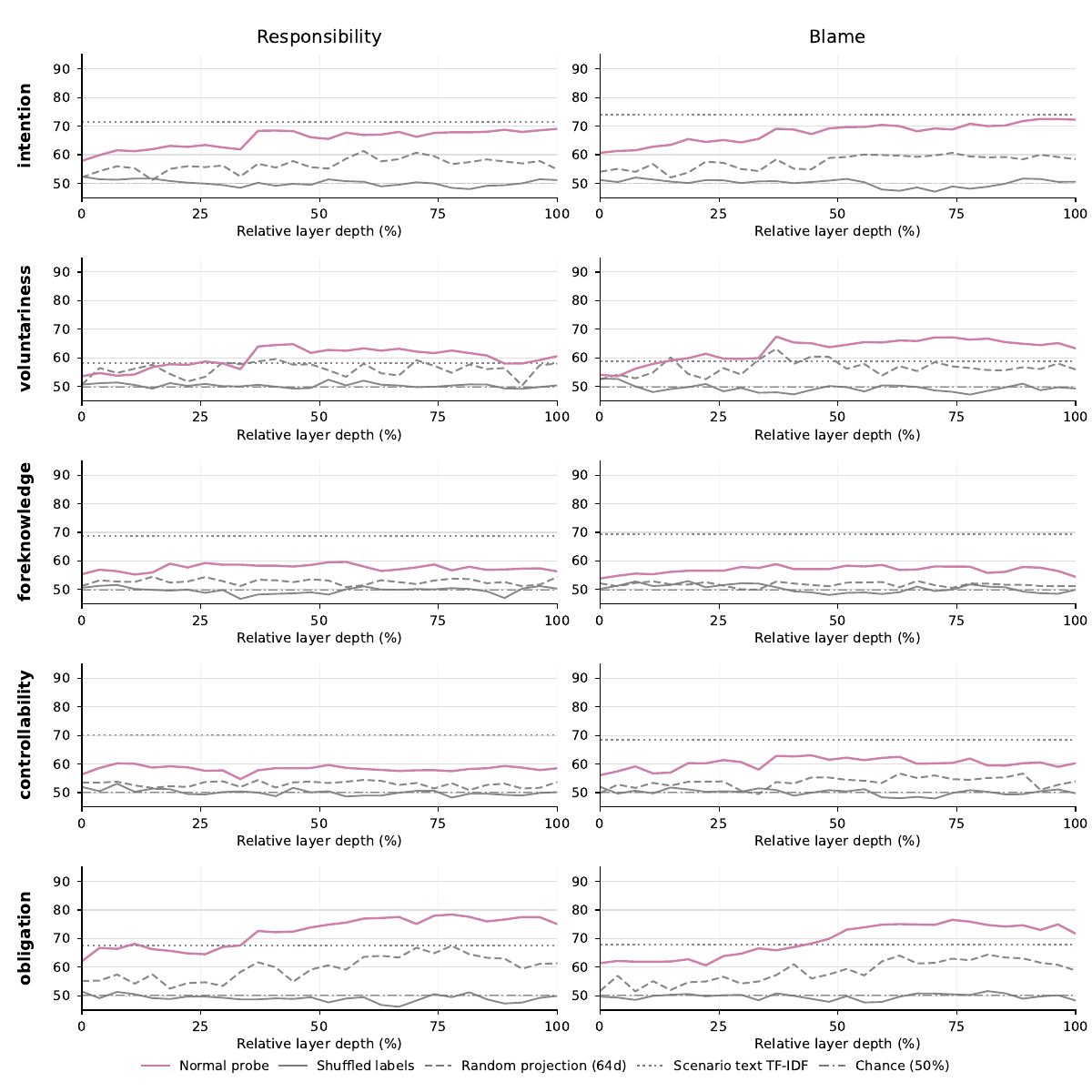}
  \caption{Sanity-check results at the label-token prediction step for DeepSeek-R1-1.5B, including shuffled-label, 64-dimensional random-projection, scenario-text lexical, and chance controls.}
  \label{fig:app-step1-sanity-deepseek-r1-15b}
\end{figure}

\begin{figure}[!htbp]
  \centering
  \includegraphics[width=1.0\linewidth]{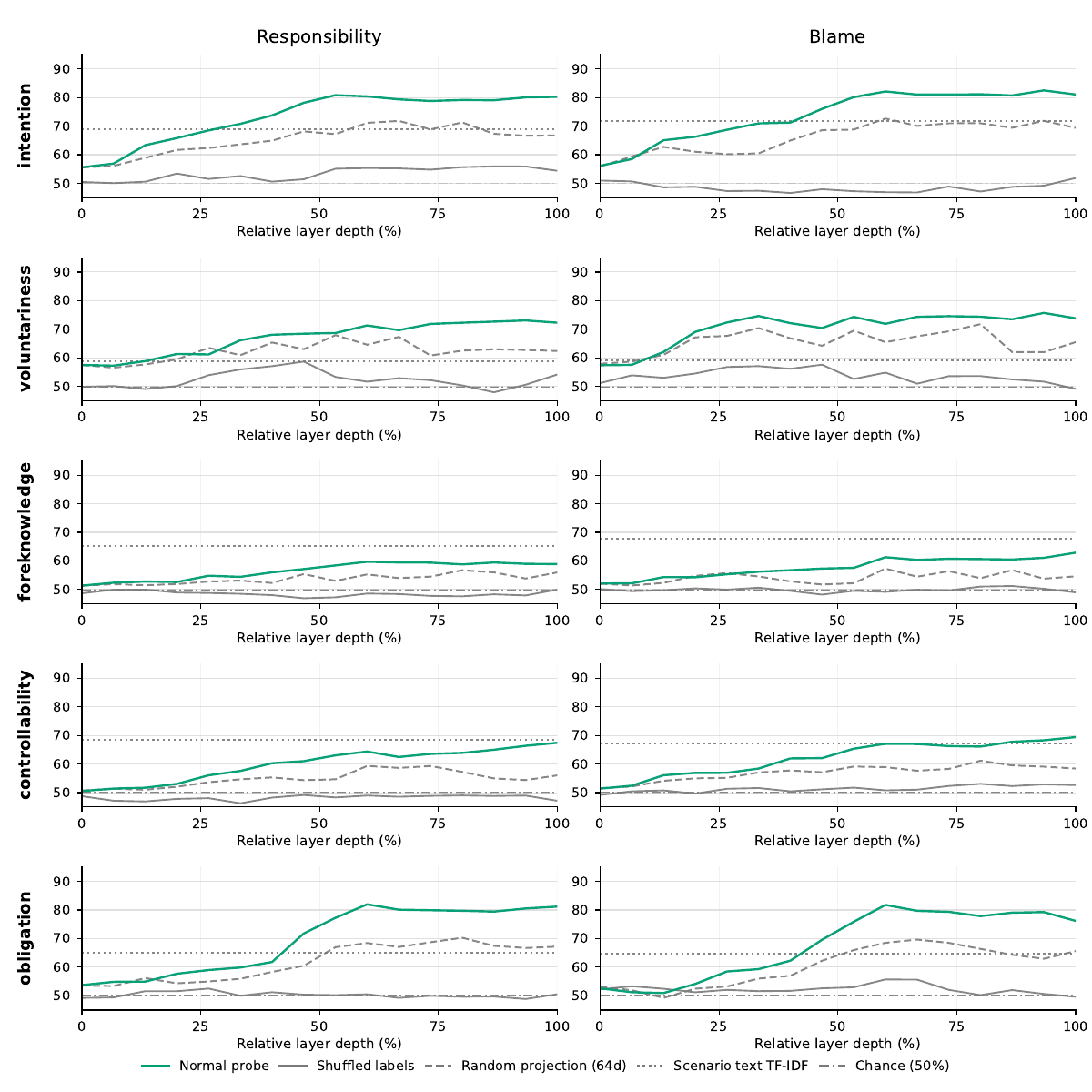}
  \caption{Sanity-check results at the label-token prediction step for Llama-3.2-1B, including shuffled-label, 64-dimensional random-projection, scenario-text lexical, and chance controls.}
  \label{fig:app-step1-sanity-llama32-1b}
\end{figure}

\begin{figure}[!htbp]
  \centering
  \includegraphics[width=1.0\linewidth]{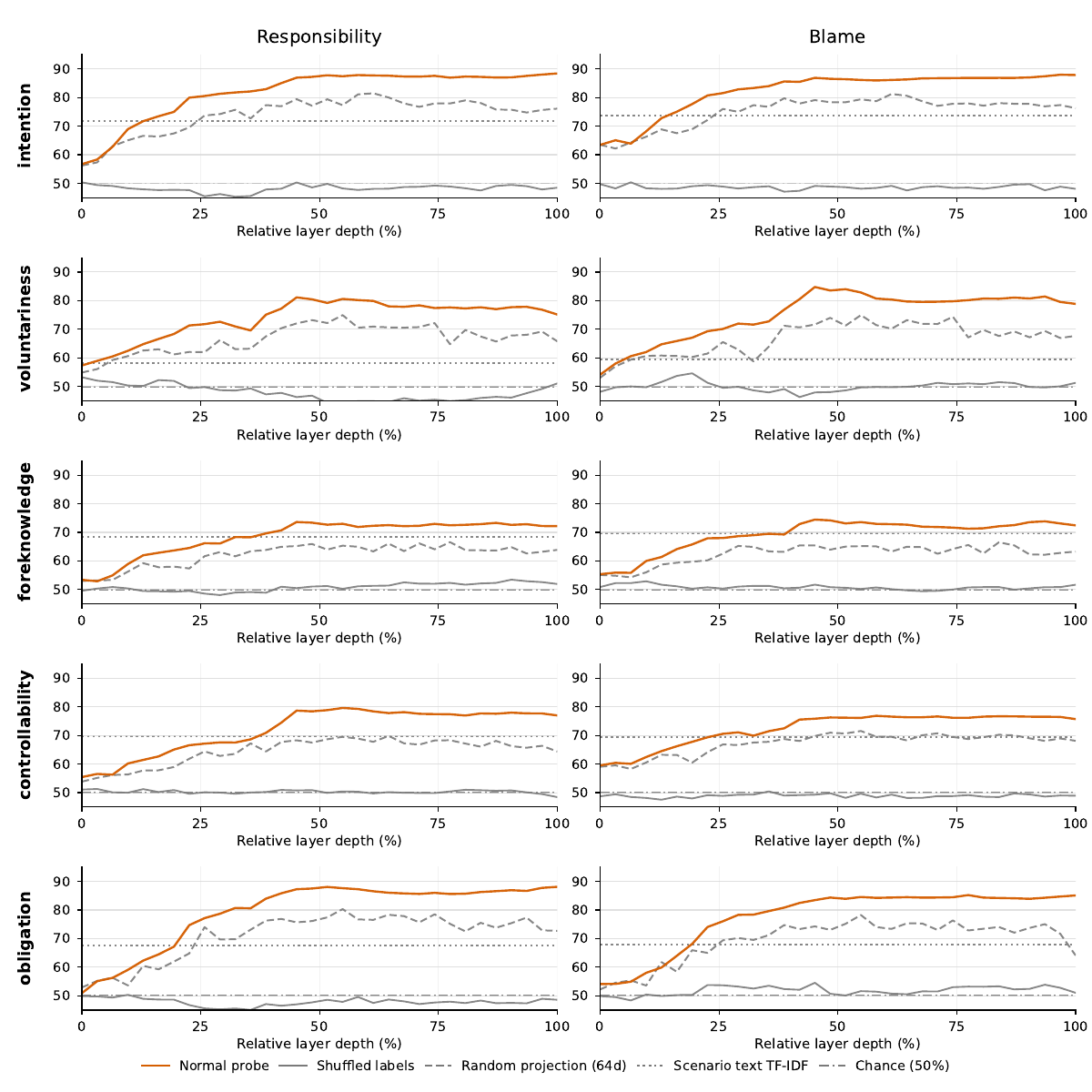}
  \caption{Sanity-check results at the label-token prediction step for Llama-3.1-8B, including shuffled-label, 64-dimensional random-projection, scenario-text lexical, and chance controls.}
  \label{fig:app-step1-sanity-llama31-8b}
\end{figure}

\begin{figure}[!htbp]
  \centering
  \includegraphics[width=1.0\linewidth]{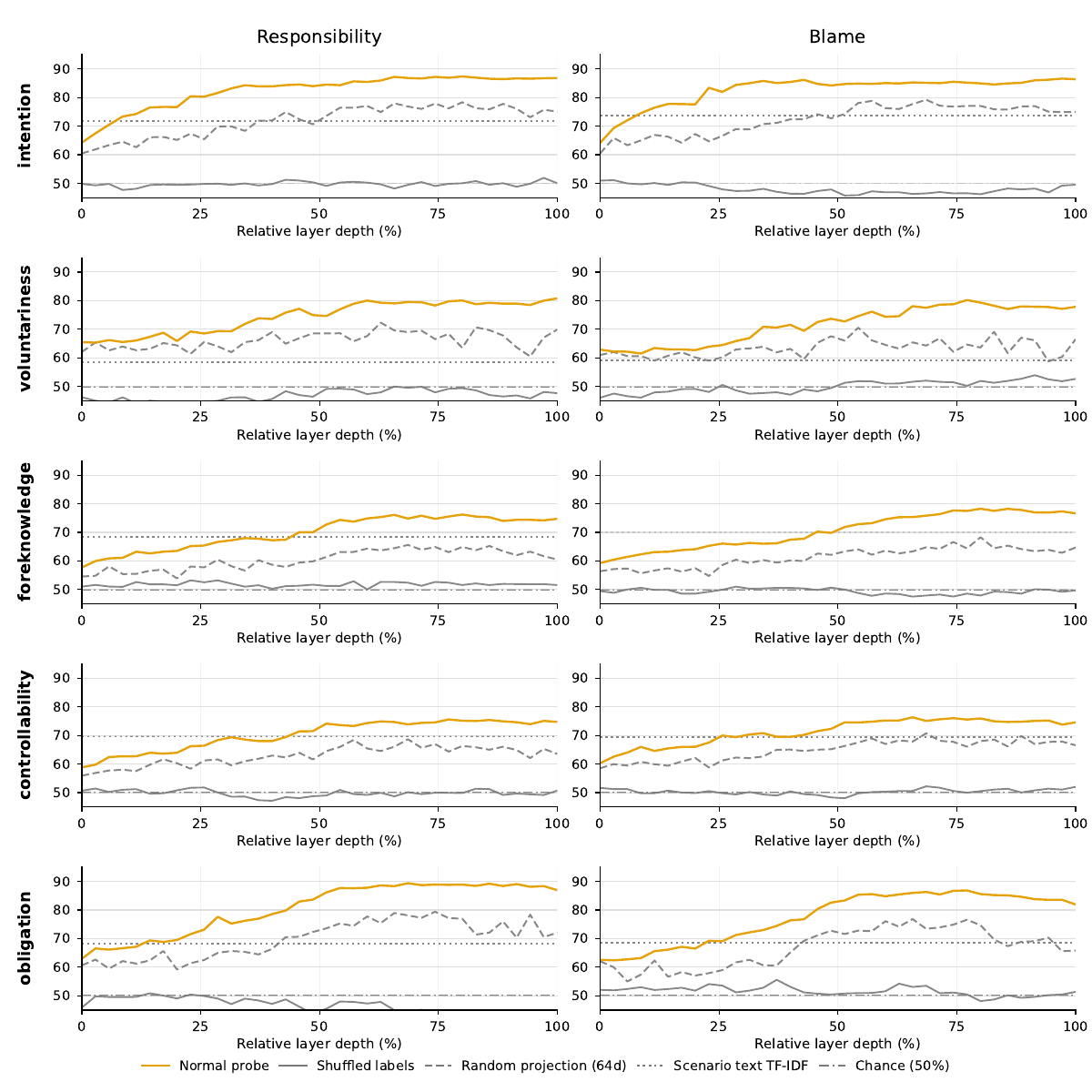}
  \caption{Sanity-check results at the label-token prediction step for Qwen3-4B, including shuffled-label, 64-dimensional random-projection, scenario-text lexical, and chance controls.}
  \label{fig:app-step1-sanity-qwen3-4b}
\end{figure}

\begin{figure}[!htbp]
  \centering
  \includegraphics[width=1.0\linewidth]{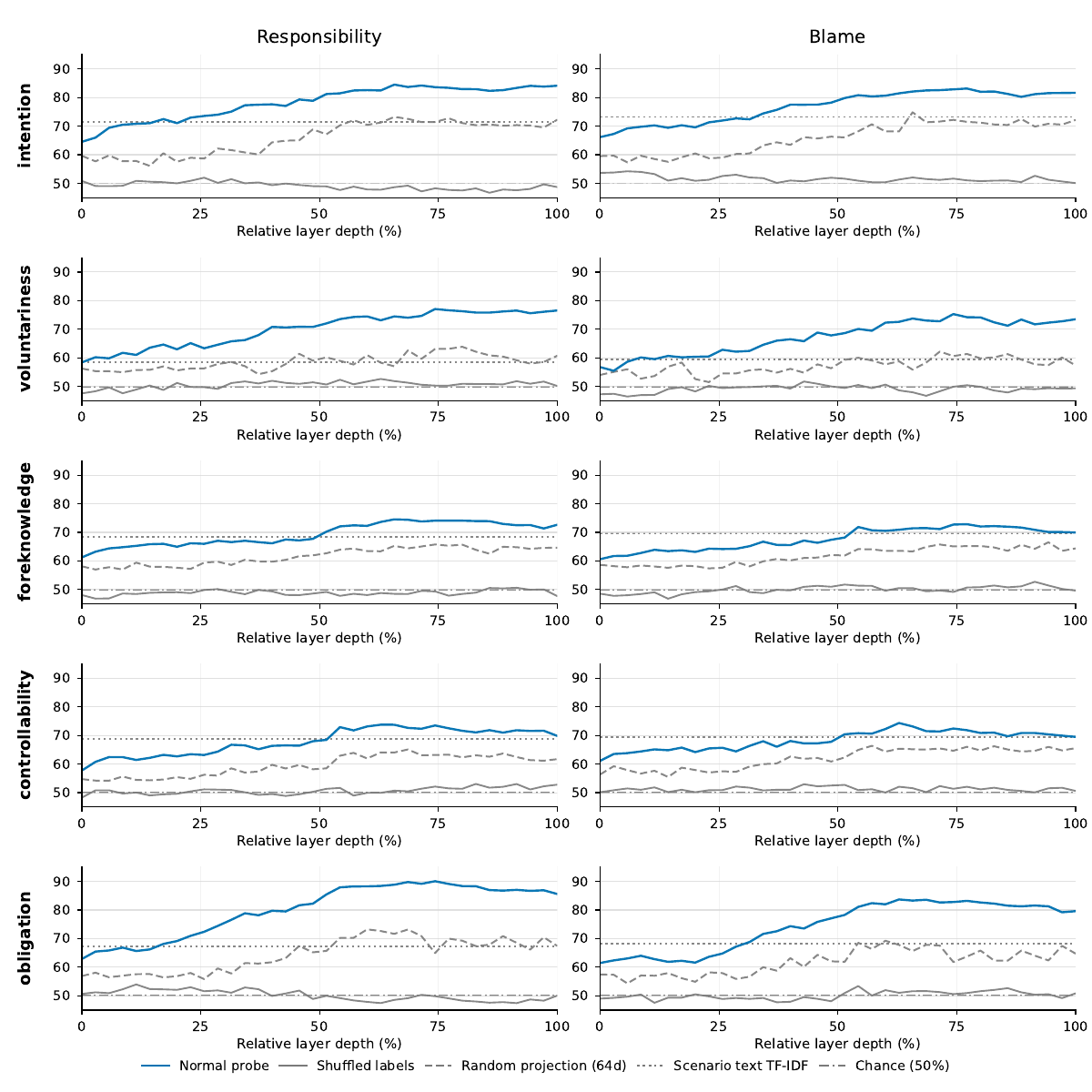}
  \caption{Sanity-check results at the label-token prediction step for Qwen3-8B, including shuffled-label, 64-dimensional random-projection, scenario-text lexical, and chance controls.}
  \label{fig:app-step1-sanity-qwen3-8b}
\end{figure}

Figures~\ref{fig:app-step1-sanity-deepseek-r1-15b}--\ref{fig:app-step1-sanity-qwen3-8b} report the results for all five backbones. The original probes consistently outperform the shuffled-label control, which remains near chance, and generally outperform the 64-dimensional random-projection control. For Llama-3.1-8B, Qwen3-4B, and Qwen3-8B, they also exceed the scenario-text lexical control across most layers. These comparisons support the usefulness of the hidden-state measurements beyond the tested controls, without implying that all lexical or contextual correlations have been removed. In particular, the lexical comparison is not uniformly favorable across every model and layer.

%% file: appendix_sections/early_token_step1.tex
\begin{figure}[!htbp]
\centering
\includegraphics[width=\linewidth]{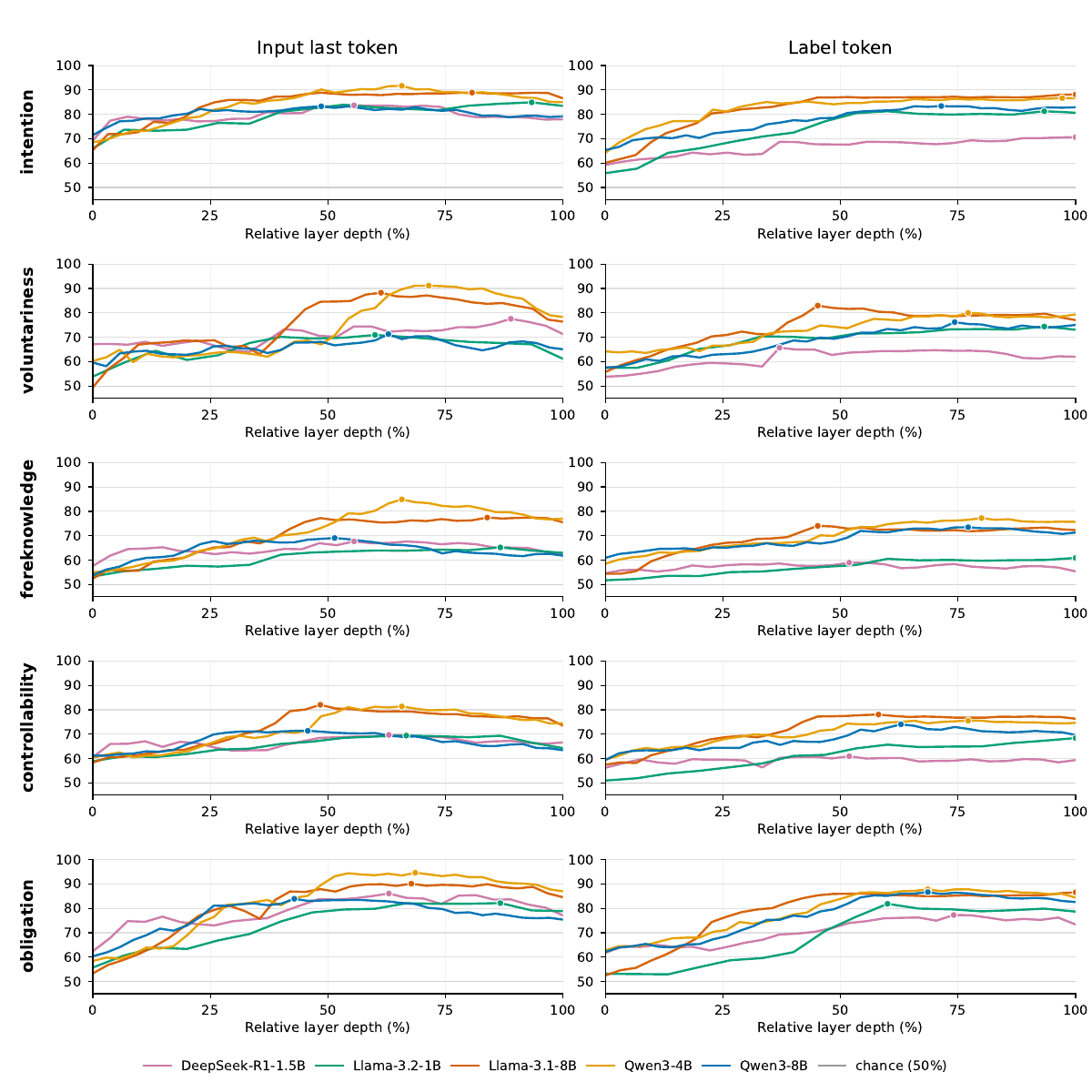}
\caption{Layerwise decodability at the last input token and the label-token prediction step. Layer depth is normalized within each model; dots indicate peak decodability.}
\label{fig:step1-decodability-by-probing-location-app}
\end{figure}

The main probes read the hidden state used to predict the final judgment label, where information from the prompt and preceding generated reasoning is available. A possible concern is that decodability at this position reflects only the immediate answer-selection stage. To examine whether attribution information is accessible earlier, we instead extract the hidden state at the last token of the input prompt, before the model generates its response, and train the same type of probes. Figure~\ref{fig:step1-decodability-by-probing-location-app} compares this last-input-token position with the label-token prediction step.

Attribution dimensions are already strongly decodable at the earlier position. Averaged over the displayed models, dimensions, and layers, balanced accuracy is 73.8\% at the last input token and 71.1\% at the label-token prediction step. Four of the five models have higher average decodability at the input position, with the largest difference in DeepSeek-R1-1.5B. Qwen3-8B is the main exception: its later position improves decoding of voluntariness, foreknowledge, and controllability. Llama-3.2-1B also improves on voluntariness at the later position.

At both positions, decodability generally rises from early layers and peaks in the middle or later layers. Intention and obligation remain among the most decodable dimensions, and the layerwise curves often have similar shapes. This comparison shows that the measured attribution information is not confined to the final label-prediction step. It does not imply that generating reasoning uniformly increases decodability, or that the two positions implement the same computation. Whether an earlier representation can also influence the final judgment is examined separately in Appendix~\ref{app:early_token_influence}.

%% file: appendix_sections/intervention_results.tex
\section{Additional Intervention Analyses}
\label{app:additional_interventions}

Decodability establishes that a dimension is accessible to a linear readout, but does not by itself show that changing the corresponding representation affects the model's judgment. The intervention analyses address this distinction. Complementing the main directional-test heatmap in Figure~\ref{fig:dimension_influence}, we test an earlier intervention position and examine how directional changes vary with perturbation strength.

\subsection{Intervention at the Last Input Token}
\label{app:early_token_influence}
\label{app:early_token_step2}

\input{appendix_sections/early_token_step2}

\subsection{Sensitivity to Intervention Strength}
\label{app:intervention_strength}

\begin{figure}[!htbp]
\centering
\includegraphics[width=\linewidth]{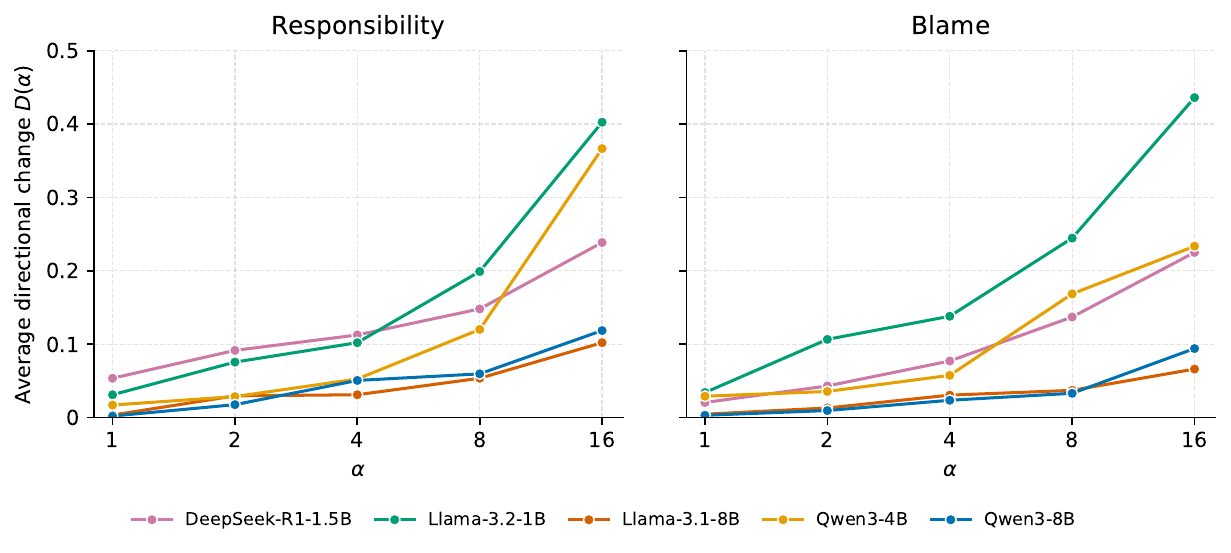}
\caption{Average signed judgment change $D(\alpha)$ at different intervention strengths, shown separately for responsibility and blame. Each model's curve averages $D_{l,k}(\alpha)$ equally across evaluated layers and dimensions. Positive values indicate a net theory-aligned change, not a directional-success rate or a significance result.}
\label{fig:alpha_analysis}
\end{figure}

The binary tests assess question-level directional success, but do not quantify the magnitude of judgment changes. We complement them with a descriptive sensitivity analysis, varying $\alpha\in\{1,2,4,8,16\}$. For a fixed model and task, let $\mathcal{I}_{l,k}$ denote the evaluated held-out questions at layer $l$ and dimension $k$. Using the same feasible signs $\mathcal{E}_i$ determined by the unedited output and the baseline-relative differences in Equation~\ref{eq:directional_difference}, we first calculate
\begin{equation}
D_{l,k}(\alpha)=\frac{1}{|\mathcal{I}_{l,k}|}
\sum_{i\in\mathcal{I}_{l,k}}
\frac{1}{|\mathcal{E}_i|}\sum_{\eta\in\mathcal{E}_i}\Delta_{i,\eta}^{\mathrm{base}}.
\label{eq:directional_change}
\end{equation}
An increase under a positive intervention and a decrease under a negative intervention contribute positively; unchanged judgments contribute zero. Each curve in Figure~\ref{fig:alpha_analysis} then gives the equal-weight mean across the evaluated layers $\mathcal{L}$ (Table~\ref{tab:probing_backbone_layers}) and the five dimensions $\mathcal{K}$:
\begin{equation}
D(\alpha)=\frac{1}{|\mathcal{L}|\,|\mathcal{K}|}
\sum_{l\in\mathcal{L}}\sum_{k\in\mathcal{K}}D_{l,k}(\alpha).
\label{eq:directional_change_aggregate}
\end{equation}
This continuous measure first averages the feasible signed changes within each question, then averages across questions and layer--dimension settings. The binary test instead assigns one success only when every feasible sign strictly satisfies the criterion. Thus, $D(\alpha)>0$ does not imply a directional-success rate above $1/2$ or that the directional tests pass. A value near zero indicates little net directional change, including unchanged judgments or changes that cancel.

Figure~\ref{fig:alpha_analysis} shows that the average directional change generally increases as $\alpha$ grows from 1 to 16 for both tasks. Over this full range, Llama-3.2-1B and Qwen3-4B show the largest increases, while the 8B models and DeepSeek-R1-1.5B show smaller positive increases. The ordering of model sensitivities can differ at individual strengths. These curves describe how average change magnitude varies with intervention strength; they do not establish that the directional-success tests pass at every strength. A baseline-relative magnitude alone also does not establish specificity to an attribution dimension, which is assessed by the matched-control tests.

Cross-model differences should be interpreted within this intervention setup. The same numerical $\alpha$ specifies the norm of a unit-direction perturbation, but need not represent the same relative change in hidden-state scale across backbones. The observed sensitivity therefore does not by itself establish that smaller models have intrinsically simpler attribution mechanisms or that larger models are generally more robust.

%% file: appendix_sections/early_token_step2.tex
The main intervention edits a hidden state at the label-token prediction step. This position directly tests whether the learned attribution directions can influence the final judgment, but its proximity to the output leaves open whether the effect is restricted to local answer selection. The earlier-token comparison addresses this concern by moving both probe training and intervention to the last token of the input prompt.

For each prompt, we extract the last-input-token hidden state and train attribution-dimension probes at that position. We then perturb that state along the corresponding probe-derived direction and allow autoregressive generation to continue until the model produces its final attribution label. The selected layers, strengths, directional hypotheses, random-direction controls, and shuffled-label probe-direction controls follow the main intervention setup. Each comparison uses the same feasible signs $\mathcal{E}_i$ determined by the unedited output: a question succeeds only if every tested sign strictly aligns, as in Equation~\ref{eq:joint_directional_success}; ties count as failures and all questions remain in $N$. We use the same one-sided exact binomial test with threshold $1/2$, requiring $p_c<0.01$ for all three comparisons (Appendix~\ref{app:directional_test}). Unlike the main one-step label prediction, the earlier edit must propagate through subsequent generation before the final judgment is observed.

\begin{figure}[!htbp]
\centering
\includegraphics[width=\linewidth]{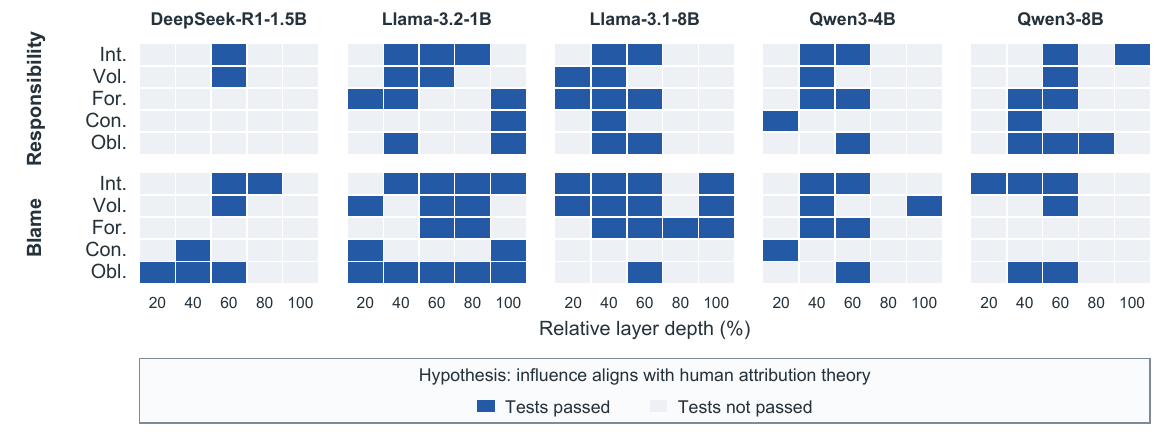}
\caption{Layerwise influence of attribution dimensions when interventions are applied at the last input token, with $\alpha=8$. Blue cells pass all three question-level directional-success tests ($p<0.01$); gray cells do not. Rows denote dimensions; 100\% denotes the final layer. Int., Vol., For., Con., and Obl. denote intention, voluntariness, foreknowledge, controllability, and obligation.}
\label{fig:early_token_influence}
\end{figure}

Figure~\ref{fig:early_token_influence} reports the earlier-position results at $\alpha=8$, using the same visual encoding as Figure~\ref{fig:dimension_influence}. Theory-aligned effects occur in several models and intermediate layers, particularly in the Llama family, but are unevenly distributed across models, dimensions, and layers. Together with Appendix~\ref{app:early_token_decodability}, this separates two questions: attribution information can already be read out before response generation, whereas the directional tests assess whether an edit at that position affects the eventual judgment.

The surviving effects show that influence along these probe-derived directions is not confined to an edit immediately before the answer label. However, the model- and layer-dependent pattern limits claims about a broadly stable mechanism throughout generation. An earlier local perturbation may be attenuated or transformed during subsequent computation; the present comparison measures its eventual effect rather than identifying the specific process responsible for that attenuation.

%% file: appendix_sections/discussion_limitations.tex
\section{Discussion and Limitations}
\label{app:discussion_limitations}

\subsection{Dataset Coverage and Human-Judgment Reference}
\label{app:dataset_scope}

The two subsets address complementary requirements. Controlled \textit{Vignette} scenarios make attribution dimensions and agent relationships explicit enough to vary systematically, supporting analysis that would be difficult with unconstrained narratives alone. The \textit{Reality} subset introduces more ambiguous situations grounded in everyday social conflicts. Together, they broaden the evaluation beyond a single form of evidence, but neither subset exhausts social attribution. Vignettes remain constructed scenarios, and reconstructed narratives omit details of the original interactions. The benchmark also focuses on negative outcomes and responsibility/blame judgments; the findings do not directly establish performance on positive credit attribution or other forms of social explanation.

The \textit{Reality} subset draws from a particular online community whose posting practices and selection of interpersonal conflicts shape the source material. Its cultural and platform coverage is therefore limited. Providing translated versions does not broaden the underlying cultural coverage.

The reference labels reflect the independently elicited judgments of two psychology experts under the supplied guidelines. They provide a consistent basis for comparison, rather than a population-wide consensus or a uniquely correct moral verdict. Keeping only questions with matching judgments avoids forcing consensus on disputed cases, but may preferentially retain clearer or less contentious items. Similar aggregate distributions before and after filtering would not rule out all such selection effects. Model accuracy should therefore be read as agreement with this defined human reference, within the dataset's coverage.

\subsection{Interpretation of Probing and Intervention Evidence}
\label{app:interpretation_limits}

The framework distinguishes three evidential questions: whether models agree with human judgments, whether attribution dimensions are linearly decodable, and whether edits along probe-derived directions influence judgments. This separation makes it possible to study intermediate variables in social reasoning rather than treating final-label agreement as the sole indicator of capability. Decodability alone, however, establishes that information is accessible to a particular classifier; it does not establish that the model uses that information in the same way as a human reasoner \citep{hewitt2019designing,belinkov2022probing}.

Several design choices strengthen the interpretation of the probing results. Vignette phrasing, names, and genders are varied, and dimension labels are not explicitly supplied in the scenario text. Its eight-fold leave-one-scenario-type-out evaluation holds each type out across all agent relations, testing generalization to unseen event settings and reducing reliance on recurring scenario content. Shuffled-label, random-projection, and lexical controls test complementary explanations for probe performance. Matched random-direction and shuffled-label interventions further test whether the observed judgment shifts depend on the learned dimension direction. These safeguards provide evidence beyond an unconstrained probe fit, while leaving open representational explanations that the controls do not directly test.

The dimensions can be correlated within narratives, and residualizing one probe weight against other probe weights removes only shared linear components captured by those probes. It does not establish complete independence among the underlying constructs or isolate every feature affected by an edit. Interventions at the label-token prediction step also demonstrate influence at a particular stage of output production; they do not recover the entire process by which the model arrived at its judgment. The earlier-token comparisons help characterize this position dependence, but do not establish equivalence to the sequential cognitive architectures of attribution theory. Dimension interactions and nonlinear representations remain incompletely characterized by the present approach.

The internal analyses cover five backbones and specific prompts, layers, token positions, and perturbation strengths. Their patterns should not be generalized to all models or inference settings. In particular, differences among the backbones confound parameter scale, model family, and training. Explanations based on more distributed representations in larger models or information distributed across a reasoning trajectory remain hypotheses rather than established mechanisms. Likewise, failure to pass the directional tests indicates insufficient evidence under the tested setting; it does not prove that the model ignores the corresponding dimension.

\subsection{Data Ethics and Responsible Use}
\label{app:ethics_statement}

The study concerns judgments about fictional or reconstructed agents for research evaluation. The \textit{Vignette} scenarios use fictional characters. For \textit{Reality}, publicly available posts serve as sources for reconstruction rather than being reproduced as the benchmark text. The preprocessing described in Appendix~\ref{app:reality_preprocessing} removes identifying information and unrelated private details, and substitutes fictional names. These measures aim to reduce privacy risks while retaining the event structure needed for attribution analysis.

Annotators were informed of the task purpose, the annotation content, and the intended use of their work. Participation was voluntary, and annotators could withdraw without providing a reason. The task did not require disclosure of personal experiences or sensitive personal information. Because scenarios include accidents, injury, conflict, and negligence, annotators were informed that they could encounter negative content; construction sought to avoid unnecessarily graphic, humiliating, or inflammatory descriptions.

The benchmark and dimension annotations can support comparative evaluation, analysis of systematic judgment differences, and investigation of representations underlying social attribution. These research uses do not establish suitability for adjudicating real disputes. Responsibility and blame involve social, moral, and sometimes legal considerations that the benchmark cannot fully represent. The labels and model outputs should not serve as the basis for decisions about real individuals in legal proceedings, disciplinary action, hiring, insurance compensation, or other consequential assessments. Agreement with the benchmark is therefore evidence about model behavior under a specified research protocol, not authorization to replace human judgment or accountable decision procedures.

%% file: iclr2027_conference.bib
@book{weiner1995judgments,
  title={Judgments of responsibility: A foundation for a theory of social conduct},
  author={Weiner, Bernard},
  year={1995},
  publisher={Guilford Press}
}

@book{weiner1986,
  title={An attributional theory of motivation and emotion},
  year={1986},
  author={Weiner, Bernard},
  publisher={Springer-Verlag}
}

@article{finch1987vignette,
  title={The vignette technique in survey research},
  author={Finch, Janet},
  journal={Sociology},
  volume={21},
  number={1},
  pages={105--114},
  year={1987},
  publisher={British Sociological Association}
}

@article{aguinis2014best_vignette,
  title={Best practice recommendations for designing and implementing experimental vignette methodology studies},
  author={Aguinis, Herman and Bradley, Kyle J},
  journal={Organizational Research Methods},
  volume={17},
  number={4},
  pages={351--371},
  year={2014},
  publisher={Sage Publications Sage CA: Los Angeles, CA}
}

@article{baguley2022vignette_statistical,
  title={Statistical modelling of vignette data in psychology},
  author={Baguley, Thom and Dunham, Grace and Steer, Oonagh},
  journal={British Journal of Psychology},
  volume={113},
  number={4},
  pages={1143--1163},
  year={2022},
  publisher={Wiley Online Library}
}

@article{malle2014PMoB,
  title={A theory of blame},
  author={Malle, Bertram F and Guglielmo, Steve and Monroe, Andrew E},
  journal={Psychological Inquiry},
  volume={25},
  number={2},
  pages={147--186},
  year={2014},
  publisher={Taylor \& Francis}
}

@article{mao2011IEEE-IS,
  title={From causal scenarios to social causality: An attributional approach},
  author={Mao, Wenji and Ge, Ansheng and Li, Xiaochen},
  journal={IEEE Intelligent Systems},
  volume={26},
  number={6},
  pages={48--57},
  year={2011},
  publisher={IEEE Computer Society}
}

@article{mao2012modeling-JAIR,
  title={Modeling social causality and responsibility judgment in multi-agent interactions},
  author={Mao, Wenji and Gratch, Jonathan},
  journal={Journal of Artificial Intelligence Research},
  volume={44},
  pages={223--273},
  year={2012}
}

@book{shaver1985,
  title={The attribution of blame: Causality, responsibility, and blameworthiness},
  author={Shaver, Kelly G},
  year={1985},
  publisher={Springer Science \& Business Media}
}

@article{shaver1986causality,
  title={On causality, responsibility, and self-blame: a theoretical note},
  author={Shaver, Kelly G and Drown, Debra},
  journal={Journal of Personality and Social Psychology},
  volume={50},
  number={4},
  pages={697-702},
  year={1986},
  publisher={American Psychological Association}
}

@inproceedings{park2023gsocial_simulate,
  title={Generative agents: Interactive simulacra of human behavior},
  author={Park, Joon Sung and O'Brien, Joseph and Cai, Carrie Jun and Morris, Meredith Ringel and Liang, Percy and Bernstein, Michael S},
  booktitle={Proceedings of UIST},
  pages={1--22},
  year={2023}
}

@article{achiam2023gpt,
  title={{GPT}-4 Technical Report},
  author={Achiam, Josh and Adler, Steven and Agarwal, Sandhini and Ahmad, Lama and Akkaya, Ilge and Aleman, Florencia Leoni and Almeida, Diogo and Altenschmidt, Janko and Altman, Sam and Anadkat, Shyamal and others},
  journal={arXiv preprint arXiv:2303.08774},
  year={2023}
}

@article{weiner1985attributional,
  title={An attributional theory of achievement motivation and emotion},
  author={Weiner, Bernard},
  journal={Psychological Review},
  volume={92},
  number={4},
  pages={548--573},
  year={1985},
  publisher={American Psychological Association}
}

@article{ji2025moralbench,
  title={Moralbench: Moral evaluation of {LLMs}},
  author={Ji, Jianchao and Chen, Yutong and Jin, Mingyu and Xu, Wujiang and Hua, Wenyue and Zhang, Yongfeng},
  journal={ACM SIGKDD Explorations Newsletter},
  volume={27},
  number={1},
  pages={62--71},
  year={2025},
  publisher={ACM New York, NY, USA}
}

@inproceedings{forbes2020socialnorm,
  title={Social Chemistry 101: Learning to Reason about Social and Moral Norms},
  author={Forbes, Maxwell and Hwang, Jena D and Shwartz, Vered and Sap, Maarten and Choi, Yejin},
  booktitle={Proceedings of EMNLP},
  pages={653--670},
  year={2020}
}

@article{jiang2021socialnorm,
  title={Can machines learn morality? {The} delphi experiment},
  author={Jiang, Liwei and Hwang, Jena D and Bhagavatula, Chandra and Bras, Ronan Le and Liang, Jenny and Dodge, Jesse and Sakaguchi, Keisuke and Forbes, Maxwell and Borchardt, Jon and Gabriel, Saadia and others},
  journal={arXiv preprint arXiv:2110.07574},
  year={2021}
}

@inproceedings{yuan2024socialnorm,
  title={Measuring Social Norms of Large Language Models},
  author={Yuan, Ye and Tang, Kexin and Shen, Jianhao and Zhang, Ming and Wang, Chenguang},
  booktitle={Findings of NAACL},
  pages={650--699},
  year={2024}
}

@inproceedings{guha2023legalbench,
  title={Legalbench: A collaboratively built benchmark for measuring legal reasoning in large language models},
  author={Guha, Neel and Nyarko, Julian and Ho, Daniel and R{\'e}, Christopher and Chilton, Adam and Chohlas-Wood, Alex and Peters, Austin and Waldon, Brandon and Rockmore, Daniel and Zambrano, Diego and others},
  booktitle={Proceedings of NeurIPS},
  pages={44123--44279},
  year={2023}
}

@inproceedings{fei2024lawbench,
  title={{LawBench}: Benchmarking legal knowledge of large language models},
  author={Fei, Zhiwei and Shen, Xiaoyu and Zhu, Dawei and Zhou, Fengzhe and Han, Zhuo and Huang, Alan and Zhang, Songyang and Chen, Kai and Yin, Zhixin and Shen, Zongwen and others},
  booktitle={Proceedings of EMNLP},
  pages={7933--7962},
  year={2024}
}

@inproceedings{dai2025law,
  title={{LAiW}: A {Chinese} legal large language models benchmark},
  author={Dai, Yongfu and Feng, Duanyu and Huang, Jimin and Jia, Haochen and Xie, Qianqian and Zhang, Yifang and Han, Weiguang and Tian, Wei and Wang, Hao},
  booktitle={Proceedings of COLING},
  pages={10738--10766},
  year={2025}
}

@article{cohen1960kappa1,
  title={A coefficient of agreement for nominal scales},
  author={Cohen, Jacob},
  journal={Educational and Psychological Measurement},
  volume={20},
  number={1},
  pages={37--46},
  year={1960},
  publisher={Sage Publications Sage CA: Thousand Oaks, CA}
}

@article{cohen1968weightedkappa,
  title={Weighted kappa: Nominal scale agreement provision for scaled disagreement or partial credit},
  author={Cohen, Jacob},
  journal={Psychological Bulletin},
  volume={70},
  number={4},
  pages={213--220},
  year={1968},
  publisher={American Psychological Association}
}

@article{fleiss1973quadratically_weighted_kappa,
  title={The equivalence of weighted kappa and the intraclass correlation coefficient as measures of reliability},
  author={Fleiss, Joseph L and Cohen, Jacob},
  journal={Educational and Psychological Measurement},
  volume={33},
  number={3},
  pages={613--619},
  year={1973},
  publisher={Sage Publications Sage CA: Thousand Oaks, CA}
}

@inproceedings{meng2022locating,
  title={Locating and editing factual associations in {GPT}},
  author={Meng, Kevin and Bau, David and Andonian, Alex and Belinkov, Yonatan},
  booktitle={Proceedings of NeurIPS},
  pages={17359--17372},
  year={2022}
}

@inproceedings{hernandezlinearity,
  title={Linearity of Relation Decoding in Transformer Language Models},
  author={Hernandez, Evan and Sharma, Arnab Sen and Haklay, Tal and Meng, Kevin and Wattenberg, Martin and Andreas, Jacob and Belinkov, Yonatan and Bau, David},
  booktitle={Proceedings of ICLR},
  year={2024}
}

@inproceedings{heollms,
  title={Do {LLMs} "know" internally when they follow instructions?},
  author={Heo, Juyeon and Heinze-Deml, Christina and Elachqar, Oussama and Chan, Kwan Ho Ryan and Ren, Shirley You and Miller, Andrew and Nallasamy, Udhyakumar and Narain, Jaya},
  booktitle={Proceedings of ICLR},
  year={2025}
}

@inproceedings{orgadllms,
  title={{LLMs} Know More Than They Show: On the Intrinsic Representation of {LLM} Hallucinations},
  author={Orgad, Hadas and Toker, Michael and Gekhman, Zorik and Reichart, Roi and Szpektor, Idan and Kotek, Hadas and Belinkov, Yonatan},
  booktitle={Proceedings of ICLR},
  year={2025}
}

@inproceedings{tak-etal-2025-mechanistic,
    title = "Mechanistic Interpretability of Emotion Inference in Large Language Models",
    author = "Tak, Ala N.  and
      Banayeeanzade, Amin  and
      Bolourani, Anahita  and
      Kian, Mina  and
      Jia, Robin  and
      Gratch, Jonathan",
    booktitle = "Findings of ACL",
    year = "2025",
    pages = "13090--13120",
}

@inproceedings{plaza2024angry,
  title={Angry Men, Sad Women: Large Language Models Reflect Gendered Stereotypes in Emotion Attribution},
  author={Plaza-del-Arco, Flor and Curry, Amanda and Curry, Alba Cercas and Abercrombie, Gavin and Hovy, Dirk},
  booktitle={Proceedings of ACL},
  pages={7682--7696},
  year={2024}
}

@inproceedings{kang-etal-2025-values,
    title = "Are the Values of {LLMs} Structurally Aligned with Humans? {A} Causal Perspective",
    author = "Kang, Yipeng  and
      Wang, Junqi  and
      Li, Yexin  and
      Wang, Mengmeng  and
      Tu, Wenming  and
      Wang, Quansen  and
      Li, Hengli  and
      Wu, Tingjun  and
      Feng, Xue  and
      Zhong, Fangwei  and
      Zheng, Zilong",
    booktitle = "Findings of ACL",
    year = "2025",
    pages = "23147--23161",
}

@inproceedings{ju2025probing,
  title={Probing then editing response personality of large language models},
  author={Ju, Tianjie and Shao, Zhenyu and Wang, Bowen and Chen, Yujia and Zhang, Zhuosheng and Fei, Hao and Lee, Mong-Li and Hsu, Wynne and Duan, Sufeng and Liu, Gongshen},
   booktitle = {Proceedings of COLM},
  year={2025}
}

@inproceedings{zhang-2025-probing,
    title = "Probing Political Ideology in Large Language Models: How Latent Political Representations Generalize Across Tasks",
    author = "Zhang, Tianyi",
    booktitle = "Findings of EMNLP",
    year = "2025",
    pages = "23349--23360"
}

@article{belinkov2019analysis,
  title={Analysis Methods in Neural Language Processing: A Survey},
  author={Belinkov, Yonatan and Glass, James},
  journal={Transactions of the Association for Computational Linguistics},
  volume={7},
  pages={49--72},
  year={2019}
}

@article{belinkov2022probing,
  title={Probing classifiers: Promises, shortcomings, and advances},
  author={Belinkov, Yonatan},
  journal={Computational Linguistics},
  volume={48},
  number={1},
  pages={207--219},
  year={2022}
}

@inproceedings{hewitt2019designing,
  title={Designing and Interpreting Probes with Control Tasks},
  author={Hewitt, John and Liang, Percy},
  booktitle={Proceedings of EMNLP},
  pages={2733--2743},
  year={2019}
}

@article{xu2024llmsocial_survey_ipm,
  title={{AI} for social science and social science of {AI}: A survey},
  author={Xu, Ruoxi and Sun, Yingfei and Ren, Mengjie and Guo, Shiguang and Pan, Ruotong and Lin, Hongyu and Sun, Le and Han, Xianpei},
  journal={Information Processing \& Management},
  volume={61},
  number={3},
  pages={103665},
  year={2024},
  publisher={Elsevier}
}

@inproceedings{zhang2025personalized_pca,
  title={Personalized text generation with contrastive activation steering},
  author={Zhang, Jinghao and Liu, Yuting and Wang, Wenjie and Liu, Qiang and Wu, Shu and Wang, Liang and Chua, Tat-Seng},
  booktitle={Proceedings of ACL},
  pages={7128--7141},
  year={2025}
}

@inproceedings{huben2023sparse_sae,
  title={Sparse autoencoders find highly interpretable features in language models},
  author={Huben, Robert and Cunningham, Hoagy and Smith, Logan Riggs and Ewart, Aidan and Sharkey, Lee},
  booktitle={Proceedings of ICLR},
  year={2024}
}

@article{malle2022attribution,
  title={Attribution theories: How people make sense of behavior},
  author={Malle, Bertram F},
  journal={Theories in Social Psychology},
  pages={93--120},
  year={2022},
  publisher={Wiley Online Library}
}

@inproceedings{goindani2020social,
  title={Social reinforcement learning to combat fake news spread},
  author={Goindani, Mahak and Neville, Jennifer},
  booktitle={Proceedings of UAI},
  pages={1006--1016},
  year={2019}
}

@inproceedings{chakraborty2025structuredmoral,
  title={Structured moral reasoning in language models: A value-grounded evaluation framework},
  author={Chakraborty, Mohna and Wang, Lu and Jurgens, David},
  booktitle={Proceedings of EMNLP},
  pages={30295--30323},
  year={2025}
}

@inproceedings{jinlanguage,
  title={Language Model Alignment in Multilingual Trolley Problems},
  author={Jin, Zhijing and Kleiman-Weiner, Max and Piatti, Giorgio and Levine, Sydney and Liu, Jiarui and Adauto, Fernando Gonzalez and Ortu, Francesco and Strausz, Andr{\'a}s and Sachan, Mrinmaya and Mihalcea, Rada and others},
  booktitle={Proceedings of ICLR},
  year={2025}
}

@inproceedings{agarwal2024ethical,
  title={Ethical reasoning and moral value alignment of {LLMs} depend on the language we prompt them in},
  author={Agarwal, Utkarsh and Tanmay, Kumar and Khandelwal, Aditi and Choudhury, Monojit},
  booktitle={Proceedings of LREC-COLING},
  pages={6330--6340},
  year={2024}
}

@article{hadan2025responsible,
  title={Who is Responsible When {AI} Fails? {Mapping} Causes, Entities, and Consequences of {AI} Privacy and Ethical Incidents},
  author={Hadan, Hilda and Hadi Mogavi, Reza and Zhang-Kennedy, Leah and Nacke, Lennart E},
  journal={International Journal of Human--Computer Interaction},
  pages={1--45},
  year={2025},
  publisher={Taylor \& Francis}
}

@article{gratch2006modeling,
  title={Modeling social emotions and social attributions},
  author={Gratch, Jonathan and Mao, Wenji and Marsella, Stacy},
  journal={Cognition and Multi-agent Interaction},
  pages={219--251},
  year={2006}
}

@article{QI2023103510,
title = {What is the limitation of multimodal {LLMs}? A deeper look into multimodal {LLMs} through prompt probing},
journal = {Information Processing \& Management},
volume = {60},
number = {6},
pages = {103510},
year = {2023},
issn = {0306-4573},
author = {Shuhan Qi and Zhengying Cao and Jun Rao and Lei Wang and Jing Xiao and Xuan Wang},
}

@article{dahl2024large_legal,
  title={Large legal fictions: Profiling legal hallucinations in large language models},
  author={Dahl, Matthew and Magesh, Varun and Suzgun, Mirac and Ho, Daniel E},
  journal={Journal of Legal Analysis},
  volume={16},
  number={1},
  pages={64--93},
  year={2024},
  publisher={Oxford University Press UK}
}

@inproceedings{moore2025expressing,
  title={Expressing stigma and inappropriate responses prevents LLMs from safely replacing mental health providers},
  author={Moore, Jared and Grabb, Declan and Agnew, William and Klyman, Kevin and Chancellor, Stevie and Ong, Desmond C and Haber, Nick},
  booktitle={Proceedings of ACM FAccT},
  pages={599--627},
  year={2025}
}
